%% file: iclr2026_conference.tex
\PassOptionsToPackage{dvipsnames,table}{xcolor}
\documentclass{article}
\usepackage{iclr2026_conference,times}
\input{preamble}

\input{math_commands.tex}

\iclrfinalcopy

\title{GTR: Gated Token Recurrence for Efficient Dense Prediction}

\author{Zhe Feng$^{1,2,3}$, Longfei Liu$^{2}$, Wei Liu$^{1}$, Kai Chen$^{5}$, Jiangang Kong$^{1}$, Wei Zhou$^{1}$, \\
\textbf{Yifeng Qian}$^{5}$, \textbf{Dexiong Chen}$^{4}$, \textbf{Xuanlong Yu}$^{2}$, \textbf{Xi Shen}$^{2}$\thanks{Corresponding author: \texttt{shenxiluc@gmail.com}} \\
$^{1}$Didi International Business Group \quad $^{2}$Intellindust AI Lab \\
$^{3}$Institute of Automation, Chinese Academy of Sciences \\
$^{4}$The Hong Kong University of Science and Technology (Guangzhou) \quad $^{5}$Didi Research
}

\begin{document}
\maketitle
\pagestyle{plain}
\input{figs_tex/teaser}

\input{sec/0_abstract}
\input{sec/1_intro}
\input{sec/3_method}
\input{sec/4_exp}

\input{sec/5_conclusion}

\bibliography{iclr2026_conference}
\bibliographystyle{iclr2026_conference}
\input{sec/X_suppl}

\end{document}

%% file: preamble.tex
\usepackage{xcolor}
\definecolor{gtrtocred}{RGB}{140, 21, 21}
\usepackage{graphicx}
\usepackage{algorithm}
\usepackage{algpseudocode}
\usepackage[normalem]{ulem}
\usepackage{wrapfig}
\usepackage{needspace}
\usepackage{caption}
\usepackage{booktabs}
\usepackage{multirow}
\usepackage{makecell}
\usepackage{xcolor}
\usepackage{colortbl}
\usepackage{adjustbox}
\usepackage{pifont}
\newcommand{\cmark}{\ding{51}}
\newcommand{\xmark}{\ding{55}}
\definecolor{gtrblue}{HTML}{47B1E1}
\newcommand{\rankfirst}[1]{\cellcolor{gtrblue!32}#1}
\newcommand{\ranksecond}[1]{\cellcolor{gtrblue!20}#1}
\newcommand{\rankthird}[1]{\cellcolor{gtrblue!10}#1}
\usepackage[colorlinks=true, citecolor=blue, linkcolor=blue, urlcolor=blue]{hyperref}
\usepackage{url}
\DeclareRobustCommand{\iconhref}[3]{\texorpdfstring{\href{#1}{\raisebox{-0.15em}{\includegraphics[height=1em]{#2}}\,#3}}{#3}}
\newcommand{\gtrlinks}{\mdseries\iconhref{https://intellindust-ai-lab.github.io/projects/GTR/}{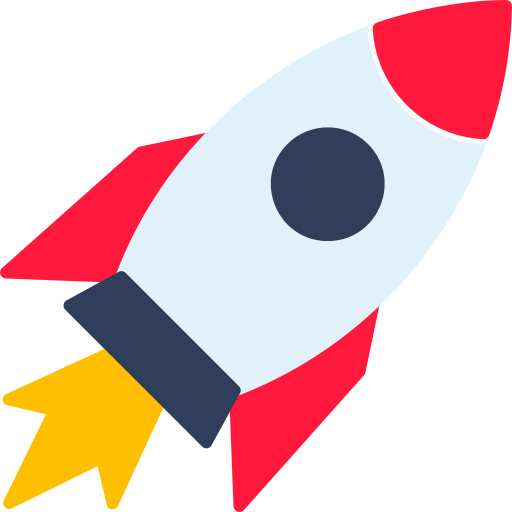}{Project Page}\quad\iconhref{https://github.com/Intellindust-AI-Lab/GTR}{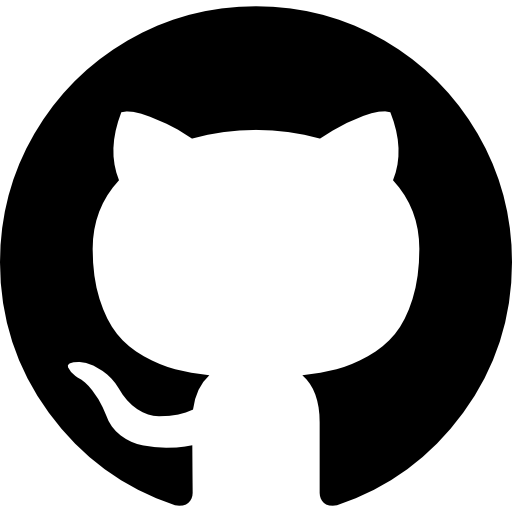}{GitHub}}
\usepackage{float}
\usepackage{placeins}

\usepackage{tablefootnote}

\definecolor{good}{RGB}{0,135,60}
\definecolor{bad}{RGB}{200,30,30}

\newlength{\figboxh}
\usepackage{titlesec}

\titlespacing*{\paragraph}
  {0pt}
  {0.75\baselineskip}
  {0.5em}

\AtBeginDocument{\setlength{\abovedisplayskip}{7pt plus 2pt minus 2pt}\setlength{\belowdisplayskip}{7pt plus 2pt minus 2pt}\setlength{\abovedisplayshortskip}{0pt plus 2pt}\setlength{\belowdisplayshortskip}{2pt plus 2pt minus 1pt}}

%% file: math_commands.tex
\usepackage{amsmath,amsfonts,bm}

\def\eqref#1{equation~\ref{#1}}

\def\1{\bm{1}}

\DeclareMathAlphabet{\mathsfit}{\encodingdefault}{\sfdefault}{m}{sl}
\SetMathAlphabet{\mathsfit}{bold}{\encodingdefault}{\sfdefault}{bx}{n}

%% file: figs_tex/teaser.tex
\begin{figure}[h]
\vspace{-20pt}
\centering
\captionsetup{skip=3pt}
  \includegraphics[width=\textwidth]{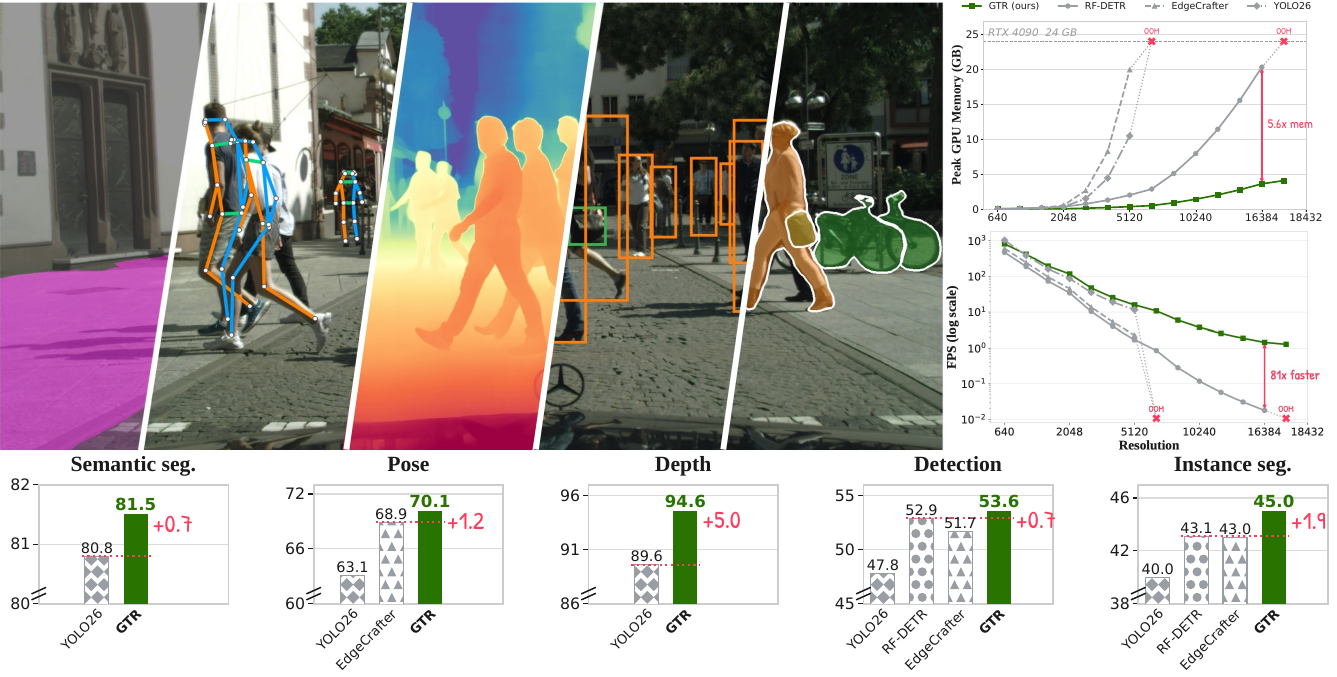}

\caption{\textbf{GTR: a shared backbone design for efficient dense prediction.}
	\textit{Left:} Task-specific models built on the same GTR backbone design produce predictions across semantic segmentation, human pose estimation, monocular depth estimation, object detection, and instance segmentation.
	\textit{Right:} Peak GPU memory and inference throughput on RTX~4090 as input resolution increases, comparing GTR-S with RF-DETR-S~\citep{robinson2025rf}, EdgeCrafter-S~\citep{liu2026edgecrafter}, and YOLO26-S~\citep{yolo26}. GTR-S shows slower peak-memory growth and higher throughput at increasing resolutions.}
  \label{fig:teaser}
\vspace{-8pt}
\end{figure}

%% file: sec/0_abstract.tex
\begin{abstract}
Self-attention-based vision backbones perform well on dense prediction, but the quadratic computational cost of global softmax attention limits their efficiency as image resolution increases. We introduce Gated Token Recurrence (GTR), a softmax-free recurrent vision backbone that combines gated linear attention, alternating spatial scan directions, and spatially enhanced SwiGLU blocks. GTR is distilled from a detection-specialized DINOv3 teacher using only final-layer patch-token alignment through a linear projection and squared $\ell_2$ loss, without masked-token prediction or intermediate-layer supervision. With Objects365 detector pre-training, GTR-L achieves 58.9 box AP on COCO \texttt{val2017} with 1.908\,ms median batch-one latency under compiled FP16 execution on an RTX~4090. The same backbone also transfers to instance segmentation, pose estimation, oriented detection, semantic segmentation, and monocular depth estimation. In an isolated kernel benchmark, our specialized chunkwise CUDA operator is $4.0\times$ faster than FLA v0.5.0 at 1.6K tokens on RTX~4090. TensorRT deployment on DRIVE AGX Thor achieves 2.282--8.769\,ms median batch-one latency across the evaluated models. These results show that recurrent token mixing can provide an efficient alternative to global softmax attention for high-resolution dense prediction and edge deployment.

\end{abstract}

%% file: sec/1_intro.tex
\section{Introduction}
\label{sec:intro}

Self-attention-based vision backbones have shown strong performance for dense prediction~\citep{liu2026edgecrafter}. However, the quadratic complexity of softmax attention with respect to the number of image tokens makes high-resolution processing increasingly expensive, particularly on resource-constrained edge devices. Linear attention offers a promising alternative: EfficientViT~\citep{cai2023efficientvit} demonstrates efficient on-device semantic segmentation, while ViG~\citep{liao2025vig} explores gated linear recurrence for detection and segmentation. Despite this progress, developing a general linear-attention backbone that performs consistently across diverse dense prediction tasks while meeting practical constraints on model size, computation, memory, and latency remains challenging. This motivates a central question: can linear attention support a general-purpose vision backbone that remains accurate across diverse dense prediction tasks while translating its theoretical efficiency into practical gains on edge hardware?

We introduce \textit{Gated Token Recurrence (GTR)}, a linear-complexity (in the number of image tokens) vision backbone designed to address this challenge. GTR builds on gated linear attention (GLA)~\citep{yang2024gla} with a softmax-free, single-scale recurrent architecture. It maintains a fixed patch grid across 12 blocks and performs one directional scan per block, cycling through four directions across depth to progressively aggregate global context. To complement this directional recurrence with explicit local spatial modeling, we introduce \textit{Spatial SwiGLU}, which incorporates depthwise convolution into the value branch of SwiGLU~\citep{shazeer2020glu} to efficiently mix neighboring patches. Together, directional recurrence and Spatial SwiGLU provide complementary global and local spatial modeling while preserving linear complexity.

Importantly, GTR uses a single distillation objective that supervises only the final patch tokens. Unlike ViT-Linearizer~\citep{wei2025vitlinearizer}, which combines intermediate activation matching with masked prediction, or ViT-AdaLA~\citep{li2026vitalada}, which aligns attention modules before the final features, our approach distills only the final patch tokens. Specifically, GTR is aligned with a frozen, detection-specialized DINOv3 teacher~\citep{simeoni2025dinov3} using the same full-image input. A linear projection and a single squared $\ell_2$ loss constitute the entire distillation objective, requiring neither masking nor intermediate supervision.

GTR scales across four model sizes (S, M, L, and X) and transfers to six dense prediction tasks: object detection, instance segmentation, human pose estimation, oriented object detection, semantic segmentation, and monocular depth estimation. With Objects365 detector pre-training~\citep{shao2019objects365}, GTR-L achieves 58.9 box AP on COCO~\citep{lin2014microsoft} \texttt{val2017} at a median forward-pass latency of only 1.908\,ms under compiled FP16, batch-one inference on an RTX~4090. Beyond detection, GTR achieves competitive results in representative comparisons across the other five tasks using the same backbone design with task-specific heads and training recipes (Section~\ref{sec:transfer_results}; Appendix~\ref{sec:experimental_setup}). As summarized in Figure~\ref{fig:teaser}, GTR-S also exhibits slower growth in peak GPU memory and higher inference throughput than the softmax-attention-based EdgeCrafter-S~\citep{liu2026edgecrafter} as input resolution increases, highlighting the practical advantage of linear-complexity attention for high-resolution dense prediction.

Finally, we bridge architectural efficiency and real-world deployment with dedicated system optimizations. We develop a specialized chunkwise CUDA operator for GLA~\citep{yang2024gla} that computes chunk summaries in parallel while propagating recurrent context through a boundary-state scan. The operator achieves a $4.0\times$ speedup over FLA v0.5.0~\citep{yang2024fla} at 1.6K tokens under FP16 and CUDA Graph execution on an RTX~4090. For end-to-end deployment on DRIVE AGX Thor, we further integrate fused kernels and graph transformations into TensorRT~\citep{nvidiatensorrt}. Across six tasks and four model scales, GTR achieves median FP16 batch-one latencies of 2.282--8.769\,ms, demonstrating its practicality for real-time dense prediction on edge hardware.

In summary, our main contributions are:
\begin{itemize}
	\item We introduce \textit{GTR}, a single-scale, linear-attention vision backbone for efficient dense prediction. GTR combines gated token recurrence with four-directional scanning across depth and Spatial SwiGLU for complementary global and local spatial modeling.
	
	\item We demonstrate the \textit{generality and scalability} of GTR across six dense prediction tasks and four model scales. The same backbone design transfers effectively across object-level and pixel-level tasks using simple final-token distillation and task-specific adaptation.
	
	\item We enable \textit{practical edge deployment} through a specialized chunkwise CUDA operator and TensorRT integration. Across six tasks and four model scales, GTR achieves low-latency FP16 inference on DRIVE AGX Thor, demonstrating low-latency deployment on automotive edge hardware.
\end{itemize}

%% file: sec/3_method.tex
\section{Method}
\label{sec:method}

Figure~\ref{fig:gtr_overview} provides an overview of GTR and its object detection pipeline. We first describe the GTR backbone, followed by its distillation strategy and specialized chunkwise inference. Given an image $\mathbf{I}\in\mathbb{R}^{H\times W\times3}$, GTR produces a stride-16 grid of $L=HW/16^2$ patch tokens and maintains the same resolution throughout 12 blocks. Each block combines gated linear attention (GLA)~\citep{yang2024gla} for long-range context with Spatial SwiGLU for local spatial mixing. The scan direction cycles through four directions across successive blocks to aggregate context from different orientations. For object detection, features from blocks 4, 8, and 12 are used to build a lightweight feature pyramid at strides 8, 16, and 32, followed by a DETR-style decoder~\citep{carion2020end,peng2025dfine}. The GTR backbone itself remains single-scale, while multi-scale features are constructed in the detection neck. Details of the neck and decoder are provided in Appendix~\ref{sec:encoder_decoder_structure}, and the four GTR model configurations in Appendix~\ref{sec:architecture_configurations}.

\input{figs_tex/gtr_pipeline.tex}

\subsection{GTR Backbone}
\label{sec:gtr_backbone}
\label{sec:preliminaries}

GTR combines three components: gated recurrence for long-range context, directional scanning for two-dimensional context, and Spatial SwiGLU for local spatial mixing.

\paragraph{Key-only gated recurrence.}
We adopt the key-only variant of GLA~\citep{yang2024gla}. For one attention head, let $\mathbf X\in\mathbb R^{L\times d}$ denote the input tokens. We first compute
\(\mathbf Q=\mathbf X\mathbf W_Q,
  \mathbf K=\mathbf X\mathbf W_K,
  \mathbf V=\mathbf X\mathbf W_V\),
where $\mathbf W_Q,\mathbf W_K\in\mathbb R^{d\times d_k}$ and
$\mathbf W_V\in\mathbb R^{d\times d_v}$. Let
$\boldsymbol x_t\in\mathbb R^{1\times d}$ be the input row at token $t$,
and let $\boldsymbol q_t,\boldsymbol k_t$, and $\boldsymbol v_t$ denote
the corresponding projected row vectors.

GLA maintains a recurrent state
\(\mathbf S_t\in\mathbb R^{d_k\times d_v}\)
that summarizes the tokens encountered along the current scan, with
\(\mathbf S_t=
\mathrm{Diag}(\boldsymbol\gamma_t)\mathbf S_{t-1}
+\boldsymbol k_t^\top\boldsymbol v_t\),
\(\boldsymbol o_t=d_k^{-1/2}\boldsymbol q_t\mathbf S_t\),
and \(\mathbf S_0=\mathbf0\).
Here,
\(\boldsymbol\gamma_t=
\sigma(\boldsymbol x_t\mathbf W_\alpha^{(1)}\mathbf W_\alpha^{(2)}
+\boldsymbol b_\alpha)^{1/\tau}\)
is a learned key-wise retention gate with
\(\boldsymbol\gamma_t\in(0,1)^{1\times d_k}\). Following GLA, the gate uses a
low-rank projection with
\(\mathbf W_\alpha^{(1)}\in\mathbb R^{d\times r}\),
\(\mathbf W_\alpha^{(2)}\in\mathbb R^{r\times d_k}\),
\(\boldsymbol b_\alpha\in\mathbb R^{1\times d_k}\), rank \(r=16\), and
temperature \(\tau=16\). The sigmoid \(\sigma\) and the power are applied
elementwise. Each gate entry scales one row of
\(\mathbf S_{t-1}\) across all \(d_v\) value channels.

GTR uses only this key-side state decay, with the value-side decay fixed
to one during both training and inference. A separate data-dependent output
gate is applied to the RMS-normalized output~\citep{zhang2019rmsnorm}, where
\(\boldsymbol r_t=
\mathrm{SiLU}(\boldsymbol x_t\mathbf W_r)\) and
\(\boldsymbol u_t=
\boldsymbol r_t\odot\mathrm{RMSNorm}(\boldsymbol o_t)\).
For multi-head GLA, the outputs of all heads are concatenated and projected back to the model dimension. Note that the output gate $\boldsymbol r_t$ is distinct from the state-retention gate $\boldsymbol\gamma_t$.

\input{figs_tex/distillation_inference}

\paragraph{Directional scanning.}
\label{sec:vision_gla}
Since the recurrence is causal along its scan order, a single scan captures context from only one direction. To model two-dimensional context, GTR cycles through four scan directions across successive blocks (see Figure~\ref{fig:gtr_overview}): left-to-right, right-to-left, top-to-bottom, and bottom-to-top. Each block performs only one directional scan and restores the original spatial order afterward. Across depth, the alternating directions allow tokens to progressively receive complementary spatial context.

\paragraph{Spatial SwiGLU.}
\label{sec:spatial_swiglu}
Directional recurrence provides long-range context, but local interactions remain dependent on the scan order. We therefore introduce Spatial SwiGLU to provide explicit local spatial mixing:
\begin{equation}
	\mathbf Y=
	\bigl[
	\mathrm{SiLU}(\mathbf X\mathbf W_g)
	\odot
	\mathrm{DWConv}_{3\times3}(\mathbf X\mathbf W_v)
	\bigr]\mathbf W_o.
	\label{eq:spatial_swiglu}
\end{equation}
The value branch is reshaped to the two-dimensional patch grid, processed by a $3\times3$ depthwise convolution, and then restored to sequence order. This provides non-causal local interaction among neighboring patches, while the gating branch remains tokenwise. \textbf{Note that GTR uses no explicit positional embeddings.}

\paragraph{Complexity discussion.}
For fixed model width and head count, GTR scales linearly with the number of tokens $L$. The recurrent update costs $\mathcal O(Ld_kd_v)$ per head, while projections, Spatial SwiGLU, and other channel operations are also linear in $L$. The recurrent state itself has constant size with respect to $L$, while memory for token activations and intermediate workspaces grows linearly with sequence length.

\subsection{Distillation}
\label{sec:distillation}

Figure~\ref{fig:distillation_pipeline} illustrates our simple final-feature distillation pipeline. We use a DINOv3 backbone~\citep{simeoni2025dinov3}, adapted for detection following EdgeCrafter~\citep{liu2026edgecrafter}, as the frozen teacher. The teacher and GTR receive the same full image, yielding spatially corresponding patch tokens. We supervise only the final patch-token representations, without masking or intermediate-layer matching.

Let $\mathbf Y_{\mathrm{tea}}\in\mathbb R^{L\times d_t}$ and
$\mathbf Y_{\mathrm{stu}}\in\mathbb R^{L\times d_s}$ denote the final
teacher and student representations at the $L$ patch positions. Since their feature
dimensions may differ, we apply a learned linear projection
$\phi:\mathbb R^{d_s}\rightarrow\mathbb R^{d_t}$ to the student features.
The distillation objective for one image is
\begin{equation}
	\mathcal L_{\mathrm{align}}=
	\frac{1}{L}\sum_{i=1}^{L}
	\left\|
	\phi(\mathbf Y_{\mathrm{stu}}^i)-\mathbf Y_{\mathrm{tea}}^i
	\right\|_2^2.
	\label{eq:loss_alignment}
\end{equation}
The loss is averaged over patch positions and excludes non-patch tokens.
Only GTR and the linear projection are optimized. This simple objective encourages the projected student features to match the teacher's final representations while
allowing GTR to learn a different internal computation. Unlike methods that
use masking, intermediate supervision, or attention alignment, our
distillation relies solely on final patch-token regression. Appendix~
\ref{sec:representation_alignment_training} provides the teacher assignments,
training images, and optimization details. The resulting GTR weights serve
as backbone initialization for subsequent detector pre-training and
task-specific training.

\subsection{Hardware-Aware Chunkwise Inference}
\label{sec:chunk_gla}

The chunkwise formulation of GLA~\citep{yang2024gla} rewrites the token-wise recurrence as efficient matrix operations within fixed-size chunks. Our implementation follows a simple three-stage procedure: we first compute chunk summaries in parallel, then propagate context across chunk boundaries through a lightweight recurrent scan, and finally compute the outputs of all chunks independently. Our inference-only operator is specialized for the GTR configuration: FP16 inputs, chunk size $C=64$, per-head dimensions $d_k=32$ and $d_v=64$, and a zero initial state. Training continues to use the reference FLA kernels~\citep{yang2024fla}. The operator takes the same queries, keys, values, and learned log-decays $\boldsymbol g_t=\log\boldsymbol\gamma_t$ as the original recurrence.

For chunk $c$ with valid length $C_c\le C$, let
$\boldsymbol A_{c,r}=\sum_{i=0}^{r}\boldsymbol g_{c,i}$ denote the cumulative log-decay. We summarize the contribution of each chunk as
\begin{equation}
	\begin{aligned}
		\widehat{\boldsymbol k}_{c,r}
		&=\boldsymbol k_{c,r}\odot
		\exp(\boldsymbol A_{c,C_c-1}-\boldsymbol A_{c,r}),&
		\mathbf U_c&=\sum_{r=0}^{C_c-1}
		\widehat{\boldsymbol k}_{c,r}^{\top}\boldsymbol v_{c,r},\\
		\boldsymbol d_c&=\exp(\boldsymbol A_{c,C_c-1}),&
		\mathbf S_{[c+1]}&=
		\mathrm{Diag}(\boldsymbol d_c)\mathbf S_{[c]}+\mathbf U_c.
	\end{aligned}
	\label{eq:chunk_boundary_main}
\end{equation}
The expensive chunk summaries $\mathbf U_c$ are computed in parallel across chunks and heads. Only the boundary states $\mathbf S_{[c]}$ are propagated sequentially, making the remaining recurrent scan lightweight. This changes only the execution schedule, not the underlying GLA recurrence.

Once the boundary states are available, all chunk outputs can again be computed independently. Defining
$\widetilde{\boldsymbol q}_{c,r}=
\boldsymbol q_{c,r}\odot\exp(\boldsymbol A_{c,r})$
and
$\widetilde{\boldsymbol k}_{c,r}=
\boldsymbol k_{c,r}\odot\exp(-\boldsymbol A_{c,r})$,
the output at position $r$ is
\(\boldsymbol o_{c,r}=d_k^{-1/2}
\left(
\widetilde{\boldsymbol q}_{c,r}\mathbf S_{[c]}
+\sum_{j=0}^{r}
\widetilde{\boldsymbol q}_{c,r}
\widetilde{\boldsymbol k}_{c,j}^{\top}
\boldsymbol v_{c,j}
\right)\).
Our output kernel fuses the within-chunk computation with the boundary-state readout. Additional implementation details, including mixed-precision accumulation, length-aware dispatch, and optional output-gate fusion, are provided in Appendix~\ref{sec:cuda_operator_impl}.

As shown in Figure~\ref{fig:chunk_gla_latency}, our operator is $2.6\times$, $4.0\times$, and $6.4\times$ faster than FLA v0.5.0~\citep{yang2024fla} at sequence lengths of 256, 1.6K, and 16.4K tokens, respectively, under FP16 and CUDA Graph execution on RTX~4090. This isolated benchmark excludes output-epilogue fusion. Full-model latency on RTX~4090 and TensorRT deployment on DRIVE AGX Thor are evaluated separately in Sections~\ref{sec:main_results} and~\ref{sec:edge_deployment}.

%% file: figs_tex/gtr_pipeline.tex
\begin{figure}[t]
    \centering
    \includegraphics[width=\textwidth]{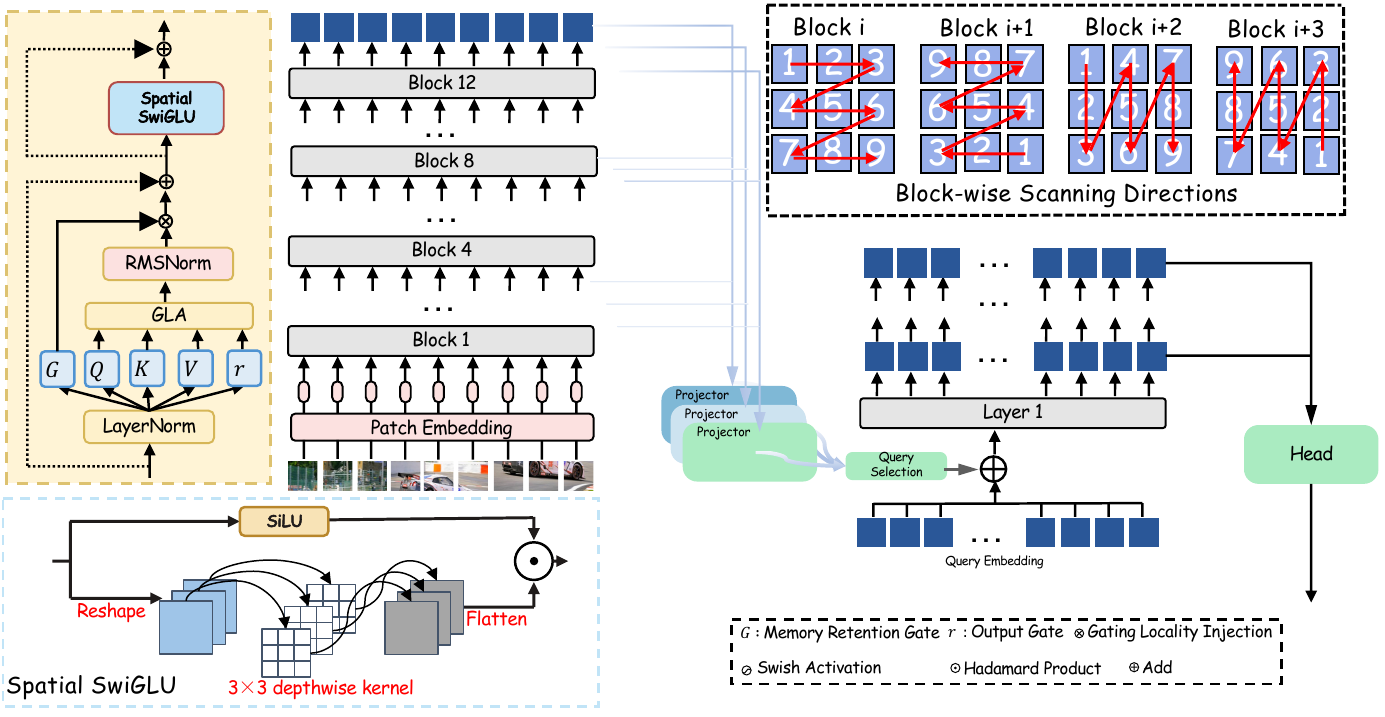}
    \caption{\textbf{Overview of GTR.}
    Each backbone block combines GLA~\citep{yang2024gla} with Spatial
    SwiGLU, whose $3\times3$ depthwise convolution mixes neighboring patch
    tokens in the value branch. The 12-block backbone alternates
    two-dimensional scan directions and exposes intermediate features to a
    lightweight three-scale projector. A query-based head decodes the
    resulting feature pyramid.}
    \label{fig:gtr_overview}
\end{figure}

%% file: figs_tex/distillation_inference.tex
\begin{figure}[t]
    \centering
    \captionsetup{font=small}
    \setlength{\figboxh}{4.6cm}
    \begin{minipage}[t]{0.49\linewidth}
        \vspace{0pt}
        \centering
        \begin{minipage}[c][\figboxh][c]{\linewidth}
            \centering
            \includegraphics[width=\linewidth,height=\figboxh,keepaspectratio]{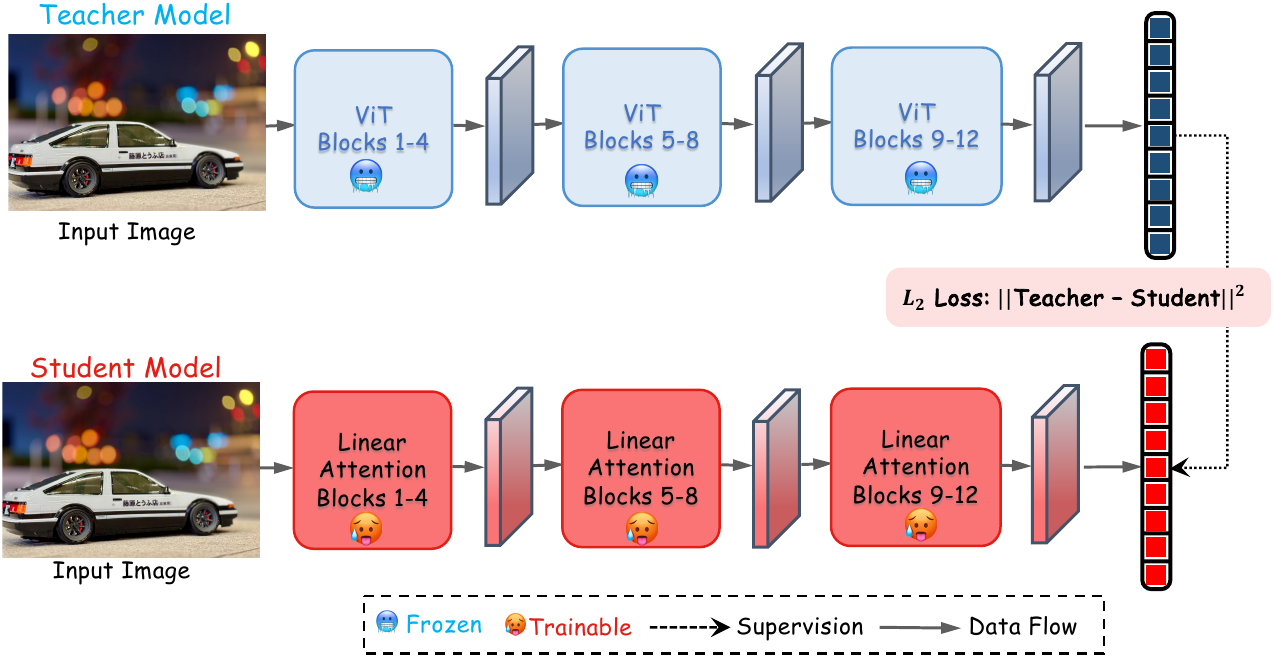}
        \end{minipage}
        \caption{\textbf{Final-output representation alignment.}
        	The frozen detection-specialized DINOv3 teacher~\citep{simeoni2025dinov3}
        	and GTR student process the same full image. We align their final patch
        	features using a squared $\ell_2$ loss after a learned linear projection
        	of the student features (Equation~\ref{eq:loss_alignment}). The projection
        	is omitted from the figure for clarity.}
        \label{fig:distillation_pipeline}
    \end{minipage}\hfill
    \begin{minipage}[t]{0.49\linewidth}
        \vspace{0pt}
        \centering
        \begin{minipage}[c][\figboxh][c]{\linewidth}
            \centering
            \includegraphics[width=\linewidth,height=\figboxh,keepaspectratio]{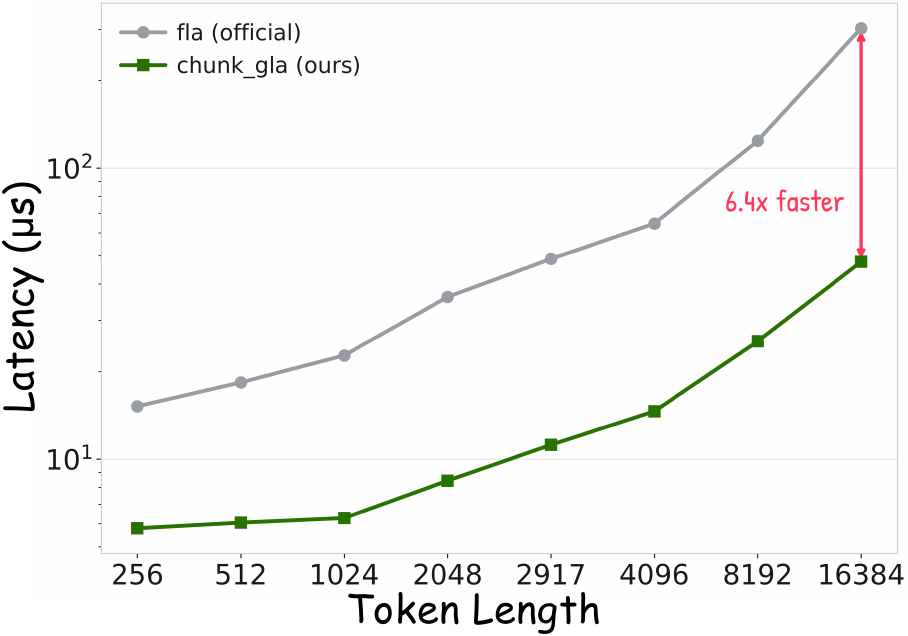}
        \end{minipage}
        \caption{\textbf{GLA operator latency on RTX~4090.}
            We compare our inference-only operator with FLA v0.5.0~\citep{yang2024fla} using FP16 inputs and CUDA Graph execution. A $640\times640$ input corresponds to 1.6K stride-16 patch tokens. Details are provided in Appendix~\ref{sec:cuda_operator_protocol}.}
        \label{fig:chunk_gla_latency}
    \end{minipage}
\end{figure}

%% file: sec/4_exp.tex
\section{Experiments}
\label{sec:experiments}

In this section, we evaluate GTR across diverse dense prediction tasks and analyze its key design choices. We first evaluate GTR on object detection, and then transfer the backbone to five additional tasks: instance segmentation, human pose estimation, oriented object detection, semantic segmentation, and monocular depth estimation. Each task uses a task-specific head, initialization, and training recipe, as detailed in Appendix~\ref{sec:experimental_setup}. The main paper focuses on recent competitive methods, with more comprehensive comparisons provided in Appendix~\ref{sec:complete_experimental_results}. We conclude with ablations of the main architectural and training choices.

\paragraph{Training details and inference latency.}
After backbone distillation, the main GTR detectors are trained for 36 epochs on Objects365~\citep{shao2019objects365}, followed by 30 epochs on COCO~\citep{lin2014microsoft} \texttt{train2017}. Direct-COCO variants skip the Objects365 stage. We evaluate on COCO \texttt{val2017} at $640\times640$ resolution. For latency, we benchmark all models on an RTX~4090 using FP16, batch size one, \texttt{torch.compile}, and CUDA Graphs, and report median CUDA-event latency after warm-up. We measure the network forward pass for DETR-style models and additionally include CUDA NMS for methods that require separate post-processing. Baseline accuracy is taken from the corresponding sources, while latency is measured under our setup.
Full training and latency protocols are provided in Appendices~\ref{sec:experimental_setup} and~\ref{sec:latency_protocol}, respectively.

\subsection{Real-Time Object Detection}
\label{sec:main_results}

\input{tables/coco_comparison_compact}
Table~\ref{tab:coco_main} compares GTR with three recent baselines: YOLO26~\citep{yolo26}, RF-DETR~\citep{robinson2025rf}, and ECDet~\citep{liu2026edgecrafter}. With Objects365 pre-training~\citep{shao2019objects365}, GTR-S/M/L/X achieve 53.6/57.3/58.9/59.4 AP, respectively, obtaining the best AP across all four scales. GTR-L reaches 58.9 AP at 1.908\,ms median latency, outperforming all L-scale baselines in both AP and latency. GTR-X achieves 59.4 AP at 2.115\,ms, compared with 58.6 AP at 3.243\,ms for RF-DETR-X. Without Objects365 pre-training, GTR consistently reduces latency relative to ECDet across all four scales. Full results are provided in Appendix~\ref{sec:coco_full_table}.

\subsection{Transfer to Other Dense-Prediction Tasks}
\label{sec:transfer_results}

We train separate task-specific models using the same GTR backbone design. Table~\ref{tab:transfer_summary} summarizes representative comparisons at the S and X scales, while Appendix~\ref{sec:complete_experimental_results} provides results across all scales and metrics.

\input{tables/transfer_compact}

\paragraph{Instance segmentation and human pose estimation.}
On COCO~\citep{lin2014microsoft}, GTR transfers effectively to both instance segmentation and human pose estimation. For instance segmentation, GTR-S/X achieve 45.0/49.8 mask AP without SAM2-generated pseudo masks~\citep{ravi2025sam}. GTR-X reaches 49.8 mask AP with 25.3\% lower latency than RF-DETR-Seg-X~\citep{robinson2025rf}. For pose estimation, GTR achieves 70.1/75.5 keypoint AP, improving over ECPose~\citep{liu2026edgecrafter} at both S and X scales while also reducing latency. Full results are provided in Appendices~\ref{sec:instance_full_table} and~\ref{sec:pose_full_table}.

\paragraph{Oriented detection and semantic segmentation.}
GTR also transfers effectively to oriented detection and semantic segmentation. On DOTA-v1.0~\citep{xia2018dota}, GTR-X achieves 81.3 AP$_{50}$ with lower median latency than YOLO26x-obb~\citep{yolo26} (3.960 vs.\ 5.085\,ms). On Cityscapes~\citep{cordts2016cityscapes}, GTR-S/X achieve 81.5/83.6 mIoU with a lightweight FCN head~\citep{long2015fully}. Compared with YOLO26-sem, GTR achieves higher mIoU at smaller scales, while GTR-X matches 83.6 mIoU with 19.5\% lower latency. Full comparisons across all scales are provided in Appendices~\ref{sec:obb_full_table} and~\ref{sec:semantic_full_table}.

\paragraph{Monocular depth estimation.}
\input{figs_tex/depth_vis}
We further evaluate GTR on monocular depth estimation using an eight-source mixture of measured and pseudo-labeled depth, without NYU-specific fine-tuning. On NYU Depth V2~\citep{silberman2012indoor,eigen2014depth}, GTR improves $\delta_1$, AbsRel, and RMSE over the evaluated YOLO26 depth variants~\citep{yolo26} at both scales. As shown in Figure~\ref{fig:depth_vis}, GTR also produces qualitatively consistent depth predictions across diverse scenes. These aligned-depth results use multi-view test-time augmentation and per-image log-affine alignment to ground-truth depth. Full results and evaluation details are provided in Appendices~\ref{sec:depth_full_table} and~\ref{sec:metric_depth_details}.

\subsection{Ablation Studies and Discussion}
\label{sec:ablation_study}

We ablate the main design choices of GTR, including the distillation objective, spatial modeling, token mixer, input resolution, and detector pre-training. All detection results are evaluated on COCO~\citep{lin2014microsoft} \texttt{val2017}. For architectural ablations, each variant is independently distilled using the same teacher and image pool, followed by the same downstream training schedule. Full settings and results are provided in Appendix~\ref{sec:ablation_details}.

\paragraph{Distillation recipe.}
Table~\ref{tab:ablation_distillation} compares our final-output alignment with ViT-Linearizer~\citep{wei2025vitlinearizer} and ViT-AdaLA~\citep{li2026vitalada} under the same teacher, GTR-S student, and COCO training schedule. For ViT-Linearizer, we use the best mask ratio (0.5) from a sweep over 0.2--0.9. Despite its simple single-stage design and objective, our method achieves 50.7 AP, outperforming ViT-Linearizer (50.0) and ViT-AdaLA (48.2), while training from scratch reaches only 31.1 AP. This suggests that simple final-output alignment is  effective for distilling GTR.

\input{tables/ablation_distillation_adapted}

\paragraph{Spatial modeling.}
Table~\ref{tab:ablation_spatial} shows that replacing learned positional embeddings with S-SwiGLU improves GTR-S by 1.2/1.3 AP at $768^2/1024^2$, with only 0.2 (47.6 to 47.8) /0.5 (83.2 to 83.7) additional GFLOPs. Scan direction provides a complementary benefit: on GTR-L, four-direction scanning achieves 55.5 AP, compared with 55.1/55.0 for horizontal/vertical bidirectional scanning and 54.4 for a single direction. Moreover, under single-direction scanning, replacing S-SwiGLU with learned positional embeddings further reduces AP to 45.6. Together, these results highlight the importance of local spatial mixing, with multi-directional scanning providing additional gains.

\input{tables/ablation_scan_direction}

\Needspace*{7\baselineskip}
\paragraph{GLA versus hybrid attention.}
Table~\ref{tab:ablation_softmax_attention} examines whether adding softmax attention benefits GTR. Replacing GLA with softmax attention in three of the twelve backbone blocks yields similar accuracy (55.4--55.6 vs.\ 55.5 AP), but increases median latency from 1.908 to 2.094--2.117\,ms and computation from 106 to 119 GFLOPs. All variants have the same 37.2M parameters and 0.19\,GB peak memory. Thus, softmax attention provides no meaningful accuracy gain while introducing additional computation and latency, supporting our homogeneous GLA design.

\input{tables/ablation_softmax_attention}

\paragraph{Resolution scaling.}
GTR-S scales effectively with input resolution (Table~\ref{tab:ablation_resolution}). Increasing the resolution from $640^2$ to $1280^2$ improves AP from 53.6 to 57.0, with a larger gain of 6.0 AP$_S$ on small objects (36.4 to 42.4). Meanwhile, median latency increases from 1.225 to 2.529\,ms and peak memory from 0.096 to 0.249\,GB. Notably, latency grows by only $2.06\times$ despite a $4\times$ increase in image tokens, demonstrating favorable scaling to high-resolution inputs.

\input{tables/ablation_resolution}

\paragraph{Edge deployment on DRIVE AGX Thor}
\label{sec:edge_deployment}
High-performance linear-attention kernels are typically designed for LLMs on server GPUs using Python/Triton stacks~\citep{qwen3.8,team2025kimi}, whereas automotive deployment requires efficient execution within a production inference engine. We deploy GTR on NVIDIA DRIVE AGX Thor with TensorRT~\citep{nvidiatensorrt} through graph--kernel co-optimization: standard TensorRT operators are combined with fused CUDA kernels for GLA and S-SwiGLU to reduce kernel launches and global-memory traffic without changing the computation. This enables low-latency inference across all six dense-prediction tasks, while maintaining at least \(0.9989\) cosine similarity to the PyTorch FP32 reference across all 48 task--scale--batch configurations. Appendix~\ref{sec:thor_deployment} provides platform details, Roofline analysis~\citep{10.1145/1498765.1498785}, and complete latency results.

%% file: tables/coco_comparison_compact.tex
\begin{table}[t]
  \centering
  \setlength{\intextsep}{4pt}

  \captionsetup{
    font=footnotesize,
    skip=2pt,
    belowskip=0pt
  }
  \caption{\textbf{COCO~\citep{lin2014microsoft} detection performance and RTX~4090 latency.}
  We compare GTR with three recent baselines; full results and COCO metrics are provided in Appendix~\ref{sec:coco_full_table}.
  GTR uses $640\times640$ inputs, while RF-DETR uses its native scale-specific resolutions.
  Latency is measured with compiled FP16 inference at batch size one (Appendix~\ref{sec:latency_protocol}).
  $^{\ddagger}$ indicates models without Objects365~\citep{shao2019objects365} pre-training.
  Per scale group, \rankfirst{best} and \ranksecond{second-best} median latency and AP are highlighted.}
  \label{tab:coco_main}

  \tiny
  \setlength{\tabcolsep}{2.15pt}
  \renewcommand{\arraystretch}{1.00}

  \begin{adjustbox}{width=\textwidth}
  \begin{tabular}{@{}lcccccccccccc@{}}
    \toprule
    \multirow{2}{*}{Model}
      & \multirow{2}{*}{\#Epochs}
      & \multirow{2}{*}{\makecell{\#Params\\(M)}}
      & \multirow{2}{*}{GFLOPs}
      & \multicolumn{3}{c}{Latency (ms) $\downarrow$}
      & \multicolumn{6}{c}{COCO AP $\uparrow$} \\    \cmidrule(lr){5-7}
    \cmidrule(lr){8-13}
      & & & & Min. & Mean & Median
      & AP$^{val}$ & AP$^{val}_{50}$ & AP$^{val}_{75}$
      & AP$^{val}_{S}$ & AP$^{val}_{M}$ & AP$^{val}_{L}$ \\    \midrule

    RF-DETR-S~\citep{robinson2025rf} & -- & 32 & 60
      & 1.207 & 1.285 & 1.299
      & \ranksecond{52.9} & \rankfirst{71.9} & \ranksecond{57.0}
      & 32.0 & \rankfirst{58.3} & \rankfirst{73.0} \\

    YOLO26-S~\citep{yolo26} & 70 & 10 & 21
      & 0.922 & 0.977 & \rankfirst{0.962}
      & 47.8 & 64.6 & 52.1 & 29.1 & 52.5 & 64.3 \\

    ECDet-S$^{\ddagger}$~\citep{liu2026edgecrafter} & 74 & 10 & 26
      & 1.511 & 1.641 & 1.652
      & 51.7 & 69.4 & 55.8
      & \ranksecond{32.3} & \ranksecond{56.4} & \ranksecond{70.5} \\

    \textbf{GTR-S (ours)}$^{\ddagger}$ & 30 & 12.1 & 33.8
      & 1.215 & 1.225 & \ranksecond{1.225}
      & 50.7 & 68.5 & 54.7 & 31.1 & 55.5 & 70.2 \\

    \textbf{GTR-S (ours)} & 30 & 12.1 & 33.8
      & 1.215 & 1.225 & \ranksecond{1.225}
      & \rankfirst{53.6} & \ranksecond{71.1} & \rankfirst{58.3}
      & \rankfirst{36.4} & \rankfirst{58.3} & 70.2 \\

    \midrule

    RF-DETR-M~\citep{robinson2025rf} & -- & 34 & 79
      & 1.326 & 1.428 & \rankfirst{1.431}
      & \ranksecond{54.7} & \ranksecond{73.5} & \ranksecond{59.2}
      & 36.1 & \ranksecond{59.7} & \ranksecond{73.8} \\

    YOLO26-M~\citep{yolo26} & 80 & 20 & 68
      & 1.459 & 1.481 & 1.473
      & 52.5 & 69.8 & 57.2 & \ranksecond{36.2} & 56.9 & 68.5 \\

    ECDet-M$^{\ddagger}$~\citep{liu2026edgecrafter} & 62 & 19 & 53
      & 1.901 & 2.080 & 2.095
      & 54.3 & 72.2 & 58.7 & 35.9 & 59.1 & 72.7 \\

    \textbf{GTR-M (ours)}$^{\ddagger}$ & 30 & 22.7 & 62.6
      & 1.450 & 1.462 & \ranksecond{1.462}
      & 54.0 & 72.2 & 58.5 & 35.3 & 59.1 & 73.1 \\

    \textbf{GTR-M (ours)} & 30 & 22.7 & 62.6
      & 1.450 & 1.462 & \ranksecond{1.462}
      & \rankfirst{57.3} & \rankfirst{75.0} & \rankfirst{62.4}
      & \rankfirst{41.1} & \rankfirst{62.0} & \rankfirst{74.1} \\

    \midrule

    RF-DETR-L~\citep{robinson2025rf} & -- & 34 & 126
      & 1.805 & 1.952 & 1.968
      & 56.5 & \ranksecond{75.1} & 61.3
      & \ranksecond{39.0} & 61.0 & 73.9 \\

    YOLO26-L~\citep{yolo26} & 60 & 25 & 86
      & 1.953 & 1.969 & \ranksecond{1.967}
      & 54.3 & 71.5 & 59.4 & 37.8 & 58.6 & 70.3 \\

    ECDet-L$^{\ddagger}$~\citep{liu2026edgecrafter} & 50 & 32 & 101
      & 2.493 & 2.713 & 2.730
      & \ranksecond{57.0} & \ranksecond{75.1} & \ranksecond{61.7}
      & 38.7 & \ranksecond{62.5} & \ranksecond{75.0} \\

    \textbf{GTR-L (ours)}$^{\ddagger}$ & 30 & 37.2 & 106
      & 1.902 & 1.908 & \rankfirst{1.908}
      & 55.5 & 73.7 & 60.3 & 36.8 & 61.1 & 74.8 \\

    \textbf{GTR-L (ours)} & 30 & 37.2 & 106
      & 1.902 & 1.908 & \rankfirst{1.908}
      & \rankfirst{58.9} & \rankfirst{76.8} & \rankfirst{64.3}
      & \rankfirst{42.4} & \rankfirst{64.0} & \rankfirst{76.0} \\

    \midrule

    RF-DETR-X~\citep{robinson2025rf} & -- & 126 & 300
      & 2.971 & 3.227 & 3.243
      & \ranksecond{58.6} & \rankfirst{77.4} & \ranksecond{63.8}
      & 40.3 & \ranksecond{63.9} & \ranksecond{76.2} \\

    YOLO26-X~\citep{yolo26} & 40 & 55 & 194
      & 3.069 & 3.088 & 3.087
      & 56.9 & 74.1 & 62.1 & \ranksecond{41.3} & 61.2 & 72.7 \\

    ECDet-X$^{\ddagger}$~\citep{liu2026edgecrafter} & 50 & 49 & 151
      & 2.950 & 3.025 & \ranksecond{3.021}
      & 57.9 & 76.0 & 62.9 & 38.7
      & 63.4 & 76.1 \\

    \textbf{GTR-X (ours)}$^{\ddagger}$ & 30 & 46.5 & 130.2
      & 2.104 & 2.117 & \rankfirst{2.115}
      & 56.2 & 74.5 & 61.1 & 37.3 & 61.8 & 74.9 \\

    \textbf{GTR-X (ours)} & 30 & 46.5 & 130.2
      & 2.104 & 2.117 & \rankfirst{2.115}
      & \rankfirst{59.4} & \ranksecond{77.3} & \rankfirst{64.7}
      & \rankfirst{42.1} & \rankfirst{64.6} & \rankfirst{76.4} \\

    \bottomrule
  \end{tabular}
  \end{adjustbox}
\end{table}

%% file: tables/transfer_compact.tex
\begin{table}[t]
  \centering
  \setlength{\intextsep}{4pt}
  \captionsetup{font=footnotesize, skip=2pt, belowskip=0pt}
  \caption{\textbf{Transfer of the GTR backbone to five additional dense prediction tasks at the S and X scales.}
  Each task uses a separately trained model with a task-specific head (Appendix~\ref{sec:experimental_setup}).
  Params in M, FLOPs in GFLOPs, Lat.\ is the median RTX~4090 latency (ms) measured as in Table~\ref{tab:coco_main}; complete comparisons are provided in Appendix~\ref{sec:complete_experimental_results}.
  $^{\dagger}$ denotes Objects365~\citep{shao2019objects365} pre-training with SAM2 pseudo masks~\citep{ravi2025sam} for instance segmentation and additional Objects365 supervision for pose estimation.
  Per task and scale, the \rankfirst{best} latency and accuracy are highlighted.}
  \label{tab:transfer_summary}

  \tiny
  \setlength{\tabcolsep}{3pt}
  \renewcommand{\arraystretch}{1.02}

  \begin{adjustbox}{width=\textwidth}
  \begin{tabular}{@{}llcccccccc@{}}
    \toprule
    \multirow{2}{*}{Task} & \multirow{2}{*}{Model}
      & \multicolumn{4}{c}{S} & \multicolumn{4}{c}{X} \\
    \cmidrule(lr){3-6}\cmidrule(lr){7-10}
      & & Params & FLOPs & Lat.\ $\downarrow$ & Metric
        & Params & FLOPs & Lat.\ $\downarrow$ & Metric \\
    \midrule

    \multirow{4}{*}{\makecell[l]{Instance Seg.\\COCO mask AP $\uparrow$}}
      & YOLO26-Seg$^{\dagger}$~\citep{yolo26}
      & 10.4 & 34.2 & \rankfirst{1.163} & 40.0
      & 62.8 & 313.5 & 4.139 & 47.0 \\
      & RF-DETR-Seg$^{\dagger}$~\citep{robinson2025rf}
      & 33.7 & 70.6 & 1.445 & 43.1
      & 38.1 & 260.0 & 3.311 & 48.8 \\
      & ECInsSeg~\citep{liu2026edgecrafter}
      & 10.3 & 33.1 & 1.755 & 43.0
      & 49.9 & 168.1 & 3.140 & 48.4 \\
      & \textbf{GTR (ours)}
      & 12.6 & 46.8 & 1.465 & \rankfirst{45.0}
      & 47.4 & 151.6 & \rankfirst{2.472} & \rankfirst{49.8} \\
    \midrule

    \multirow{4}{*}{\makecell[l]{Pose\\COCO keypoint AP $\uparrow$}}
      & YOLO26-Pose~\citep{yolo26}
      & 10.4 & 23.9 & \rankfirst{0.813} & 63.1
      & 57.6 & 201.7 & 2.788 & 71.6 \\
      & DETRPose$^{\dagger}$~\citep{janampa2025detrpose}
      & 11.5 & 33.1 & 1.218 & 67.0
      & 73.3 & 239.5 & 4.145 & 73.3 \\
      & ECPose~\citep{liu2026edgecrafter}
      & 9.9 & 30.4 & 1.519 & 68.9
      & 50.6 & 172.2 & 3.281 & 74.8 \\
      & \textbf{GTR (ours)}
      & 11.9 & 37.0 & 1.455 & \rankfirst{70.1}
      & 47.6 & 142.9 & \rankfirst{2.590} & \rankfirst{75.5} \\
    \midrule

    \multirow{2}{*}{\makecell[l]{Oriented Det.\\DOTA AP$_{50}$ $\uparrow$}}
      & YOLO26-obb~\citep{yolo26}
      & 9.8 & 56.7 & \rankfirst{1.099} & \rankfirst{80.9}
      & 57.6 & 520.1 & 5.085 & \rankfirst{81.7} \\
      & \textbf{GTR (ours)}
      & 12.1 & 82.8 & 1.946 & 80.1
      & 46.3 & 324 & \rankfirst{3.960} & 81.3 \\
    \midrule

    \multirow{2}{*}{\makecell[l]{Semantic Seg.\\Cityscapes mIoU $\uparrow$}}
      & YOLO26-sem~\citep{yolo26}
      & 6.5 & 88.8 & \rankfirst{0.758} & 80.8
      & 40.2 & 861.7 & 4.325 & \rankfirst{83.6} \\
      & \textbf{GTR (ours)}
      & 6.9 & 82.8 & 1.495 & \rankfirst{81.5}
      & 32.2 & 317.1 & \rankfirst{3.482} & \rankfirst{83.6} \\
    \midrule

    \multirow{2}{*}{\makecell[l]{Depth\\NYU $\delta_1$ $\uparrow$}}
      & YOLO26-depth~\citep{yolo26}
      & 13.2 & 67.9 & \rankfirst{0.850} & 0.896
      & 57.0 & 302.0 & 2.730 & 0.933 \\
      & \textbf{GTR (ours)}
      & 11.5 & 65.0 & 1.296 & \rankfirst{0.946}
      & 41.1 & 159.4 & \rankfirst{2.158} & \rankfirst{0.954} \\
    \bottomrule
  \end{tabular}
  \end{adjustbox}
\end{table}

%% file: figs_tex/depth_vis.tex
\begin{figure}[t]
    \centering
    \includegraphics[width=\textwidth,pagebox=cropbox]{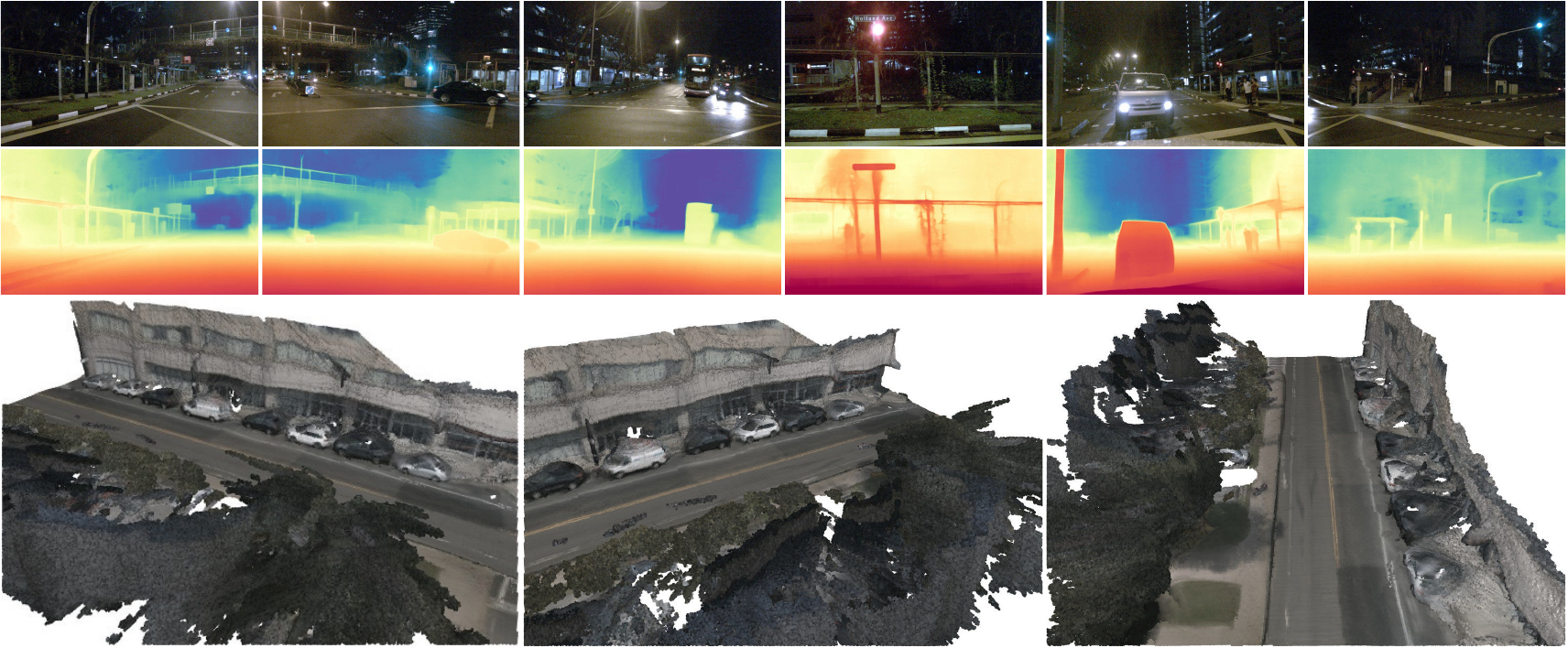}
    \caption{\textbf{GTR-L depth prediction and reconstruction on nuScenes~\citep{caesar2020nuscenes}, without dataset-specific fine-tuning.}
    	\textit{Top/middle:} Six nighttime surround-view images and the corresponding GTR-L depth predictions.
    	\textit{Bottom:} Multi-view visualization of a reconstructed street scene, obtained by temporally fusing predicted depth after LiDAR-assisted per-frame scale alignment (Appendix~\ref{sec:depth_nuscenes_visualization}).}
    \label{fig:depth_vis}
\end{figure}

%% file: tables/ablation_distillation_adapted.tex
\begin{table}[t]
  \centering
  \setlength{\intextsep}{4pt}

  \captionsetup{
    font=footnotesize,
    skip=2pt,
    belowskip=0pt
  }
  \caption{\textbf{Cross-architecture distillation on GTR-S without Objects365 pre-training.}
  	We include a from-scratch baseline for reference. Methods use the same teacher, student, and COCO training schedule. }

  \label{tab:ablation_distillation}
  \footnotesize
  \setlength{\tabcolsep}{2.5pt}
  \renewcommand{\arraystretch}{1.06}
  \begin{adjustbox}{max width=\textwidth}
  \begin{tabular}{@{}lcccccccc@{}}
    \toprule
    \multirow{2}{*}{Initialization / recipe}
      & \multirow{2}{*}{\makecell{Distill.\\stages}}
      & \multirow{2}{*}{\makecell{\#Distill\\losses}}
      & \multicolumn{6}{c}{COCO AP $\uparrow$} \\
    \cmidrule(lr){4-9}
      & &
      & AP$^{val}$ & AP$^{val}_{50}$ & AP$^{val}_{75}$
      & AP$^{val}_{S}$ & AP$^{val}_{M}$ & AP$^{val}_{L}$ \\
    \midrule
    Random initialization (from scratch)
      & 0 & 0
      & 31.1 & 44.9 & 33.3 & 17.0 & 33.1 & 43.7 \\
    \midrule
    ViT-AdaLA (adapted)~\citep{li2026vitalada}
      & 2 & 1
      & 48.2 & 65.6 & 52.1 & 28.2 & 52.4 & 66.9 \\
    ViT-Linearizer (adapted)~\citep{wei2025vitlinearizer}
      & 1 & 2
      & 50.0 & 67.7 & 53.6 & \textbf{31.1} & 54.7 & 69.4 \\
    \rowcolor{gtrblue!10}
    \textbf{Final-output $\ell_2$ (ours)}
      & \textbf{1} & \textbf{1}
      & \textbf{50.7} & \textbf{68.5} & \textbf{54.7}
      & \textbf{31.1} & \textbf{55.5} & \textbf{70.2} \\
    \bottomrule
  \end{tabular}
  \end{adjustbox}
\end{table}

%% file: tables/ablation_scan_direction.tex
\begin{table}[t]
	\centering
	\setlength{\intextsep}{4pt}

	\captionsetup{
		font=footnotesize,
		skip=2pt,
		belowskip=0pt
	}

	\caption{\textbf{Ablation of spatial modeling in GTR.}
		We study local spatial mixing with S-SwiGLU on GTR-S and scan directions on GTR-L.
		LR/RL/TB/BT denote left-to-right, right-to-left, top-to-bottom, and bottom-to-top scans.}
	\label{tab:ablation_spatial}

	\footnotesize
	\setlength{\tabcolsep}{4.0pt}
	\renewcommand{\arraystretch}{1.12}

	\begin{tabular}{@{}llcccccc@{}}
		\toprule
		\multirow{2}{*}{Setting}
		& \multirow{2}{*}{Variant}
		& \multicolumn{6}{c}{COCO AP $\uparrow$} \\
		\cmidrule(lr){3-8}
		&
		& AP$^{val}$ & AP$^{val}_{50}$ & AP$^{val}_{75}$
		& AP$^{val}_{S}$ & AP$^{val}_{M}$ & AP$^{val}_{L}$ \\
		\midrule

		\multicolumn{8}{l}{\textit{Local spatial mixing (GTR-S)}} \\

		$768^2$
		& Learned positional embedding
		& 53.9 & 71.6 & 58.6 & 38.0 & 57.8 & 70.2 \\

		\rowcolor{gtrblue!10}
		$768^2$
		& \textbf{S-SwiGLU (ours)}
		& \textbf{55.1} & \textbf{72.5} & \textbf{60.2}
		& \textbf{38.7} & \textbf{59.6} & \textbf{70.7} \\

		$1024^2$
		& Learned positional embedding
		& 55.3 & 72.7 & 60.3 & 40.6 & 58.7 & \textbf{70.7} \\

		\rowcolor{gtrblue!10}
		$1024^2$
		& \textbf{S-SwiGLU (ours)}
		& \textbf{56.6} & \textbf{73.9} & \textbf{61.9}
		& \textbf{42.1} & \textbf{60.8} & 70.4 \\

		\midrule

		\multicolumn{8}{l}{\textit{Scan direction (GTR-L)}} \\

		Horizontal
		& LR/RL + S-SwiGLU
		& 55.1 & 73.4 & 59.8 & 36.1 & 60.5 & \textbf{75.2} \\

		Vertical
		& TB/BT + S-SwiGLU
		& 55.0 & 73.3 & 59.8 & 35.8 & 60.7 & 74.6 \\

		Single direction
		& S-SwiGLU
		& 54.4 & 72.6 & 59.0 & 35.1 & 59.7 & 73.9 \\

		Single direction
		& Learned positional embedding
		& 45.6 & 62.3 & 49.1 & 27.5 & 49.1 & 63.5 \\

		\rowcolor{gtrblue!10}
		Four directions
		& \textbf{LR/RL/TB/BT + S-SwiGLU}
		& \textbf{55.5} & \textbf{73.7} & \textbf{60.3}
		& \textbf{36.8} & \textbf{61.1} & 74.8 \\
		\bottomrule
	\end{tabular}
\end{table}

%% file: tables/ablation_softmax_attention.tex
\begin{table}[t]
	\centering
	\setlength{\intextsep}{4pt}

	\captionsetup{
		font=footnotesize,
		skip=2pt,
		belowskip=0pt
	}

	\caption{\textbf{GLA versus hybrid GLA--softmax attention on GTR-L.}
		Each hybrid replaces GLA with softmax attention in three of the twelve backbone blocks.}
	\label{tab:ablation_softmax_attention}

	\footnotesize
	\setlength{\tabcolsep}{3.2pt}
	\renewcommand{\arraystretch}{1.10}

	\begin{tabular}{@{}lccccccc@{}}
		\toprule
		\multirow{2}{*}{Global-mixer layout}
		& \multirow{2}{*}{\makecell{Latency\\(ms) $\downarrow$}}
		& \multicolumn{6}{c}{COCO AP $\uparrow$} \\
		\cmidrule(lr){3-8}
		&
		& AP$^{val}$ & AP$^{val}_{50}$ & AP$^{val}_{75}$
		& AP$^{val}_{S}$ & AP$^{val}_{M}$ & AP$^{val}_{L}$ \\
		\midrule

		\rowcolor{gtrblue!10}
		\textbf{Pure GLA (ours)}
		& \textbf{1.908}
		& 55.5 & 73.7 & 60.3
		& 36.8 & 61.1 & 74.8 \\

		Softmax in last 3 blocks
		& 2.117
		& \textbf{55.6} & 73.8 & \textbf{60.7}
		& 37.2 & \textbf{61.7} & 74.6 \\

		Softmax in 3 uniformly spaced blocks
		& 2.109
		& \textbf{55.6} & \textbf{74.0} & 60.0
		& \textbf{37.3} & 60.9 & \textbf{75.1} \\

		Softmax in first 3 blocks
		& 2.094
		& 55.4 & 73.6 & 60.1
		& 35.5 & 61.0 & 74.6 \\

		\bottomrule
	\end{tabular}
\end{table}

%% file: tables/ablation_resolution.tex
\begin{table}[t]
	\centering
	\setlength{\intextsep}{4pt}

	\captionsetup{
		font=footnotesize,
		skip=2pt,
		belowskip=0pt
	}

	\caption{\textbf{Resolution scaling of GTR-S.}
		Higher resolution consistently improves accuracy, particularly for small objects, at increased computational cost.}
	\label{tab:ablation_resolution}

	\footnotesize
	\setlength{\tabcolsep}{4.0pt}
	\renewcommand{\arraystretch}{1.08}

	\begin{tabular}{@{}cccccccc@{}}
		\toprule
		Resolution
		& GFLOPs
		& Memory (GB)
		& Latency (ms) $\downarrow$
		& AP $\uparrow$
		& AP$_S$ $\uparrow$
		& AP$_M$ $\uparrow$
		& AP$_L$ $\uparrow$ \\
		\midrule

		$640^2$  & 33.8  & 0.096 & 1.225 & 53.6 & 36.4 & 58.3 & 70.2 \\
		$704^2$  & 40.4  & 0.121 & 1.317 & 54.3 & 37.5 & 59.0 & 70.2 \\
		$768^2$  & 47.8  & 0.138 & 1.421 & 55.1 & 38.7 & 59.6 & 70.7 \\
		$832^2$  & 55.8  & 0.142 & 1.550 & 55.5 & 39.8 & 60.0 & \textbf{70.8} \\
		$896^2$  & 64.5  & 0.156 & 1.714 & 56.0 & 41.4 & 60.0 & 70.5 \\
		$960^2$  & 73.7  & 0.172 & 1.784 & 56.2 & 40.7 & 60.4 & 70.0 \\
		$1024^2$  & 83.7  & 0.171 & 1.844 & 56.6 & 42.1 & 60.8 & 70.4 \\
		$1152^2$  & 105.4 & 0.224 & 2.157 & 56.7 & 41.9 & 60.8 & 70.3 \\

		\rowcolor{gtrblue!10}
		$1280^2$  & 129.8 & 0.249 & 2.529
		& \textbf{57.0}
		& \textbf{42.4}
		& \textbf{61.0}
		& 70.0 \\

		\bottomrule
	\end{tabular}
\end{table}

%% file: sec/5_conclusion.tex
\section{Conclusion}
\label{sec:conclusion}
We presented GTR, an efficient softmax-free backbone for dense prediction. GTR combines a homogeneous GLA architecture with simple final-output distillation and optimized inference execution. Across object detection and five additional dense-prediction tasks, GTR achieves a strong accuracy--efficiency trade-off, while deployment on NVIDIA DRIVE AGX Thor demonstrates its practicality on automotive hardware. These results show that linear attention can serve as an effective and efficient alternative to softmax attention for dense visual prediction.
\pagebreak

\subsection*{AI use statement}

Generative AI tools were used to assist with literature review, manuscript organization, language editing, interpretation and synthesis of the reported experimental results, and limited code development and debugging. Their use in result interpretation was limited to summarizing and describing trends and comparisons already present in the figures and tables. Generative AI tools were not used to generate, modify, or fabricate experimental data. All AI-assisted text and code were reviewed and verified by the authors, who take full responsibility for the final manuscript.

%% file: sec/X_suppl.tex
\clearpage
\appendix
\renewcommand{\thepage}{A\arabic{page}}
\setcounter{page}{0}
\thispagestyle{empty}
\makeatletter
\setlength{\@fptop}{0pt}
\setlength{\@dblfptop}{0pt}
\makeatother
\renewcommand{\topfraction}{0.95}
\renewcommand{\textfraction}{0.05}
\setcounter{topnumber}{4}
\raggedbottom

\begin{center}
    {\large\sffamily\bfseries
    GTR: Gated Token Recurrence for Efficient Dense Prediction}\\[12pt]
    {\LARGE Appendix}
\end{center}
\vspace{20pt}

\newcommand{\apptocsection}[2]{\par\noindent
  \begingroup\large\sffamily\bfseries
  \hyperref[#1]{\textcolor{gtrtocred}{\makebox[2.2em][l]{\ref*{#1}}#2}}\nobreak\leaders\hbox to 0.75em{\hss.\hss}\hfill
  \nobreak\hyperref[#1]{\textcolor{black}{\pageref*{#1}}}\par
  \endgroup\vspace{7pt}}
\newcommand{\apptocsubsection}[2]{\par\noindent\hspace*{2.2em}
  \begingroup\normalsize
  \hyperref[#1]{\textcolor{gtrtocred}{\makebox[3.2em][l]{\ref*{#1}}#2}}\nobreak\leaders\hbox to 0.75em{\hss.\hss}\hfill
  \nobreak\hyperref[#1]{\textcolor{black}{\pageref*{#1}}}\par
  \endgroup\vspace{3pt}}

\apptocsection{sec:related_work}{Related Work}
\apptocsection{sec:encoder_decoder_structure}{GTR Architecture Details}
\apptocsubsection{sec:encoder}{Multi-scale Feature Encoder}
\apptocsubsection{sec:decoder}{Query-based Detection Decoder}
\apptocsection{sec:complete_experimental_results}{Complete Experimental Results}
\apptocsubsection{sec:coco_full_table}{Full COCO Object Detection Comparison}
\apptocsubsection{sec:instance_full_table}{Full COCO Instance Segmentation Comparison}
\apptocsubsection{sec:pose_full_table}{Full COCO Human Pose Estimation Comparison}
\apptocsubsection{sec:obb_full_table}{Full DOTA-v1.0 Oriented Object Detection Comparison}
\apptocsubsection{sec:semantic_full_table}{Full Cityscapes Semantic Segmentation Comparison}
\apptocsubsection{sec:depth_full_table}{Full NYU Depth Estimation Comparison}
\apptocsubsection{sec:ablation_details}{Detailed Ablation Results}
\apptocsection{sec:experimental_setup}{Experimental Details}
\apptocsubsection{sec:detailed_training}{Training Configurations}
\apptocsection{sec:efficiency_details}{Efficiency and Reproducibility Details}

\apptocsubsection{sec:thor_deployment}{DRIVE AGX Thor Deployment}
\apptocsubsection{sec:latency_protocol}{GPU Latency Measurement Protocol}
\apptocsubsection{sec:cuda_operator_protocol}{Custom CUDA Operator}
\apptocsection{sec:qualitative_results}{Qualitative Results}
\apptocsubsection{sec:detection_visualization}{Object Detection Results}
\apptocsubsection{sec:segmentation_visualization}{Instance Segmentation Results}
\apptocsubsection{sec:pose_visualization}{Human Pose Estimation Results}
\apptocsubsection{sec:semantic_segmentation_visualization}{Semantic Segmentation Results}
\apptocsubsection{sec:depth_visualization}{Zero-Shot Monocular Depth Estimation Results}
\apptocsubsection{sec:obb_visualization}{Oriented Object Detection Results}

\clearpage

\input{sec/2_related}

\Needspace{5\baselineskip}
\section{GTR Architecture Details}
\label{sec:encoder_decoder_structure}
\subsection{Multi-scale Feature Encoder}
\label{sec:encoder}

Given an input image $\mathbf{I}\in\mathbb{R}^{H\times W\times 3}$, the GLA
backbone extracts features from three depths:
\begingroup
\normalsize
\begin{equation}
\{\mathbf{X}_4,\mathbf{X}_8,\mathbf{X}_{12}\}
=
\mathrm{GTRBackbone}(\mathbf{I}),
\label{eq:backbone_outputs}
\end{equation}
\endgroup
where $\mathbf{X}_4,\mathbf{X}_8,\mathbf{X}_{12}\in
\mathbb{R}^{\frac{H}{16}\times\frac{W}{16}\times C}$ are the restored and
reshaped feature maps from the 4th, 8th, and 12th backbone blocks.
For each target stride $s\in\{8,16,32\}$, we first align all selected features to the same spatial resolution:
\begingroup
\normalsize
\begin{equation}
\widetilde{\mathbf{X}}_{l}^{(s)}
=
\mathcal{R}_{s}(\mathbf{X}_l),
\quad
l\in\{4,8,12\},
\label{eq:resample_features}
\end{equation}
\endgroup
where $\mathcal{R}_{s}(\cdot)$ resizes each feature map to the target stride
$s\in\{8,16,32\}$ using bilinear interpolation; no resampling is applied when the source and
target strides are identical.
For each target scale, the aligned features from different semantic depths are concatenated along the channel dimension and fused by a lightweight C2f module~\citep{yolov8}:
\begingroup
\normalsize
\begin{equation}
\mathbf{P}^{(s)}
=
\mathrm{C2f}_{s}
\left(
\mathrm{Concat}
[
\widetilde{\mathbf{X}}_{4}^{(s)},
\widetilde{\mathbf{X}}_{8}^{(s)},
\widetilde{\mathbf{X}}_{12}^{(s)}
]
\right),
\quad
s\in\{8,16,32\}.
\label{eq:c2f_fusion}
\end{equation}
\endgroup
This produces a three-level feature pyramid
$\{\mathbf{P}^{(8)},\mathbf{P}^{(16)},\mathbf{P}^{(32)}\}$ with unified
channel dimensions.
Compared with heavier feature-pyramid encoders, this design keeps the encoder compact while aggregating early, middle, and late backbone representations at every detection scale.
The three pyramid levels are then flattened and concatenated as the
multi-scale memory for the decoder.

\subsection{Query-based Detection Decoder}
\label{sec:decoder}

The decoder follows the DETR-style set prediction
paradigm~\citep{carion2020end,peng2025dfine}.
Given the multi-scale features $\{\mathbf{P}^{(8)},\mathbf{P}^{(16)},\mathbf{P}^{(32)}\}$, we flatten and concatenate them into memory tokens:
\begingroup
\normalsize
\begin{equation}
\mathbf{M}
=
\mathrm{Concat}
[
\mathrm{Flatten}(\mathbf{P}^{(8)}),
\mathrm{Flatten}(\mathbf{P}^{(16)}),
\mathrm{Flatten}(\mathbf{P}^{(32)})
].
\label{eq:memory_tokens}
\end{equation}
\endgroup
Initial anchor proposals are generated from the multi-scale memory, and the top-$k$ candidates with the highest classification responses are selected to initialize object queries.

During training, we adopt grouped queries~\citep{chen2023group} by dividing detection queries into $G=3$ groups with shared decoder parameters. This provides denser supervision and improves query diversity, while only the primary group is used during inference.

\section{Complete Experimental Results}
\label{sec:complete_experimental_results}

The main paper uses representative detector families or model scales to keep
the quantitative presentation concise. This section reports the complete
comparisons used in our analysis for COCO object detection, instance segmentation,
and human pose estimation, DOTA-v1.0 oriented object detection, Cityscapes semantic
segmentation, and ground-truth-aligned monocular depth on NYU Depth V2.

\subsection{Full COCO Object Detection Comparison}
\label{sec:coco_full_table}

\paragraph{Accuracy--latency comparison.}
Table~\ref{tab:coco_main_full} expands the main-paper comparison with a
broader set of CNN- and Transformer-based detectors, model complexity, all
three latency statistics, and scale-specific AP. With Objects365
pre-training, GTR-S/M/L/X attain the highest overall AP in all four scale
groups, reaching 53.6/57.3/58.9/59.4 AP. GTR-L and GTR-X also provide the
lowest median latency in their groups. Without Objects365 pre-training, GTR-S/M/L/X achieve
50.7/54.0/55.5/56.2 AP under the same scale-specific COCO configurations.

\input{tables/coco_main_table_layout}
\FloatBarrier

\subsection{Full COCO Instance Segmentation Comparison}
\label{sec:instance_full_table}
Table~\ref{tab:coco_instance_segmentation} gives the full mask-metric breakdown. GTR has the highest overall mask AP in each reported scale group. GTR-X improves both mask AP and latency over RF-DETR-Seg-X. GTR attains these results without the SAM2 pseudo-mask supervision used by the $\dagger$-marked baselines.
\input{tables/coco_instance_segmentation_layout}
\FloatBarrier

\subsection{Full COCO Human Pose Estimation Comparison}
\label{sec:pose_full_table}

Table~\ref{tab:coco_pose_full} complements the representative main-paper
comparison with RTMO, YOLO11-Pose, and ED-Pose where the corresponding model
scales are available. Across this broader comparison, GTR attains the highest
keypoint AP and AR at every scale.

\input{tables/coco_pose}
\FloatBarrier

\subsection{Full DOTA-v1.0 Oriented Object Detection Comparison}
\label{sec:obb_full_table}

Table~\ref{tab:dota_obb_full} expands the representative main-paper
comparison with the smaller YOLO26-obb and RTMDet-R variants. Compared with RTMDet-R-l, GTR-X gains 0.8 AP$_{50}$ with 6.0M fewer
parameters. Relative to the single-scale
Oriented-DETR (Swin-T)~\citep{zhao2024oriented}, GTR-X gains 1.5 AP$_{50}$ and
is $3.9\times$ faster. In the class-wise breakdown, GTR-X attains the highest
plane and harbor AP$_{50}$ among all detectors with reported per-class results
(89.5 and 84.5).

\input{tables/dota_obb_full}
\FloatBarrier

\subsection{Full Cityscapes Semantic Segmentation Comparison}
\label{sec:semantic_full_table}

\input{tables/cityscapes_semantic}
\FloatBarrier

Table~\ref{tab:cityscapes_semantic_full} reports all four evaluated model
scales. GTR-S/M/L outperform their size-matched YOLO26-sem counterparts by
0.7/1.0/0.3 mIoU points, respectively. GTR-X matches YOLO26x-sem at 83.6 mIoU
while using 20.0\% fewer parameters, 63.2\% fewer FLOPs, and 19.5\% lower
median latency.

\subsection{Full NYU Depth Estimation Comparison}
\label{sec:depth_full_table}

\input{tables/nyu_depth_aligned}
\FloatBarrier

Table~\ref{tab:nyu_depth_full} reports all four evaluated YOLO26/GTR scales
and ZipDepth~\citep{tosi2026zipdepth}. At
every scale, GTR improves $\delta_1$, AbsRel, and RMSE over the corresponding
YOLO26 depth variant while using fewer GFLOPs. GTR-X additionally reduces
median latency from 2.730 to 2.158~ms.

ZipDepth is re-evaluated on the same filled ground-truth data as YOLO26 and
GTR, yielding $\delta_1=0.919$, AbsRel $=0.091$, and RMSE $=0.385$.
The original ZipDepth paper reports $\delta_1=0.933$ and AbsRel $=0.084$
under its own evaluation protocol.

\subsection{Detailed Ablation Results}
\label{sec:ablation_details}

\paragraph{Spatial mechanism.}
Table~\ref{tab:ablation_positional_encoding} compares S-SwiGLU with learned positional embeddings. S-SwiGLU improves overall AP at both resolutions with a small FLOP increase.

\input{tables/ablation_positional_encoding}

\paragraph{Scan schedule.}
Table~\ref{tab:ablation_scan_direction} reports GTR-L results for all evaluated scan schedules. Four directions yield the highest overall AP. The single-direction comparison shows a larger change when removing S-SwiGLU than when reducing the number of scan directions.

\input{tables/ablation_scan_direction_full}

\paragraph{Hybrid attention.}
Table~\ref{tab:ablation_softmax_attention} replaces GLA with softmax attention in three backbone blocks at early, uniform, or late positions. The hybrids use FlashAttention-2~\citep{dao2023flashattention2}. They increase GFLOPs from 106 to 119 and median latency by 9.7--11.0\%, while overall AP remains within 0.1 of pure GLA. The reported memory values coincide at the displayed precision.

\paragraph{Resolution scaling.}
Table~\ref{tab:ablation_resolution} reports the full GTR-S sweep with Objects365-pre-trained detector initialization~\citep{shao2019objects365}. Most of the accuracy gain occurs for small and medium objects. The batch-one rate is computed as the reciprocal of median latency.

\paragraph{Detector pre-training.}
Table~\ref{tab:objects365_ablation} gives the GTR-S metric breakdown with and without Objects365~\citep{shao2019objects365} detector pre-training. Overall AP increases by 2.9 and small-object AP by 5.3. The corresponding four-scale results appear in Table~\ref{tab:coco_main_full}.
\input{tables/object365_pretraining_ablation}
\FloatBarrier

\section{Experimental Details}
\label{sec:experimental_setup}

\paragraph{Datasets and metrics.}
For the main detector comparison, we report GTR with and without Objects365~\citep{shao2019objects365} detector pre-training before COCO~\citep{lin2014microsoft} \texttt{train2017} fine-tuning.
We evaluate on COCO \texttt{val2017} with AP averaged over IoU thresholds
from $0.50$ to $0.95$, together with AP$_{50}$, AP$_{75}$, and AP for small,
medium, and large objects.
We report box AP for detection and mask AP for instance segmentation.
For transfer evaluation, we report standard keypoint AP and AR on COCO
\texttt{val2017}, and AP$_{50}$ for oriented object detection on the
DOTA-v1.0 test set~\citep{xia2018dota}.

\paragraph{Training protocol.}
All models are trained on eight NVIDIA A100 GPUs. For the main comparison, the S/M/L/X backbones are initialized with the corresponding distilled GLA checkpoints, followed by pretraining of the full detectors on Objects365 for 36 epochs with a global batch size of 128.

We then train all four GTR scales for 30 epochs with a global batch size of 32.
The direct-COCO variants instead start from the output-aligned backbone and
otherwise use the same scale-specific COCO fine-tuning configurations as their
Objects365-pre-trained counterparts.
The Objects365 and COCO stages use AdamW~\citep{loshchilov2017decoupled} with a
flat-cosine learning-rate schedule and an exponential moving average;
evaluation uses $640\times640$ inputs.
Appendix~\ref{sec:detailed_training} reports the remaining scale-specific
learning rates, augmentations, loss weights, and model configurations.

\paragraph{Efficiency evaluation.}
We remeasure every detection and segmentation model in a unified FP16, batch-size-one environment on a single NVIDIA GeForce RTX~4090.
Tables~\ref{tab:coco_main_full} and~\ref{tab:coco_instance_segmentation} report the minimum, arithmetic mean, and median per-image latency; all comparisons use the median. Compilation, graph capture, autotuning, and warm-up are excluded from timing.
For DETR-style models, the measured latency covers the network forward pass
alone; for YOLO variants that require NMS, it additionally includes CUDA NMS.
Because deployment conversion, input resolution, and post-processing differ across model families, the complete protocol is specified in Appendix~\ref{sec:latency_protocol}.

\subsection{Training Configurations}
\label{sec:detailed_training}

\subsubsection{Architecture Configurations}
\label{sec:architecture_configurations}

All GTR variants employ a 12-layer, stride-16 GLA backbone and extract
multi-scale features from intermediate layers. As summarized in
Table~\ref{tab:gtr_architecture}, the decoder depth, object-query count, and
number of training query groups remain fixed across scales. From GTR-S to
GTR-L, the embedding dimension and number of GLA heads increase from 192/3 to
384/6. GTR-X retains the GTR-L token-mixing width while increasing the
S-SwiGLU expansion ratio from 4 to 6 and the decoder FFN dimension from 1,024
to 2,048.

\input{tables/gtr_architecture}

\subsubsection{Final-Output Representation Alignment}
\label{sec:representation_alignment_training}

\paragraph{Teacher configuration.}
Following~\citet{liu2026edgecrafter}, ECTeacher-S and ECTeacher-B are obtained by fine-tuning pretrained DINOv3-S and DINOv3-B in ECDet-L and ECDet-X, respectively, on COCO \texttt{train2017}. ECTeacher-S supervises GTR-S, while ECTeacher-B supervises GTR-M/L/X. During distillation, the adapted teacher backbone is frozen and provides the final patch-token targets in Equation~\ref{eq:loss_alignment}.

\paragraph{Distillation data.}
Following EdgeCrafter~\citep{liu2026edgecrafter}, the image pool combines ImageNet-1K~\citep{deng2009imagenet} and COCO~\citep{lin2014microsoft} \texttt{train2017}.

\paragraph{Student optimization.}
We adopt the optimization recipe of \citet{peng2023unimim} without additional hyperparameter tuning. GTR-S/M use the Small/Base presets, while GTR-L/X use the Large preset. As shown in Table~\ref{tab:distillation_recipe}, all variants are trained for 300 epochs with AdamW and a global batch size of 2,048. GTR-L/X use a longer warm-up (20 vs. 10 epochs) and a higher drop-path rate (0.2 vs. 0.1); all other optimization and augmentation settings are shared.

\input{tables/distillation}

Figure~\ref{fig:distillation_comparison} compares the supervision structure of GTR with representative cross-architecture approaches~\citep{wei2025vitlinearizer,li2026vitalada,moudgil2026attention,bick2024transformers}. GTR's single stage here means one representation-distillation stage; teacher preparation and downstream supervised training remain separate.
\input{figs_tex/compare_distillation}

\subsubsection{Objects365 Pre-training}
\label{sec:objects365_training}

Each detector is initialized with its corresponding output-aligned GLA
backbone and subsequently pre-trained on Objects365. As shown in
Table~\ref{tab:objects365_hparams}, all scales use the same 36-epoch schedule,
global batch size of 128, base learning rate of $2\times10^{-3}$, and 2,000
warm-up iterations. The backbone learning rate decreases as model capacity increases,
from $10^{-4}$ for GTR-S/M to $2\times10^{-5}$ and $10^{-5}$ for GTR-L and
GTR-X, respectively.

\input{tables/obj365_pre_training}

\subsubsection{COCO Fine-tuning}
\label{sec:coco_training}

For COCO fine-tuning after Objects365 pre-training, we load the complete
detector checkpoint except the classification head. As shown in
Table~\ref{tab:coco_hparams}, the 30-epoch schedule, global batch size, and base
learning rate are fixed across scales. All variants are
trained on \texttt{train2017} and evaluated on \texttt{val2017}.

\input{tables/coco_finetune}

\subsubsection{Human Pose Estimation}
\label{sec:pose_training}

For human pose estimation, we initialize only the GTR backbone from the COCO detection checkpoint and train all pose-specific parameters from scratch. The training protocol is summarized in
Table~\ref{tab:pose_hparams}. All variants use AdamW, a base learning rate of
$5\times10^{-4}$, a global batch size of 32, and the same augmentation
probabilities and loss weights.

\input{tables/coco_pose_training}
\FloatBarrier

\subsubsection{Semantic Segmentation}
\label{sec:semantic_segmentation_details}

For semantic segmentation, we initialize only the GTR backbone from the pretrained backbone checkpoint and train all task-specific parameters from scratch.

We train and evaluate semantic segmentation on Cityscapes~\citep{cordts2016cityscapes}
with fine annotations only, and report mIoU on the validation set. During training,
images are augmented with random resizing by a factor sampled from $[0.5, 2.0]$,
random horizontal flipping, and random cropping to $1024 \times 1024$. At inference,
we adopt sliding-window evaluation with a $1024 \times 1024$ window and a stride of
768 pixels.

For semantic segmentation, we retain only the \texttt{scale=[2.0]} branch of the
semantic neck, which yields stride-8 features. On top of these features we attach a
lightweight FCN head consisting of a $3\times3$ Conv-BN-ReLU block, dropout
($p=0.1$), and a $1\times1$ convolutional classifier over the 19 Cityscapes classes.
The output logits are bilinearly upsampled to the input resolution and supervised by
a per-pixel cross-entropy loss. Quantitative results are reported in
Section~\ref{sec:transfer_results}, and qualitative results are provided
in Appendix~\ref{sec:semantic_segmentation_visualization}.

Table~\ref{tab:cityscapes_hparams} lists the shared 30-epoch, batch-eight schedule and the scale-specific backbone learning rates. The reported task accuracy uses sliding-window inference, while latency measures one fixed-size network invocation.
\input{tables/cityscapes_training}
\FloatBarrier

\subsubsection{Monocular Depth Estimation with Diverse Training Data}
\label{sec:metric_depth_details}
For monocular depth estimation, we initialize only the GTR backbone from the pretrained backbone checkpoint and train all task-specific parameters from scratch.

\paragraph{Training sources and data acquisition.}
We adopt the same eight-source training mixture used for depth training in
YOLO26~\citep{yolo26}, comprising ARKitScenes~\citep{dehghan2021arkitscenes},
SUN RGB-D~\citep{song2015sun}, DIODE~\citep{vasiljevic2019diode},
Hypersim~\citep{roberts2021hypersim}, TartanAir~\citep{wang2020tartanair},
Virtual KITTI 2~\citep{cabon2020virtual}, KITTI~\citep{geiger2013vision}, and
pseudo-labeled ImageNet~\citep{deng2009imagenet}. To reduce storage and bandwidth
overhead, we adopt source-specific acquisition strategies. For Hypersim, we
download only the \texttt{tonemap.jpg} and \texttt{depth\_meters.hdf5} files
instead of the full 1.9~TB release, reducing the local storage footprint to
approximately 11~GB. For KITTI, we retrieve the raw archives with batched HTTP
range requests and keep only the color-camera streams, discarding the Velodyne point clouds, grayscale images, and
OXTS records. This reduces the amount of data transferred by about 75\%.

\paragraph{Depth decoding.}
All depth targets are decoded as \texttt{float32} arrays in meters, and values
$\leq 0$ are treated as invalid. For SUN RGB-D, we follow the official toolbox
and recover depth through the 16-bit rotation
$d=((r\mathbin{\gg}3)\mathbin{|}(r\mathbin{\ll}13))/1000$.
We use the official \texttt{depth\_bfx} targets, which leave measured pixels
unchanged while inpainting missing regions and therefore provide nearly dense
supervision. Hypersim stores the Euclidean distance $d$ from the optical center
rather than planar depth, i.e., the $z$ coordinate in camera space. Using the
official $1024\times768$ intrinsics with focal length $f=886.81$~pixels, we
convert each pixel to planar depth as
\begin{equation}
z(u,v)=d(u,v)\frac{f}{\sqrt{(u-c_x)^2+(v-c_y)^2+f^2}},
\label{eq:hypersim_plane_depth}
\end{equation}
where $(c_x,c_y)$ is the principal point. Any resulting NaN is set to zero and
thus marked invalid. KITTI depth is decoded from \texttt{uint16} as $d=r/256$,
with zeros retained as invalid. For DIODE, we multiply the depth by the provided
validity mask, which maps invalid pixels to zero. All depth maps are resized
using nearest-neighbor interpolation only, so that invalid zeros are never
blended with valid metric depths.

We further apply dataset-specific preprocessing before enforcing the common training range. In Virtual KITTI 2, sky pixels are encoded with a depth of 655.35~m; we clip these values to 80~m while retaining them as valid. In contrast, sky depths in TartanAir are on the order of $10^{4}$~m; these values are set to zero and treated as invalid. For DIODE and SUN RGB-D, depth values are clipped to 80~m and 10~m, respectively. Across all datasets, training and evaluation use a common maximum depth of 100~m, with depths in the range $[10^{-3}, 100]$~m considered valid.

\paragraph{Depth teacher.}
We generate metric-depth pseudo-labels with the nested variant of Depth
Anything~3 (DA3)~\citep{lin2026depthanything3}, which jointly estimates camera
intrinsics and metric depth and therefore needs no additional scale
calibration. The alternative canonical-depth DA3 variant instead predicts depth
up to a focal-length-dependent scale and must be rescaled as
$d_{\mathrm{metric}} = (f/300)\, d_{\mathrm{canonical}}$, where $f$ is the
focal length in pixels and $300$~pixels is the canonical focal length used
during training. This rescaling is not applicable in our setting: standard
ImageNet preprocessing discards the camera metadata and resizes and crops each
image, so the effective focal length cannot be recovered reliably. Estimating
$f$ from an assumed field of view would inject an uncontrolled, per-image scale
error into the pseudo-labels, which the nested variant avoids by construction.

\paragraph{Batched teacher inference.}
The default DA3 inference pipeline assumes multi-view inputs from a common
scene rather than a batch of independent images: it couples the views in a
batch through cross-view attention, shared scale alignment, and joint
percentile-based depth normalization. Batching unrelated ImageNet images
therefore makes each prediction depend on the other images in the batch; we
measured median relative deviations of $68$--$233\%$ from single-image
inference. To batch efficiently while preserving single-image behavior, we feed
each sample as a single-view input, yielding tensors of shape $(B,1,3,H,W)$,
and run global alignment and post-processing independently for each batch
element. The resulting predictions match single-image inference to within
$0.1\%$ relative error. We further bucket images by their resolution after
teacher preprocessing rather than by their original resolution, so that every
batch shares a single resolution while images with common aspect ratios can
still be grouped together. Overall, batching with resolution bucketing reduces
the number of teacher forward passes by a factor of approximately $8$.

\paragraph{Sampling and augmentation.}
The measured-depth datasets differ by more than an order of magnitude in size,
ranging from roughly 10K images in SUN RGB-D to over 300K in TartanAir. To
mitigate this imbalance, we oversample each dataset with an integer repetition
factor, using $[6, 3, 2, 2, 2, 1, 3]$ in data-loader order. The pseudo-labeled ImageNet pool is drawn from a separate
sampling stream, so that its ratio to the measured-depth data can be controlled
independently. For augmentation, we retain only the metric-depth-preserving
transformations of Depth Anything~V2~\citep{yang2024depthanything2}: color
jitter, horizontal flipping, aspect-ratio-preserving short-edge resizing, and
random square cropping. Composition-based geometric augmentations such as
Mosaic, ZoomOut, and IoUCrop are disabled, since they alter the relationship
between image scale and metric depth and thus break geometric consistency.

\paragraph{Head, objective, and decoding.}
Our depth head follows a DPT-style RefineNet hierarchy~\citep{ranftl2021dpt}: starting from
stride-32 features, it progressively fuses those at strides 16, 8, and 4, and upsamples the
final prediction to the input resolution. Let $\hat{d}_i$ and $d_i$ denote the predicted and
ground-truth depth at pixel $i$, let $N$ be the number of valid pixels, and let
$g_i=\log\hat{d}_i-\log d_i$. We minimize the scale-invariant logarithmic (SiLog)
loss~\citep{eigen2014depth}, a standard objective for metric depth regression,
\begin{equation}
\mathcal{L}_{\mathrm{SiLog}}
=\sqrt{\frac{1}{N}\sum_i g_i^{2}
-\lambda\left(\frac{1}{N}\sum_i g_i\right)^{2}},
\qquad \lambda=0.5 .
\label{eq:depth_silog}
\end{equation}
Unlike the fully scale-invariant case ($\lambda=1$), $\lambda=0.5$ retains a partial penalty
on the global log-scale error. SiLog is the only objective used to train the depth head.
For decoding, we follow YOLO26~\citep{yolo26} and regress log-depth,
$\hat{d}=\exp\!\left(\operatorname{clamp}(z,-4,5)\right)$, where $z$ is the raw head output;
this yields an effective range of approximately $0.02$--$148$\,m. The clamp serves only for
numerical stability and its bounds are dataset-agnostic, so predictions are not saturated at
a dataset-specific \texttt{max\_depth}.

\paragraph{Ground-truth-aligned evaluation protocol.}
We evaluate on the 654-image Eigen test split of NYU Depth
V2~\citep{silberman2012indoor,eigen2014depth} without NYU-specific fine-tuning. We use filled ground-truth depth maps for evaluation.
Each RGB image is first resized
with its shorter side set to 640 pixels while preserving aspect ratio; hence, a
$480\times640$ image becomes $640\times853$. Because the backbone operates on
a square patch grid, two overlapping $640\times640$ windows cover the resized
image, and predictions in their overlap are averaged. We use three image
scales, $\{0.75,1,1.25\}$, with and without horizontal flipping, yielding six
test-time views. Each prediction is bilinearly resized back to the native
$480\times640$ resolution, after which the six depth maps are averaged
linearly. Within the Eigen crop $[45{:}471,\,41{:}601]$, we fit a two-parameter
log-affine alignment to the ground-truth depth independently for each image by
least squares. For an aligned prediction $\tilde d_i$, ground-truth depth $d_i$,
and the set $\mathcal{V}$ of $N$ valid pixels in the crop, we define the three
metrics as
\begin{align}
\delta_1
&= \frac{1}{N}\sum_{i\in\mathcal{V}}
\mathbf{1}\!\left[
\max\!\left(\frac{\tilde d_i}{d_i},\frac{d_i}{\tilde d_i}\right)<1.25
\right], \\
\mathrm{AbsRel}
&= \frac{1}{N}\sum_{i\in\mathcal{V}}
\frac{\lvert \tilde d_i-d_i\rvert}{d_i}, \\
\mathrm{RMSE}
&= \sqrt{\frac{1}{N}\sum_{i\in\mathcal{V}}(\tilde d_i-d_i)^2}.
\end{align}
We compute each metric separately for every image and report its arithmetic mean
over all 654 test images. This protocol follows YOLO26~\citep{yolo26}.
Section~\ref{sec:transfer_results} reports the quantitative comparison, and
Appendix~\ref{sec:depth_visualization} provides additional zero-shot results on
indoor and outdoor benchmarks.

Table~\ref{tab:depth_pretrain_hparams} gives the mixed-data depth schedule. The four models share the 30-epoch duration and batch size of 256, while the L/X backbone learning rates are lower than those of S/M. The ground-truth alignment above is applied only when computing the reported NYU evaluation metrics.
\input{tables/depth_pretraining}
\FloatBarrier

\subsubsection{Oriented Object Detection}
\label{sec:obb_implementation_details}

For oriented object detection, we initialize only the GTR backbone from the pretrained backbone checkpoint and train all task-specific parameters from scratch.

For DOTA-v1.0, we follow the multi-scale training and testing protocol of
RiO-DETR~\citep{hu2026rio}: during both training and testing, each image is
rescaled by factors of
$\{0.5, 1.0, 1.5\}$ and then cropped into $1024\times1024$ patches with a
stride of $524$ pixels, i.e., an overlap of $500$ pixels. The task-specific
OBB decoder consists of four layers and introduces three modifications for
oriented object detection. First, we adopt angle distribution
refinement (ADR)~\citep{ding2026oriented}, which represents each oriented box
by six discrete distributions: four distances to the sides of its
axis-aligned enclosing box and two vertex offsets. These distributions are
decoded into oriented boxes by a distance-to-rotated-box conversion, which
avoids the angular boundary discontinuity of direct angle regression.
Second, we rotate the sampling offsets of deformable attention by the
predicted angle $\theta$ of each query, so that the sampling pattern aligns
with the principal direction of the target. Third, oriented contrastive
denoising (OCD)~\citep{ding2026oriented} perturbs only the axis-aligned
coordinates $(x_1,y_1,x_2,y_2)$ of the denoising queries and leaves the
angle channel unchanged.

The training objective combines one classification and two regression terms.
For classification, the matching-aware loss (MAL)~\citep{huang2025deim} uses a
soft target weighted by the rotated IoU with the ground truth. For
regression, an $\ell_1$ loss supervises the five box parameters
$(x, y, w, h, \theta)$, and the Gaussian Kullback--Leibler divergence (KLD)
loss~\citep{yang2021kld} replaces the GIoU loss used in the horizontal-box
setting: it models an oriented box as a 2-D Gaussian and thus provides
geometry-aware supervision for rotated boxes. As in the original
formulation, the divergence $D$ is mapped to a loss by
$1 - 1/(\tau + \log(1+D))$ with $\tau_{\mathrm{KLD}} = 1$.
Table~\ref{tab:dota_obb_hparams} summarizes the complete fine-tuning
configuration.

\input{tables/dota_obb_training}
\FloatBarrier

\section{Efficiency and Reproducibility Details}
\label{sec:efficiency_details}
\subsection{DRIVE AGX Thor Deployment}
\label{sec:thor_deployment}

\paragraph{Platform and execution path.}
Table~\ref{tab:thor_platform} specifies the on-device environment. The Thor-targeted operators are compiled for \texttt{sm\_110} and
integrated into TensorRT; they are distinct builds from the \texttt{sm\_89}
PyTorch-extension GLA operator benchmarked in
Appendix~\ref{sec:cuda_operator_protocol}.

\begin{table}[t]
  \centering
  \caption{\textbf{DRIVE AGX Thor deployment environment.} All results in
  Tables~\ref{tab:thor_bs1} and~\ref{tab:thor_bs8} use FP16.}
  \label{tab:thor_platform}
  \small
  \setlength{\tabcolsep}{5pt}
  \begin{tabular}{@{}p{0.29\columnwidth}p{0.62\columnwidth}@{}}
    \toprule
    Component & Configuration \\
    \midrule
    Board & NVIDIA DRIVE AGX Thor \\
    GPU & Blackwell (\texttt{sm\_110}) \\
    Operating system & DRIVE OS~7.0.5.0 / Ubuntu~24.04.3 LTS \\
    CUDA & 13.0 \\
    TensorRT & 10.14.2.2 \\
    Precision & FP16 \\
    \bottomrule
  \end{tabular}
\end{table}

\paragraph{Roofline analysis.}
Figure~\ref{fig:thor_roofline} motivates optimizing memory traffic rather than peak compute. Across six tasks and four model scales, the plotted baseline and optimized points lie well left of the \(1{,}032\,\mathrm{FLOP/byte}\) ridge point and near the bandwidth roofs, consistent with memory-traffic limitations on Thor. Compared with the TensorRT baseline, our graph transformations and fused CUDA kernels shift the operating points upward and to the right by eliminating intermediate memory traffic. The optimized models reach \(89\text{--}141\%\) of the nominal DRAM roof; values above \(100\%\) are compatible with on-chip reuse under the plotted traffic model.

\input{figs_tex/roofline_model}

\paragraph{Deployment results.}
Tables~\ref{tab:thor_bs1} and~\ref{tab:thor_bs8} report the complete TensorRT measurements. At batch size one, latency ranges from 2.282 to 8.769\,ms across all tasks and model scales, while the cosine similarity between FP16 TensorRT outputs and their FP32 PyTorch references remains consistently high, ranging from 0.9989 to 1.0000. Increasing the batch size to eight raises the peak throughput to 571 images/s for GTR-S detection. For the \(1024\times1024\) oriented-detection and semantic-segmentation workloads, batch-one inference provides higher throughput than batch eight, consistent with the memory-bound regime identified in Figure~\ref{fig:thor_roofline}.

\input{tables/thor_deployment_bs1}
\input{tables/thor_deployment_bs8}

\subsection{GPU Latency Measurement Protocol}
\label{sec:latency_protocol}

\subsubsection{Hardware, Software, and Deployment}

Our protocol measures steady-state GPU latency under a unified deployment, compilation, and execution pipeline.
The model forward-pass measurements use a single NVIDIA GeForce RTX~4090 GPU
with Python~3.11, PyTorch~2.11.0+cu128, torchvision~0.26.0+cu128, and
CUDA~12.8.
We use a batch size of one and FP16 precision throughout.

For object detection, RF-DETR-S/M/L/X~\citep{robinson2025rf} are evaluated at
their released native resolutions of $512/576/704/700$, respectively; the
non-monotonic L/X resolutions follow the official model configurations. All
other detectors use $640\times640$ inputs.
For instance segmentation, RF-DETR-Seg-S/M/L/X use their native resolutions of $384/432/504/624$, respectively, while all other models use $640\times640$ inputs.

Before compilation, we invoke the official \texttt{deploy()}, \texttt{export()}, or \texttt{fuse()} routine, when available, to apply deployment conversion and Conv--BN fusion.
TorchInductor freezing is enabled for implementations without an explicit deployment interface.
Each model is compiled with
\texttt{torch.compile(mode="max-autotune", fullgraph=True)}
and executed through a CUDA Graph.
The \texttt{fullgraph=True} constraint prevents graph breaks or device synchronizations from silently causing a fallback to eager execution.

Profiler traces confirm that each steady-state forward pass incurs exactly one CUDA Graph launch, without GPU-to-CPU scalar synchronization or device-to-host transfer.
The official RF-DETR~\citep{robinson2025rf} and
LW-DETR~\citep{chen2024lwdetr} implementations use GPU-resident shape scalars
and tensor assertions that prevent full-graph compilation.
We replace these operations with compile-time shape constants, producing bitwise-identical outputs.
Compilation, autotuning, graph capture, and warm-up are excluded from all reported measurements.

\subsubsection{Timed Scope and Post-processing}

For DETR-style models, we measure only the forward pass and exclude output decoding and post-processing.
For YOLOv9~\citep{wang2024yolov9}, YOLO11~\citep{yolo11}, and
YOLOv12~\citep{tian2025yolov12}, the reported latency includes both the
forward pass and CUDA NMS implemented with \texttt{torchvision.ops.nms}.
We use \texttt{conf=0.001}, \texttt{iou=0.7},
\texttt{multi\_label=True}, \texttt{agnostic=False}, and
\texttt{max\_det=300}.

Because NMS cost depends on the number of candidate boxes, these models are
evaluated over all 5,000 images in COCO~\citep{lin2014microsoft}
\texttt{val2017}.
YOLOv10~\citep{wang2024yolov10} and YOLO26~\citep{yolo26} perform in-network
top-$k$ selection and require no separate NMS stage.
All NMS-free detectors are evaluated on a fixed subset of 1,000 COCO images.

\subsubsection{Warm-up and Statistics}

YOLO models are warmed up for 200 forward iterations.
YOLO11 and YOLOv12 receive an additional 30 warm-up iterations covering the complete prediction and NMS pipeline.
YOLOv9 uses no separate NMS warm-up; its one-time NMS initialization cost is negligible when averaged over 5,000 images.
All other detection and segmentation models are warmed up for 300 iterations.

Latency is measured using CUDA events on the default stream while the GPU is otherwise idle.
Each model is benchmarked once in an independent process.
Table~\ref{tab:coco_main_full} reports the minimum, arithmetic mean, and median per-image latency. We use the median for every model comparison.

\FloatBarrier
\subsection{Custom CUDA Operator}
\label{sec:cuda_operator_protocol}

\subsubsection{Implementation Details}
\label{sec:cuda_operator_impl}

\paragraph{Scope and decomposition.}
The reference FLA implementation~\citep{yang2024fla} targets general training
and inference settings, supporting variable head dimensions, sequence
lengths, recurrent states, backward propagation, and final-state outputs. In
contrast, our CUDA operator is specialized for GTR inference with batch size
one, FP16 inputs, a zero initial state, no backward propagation or
final-state output, chunk size \(C=64\), and per-head dimensions \(d_k=32\)
and \(d_v=64\).

As described in Section~\ref{sec:preliminaries}, GTR fixes the value-side
state-decay gate to \(\boldsymbol{\beta}_t\equiv\mathbf{1}_{1\times d_v}\)
during both training and inference. Consequently, the training-time FLA
kernels and our custom operator consume identical inputs---\(\mathbf{Q}\),
\(\mathbf{K}\), \(\mathbf{V}\), and the learned key-side log-decay gate
\(\boldsymbol{g}\)---and evaluate the same recurrence. The data-dependent output gate is
likewise preserved and may be fused into the deployment epilogue. Up to
mixed-precision roundoff, our operator therefore changes only the execution
schedule and fusion pattern, not the underlying computation.

A remaining inefficiency in FLA v0.5.0 is that each state-tile program
processes chunks sequentially, performing a matrix contraction at every step
and exposing limited parallelism in our small-batch regime. Let
\(N_c=\lceil L/C\rceil\) denote the number of chunks. We instead parallelize
the expensive chunk summaries \(\mathbf{U}_c\) in
Equation~\ref{eq:chunk_boundary_main} over the chunk--head grid, leaving only
the lightweight boundary-state recurrence as a sequential scan; adjacent
lanes process adjacent elements along the contiguous \(d_v\) dimension,
enabling coalesced memory accesses. While a tree-based prefix scan achieves
\(\mathcal{O}(\log N_c)\) depth, it requires additional global-memory passes
and strided accesses in our setting. We find the coalesced linear scan faster
in the target deployment regime while still retaining \(\mathcal{O}(L)\) work
for fixed head dimensions and chunk size.

\paragraph{Fused tensor-core dataflow.}
Building on this schedule, our first kernel fuses chunk-local cumulative sums
with the construction of \(\mathbf{U}_c\). Prefix sums are computed in FP32
registers, while value tiles are staged in padded shared memory via
\texttt{cp.async}. The fixed configuration \(C=64\), \(d_k=32\), \(d_v=64\)
maps efficiently to \texttt{ldmatrix} loads and tensor-core MMA instructions
with FP32 accumulation: chunk-summary and score--value contractions use FP16
operands, whereas the decay-rescaled query/key operands and boundary states
use BF16 to accommodate their wider dynamic range. Kernel tiling and dispatch
parameters are tuned for RTX~4090 \texttt{sm\_89}.

For each chunk--head pair, a second kernel jointly computes the
causal \(\widetilde{\mathbf{Q}}\widetilde{\mathbf{K}}^\top\) scores, their
product with \(\mathbf{V}\), and the boundary-state readout within a single
thread block. The resulting \(64\times64\) causal score tile is consumed
directly from shared memory rather than materialized to global memory,
eliminating the standalone cumulative-sum launch and an intermediate
intra-chunk score tensor of size \(\mathcal{O}(LHC)\). Chunk-local cumulative
sums are still exchanged between the summary and output kernels through an
FP16 workspace.

\paragraph{Length-aware scheduling and deployment fusion.}
Because the relative cost of the scan and output stages depends on \(N_c\),
we adopt a length-aware dispatch policy. For short sequences, a fallback
output kernel reconstructs chunk prefixes inline, avoiding a separate scan
launch; under our default RTX~4090 dispatch policy, this path is used for
\(N_c<6\). For eligible non-wave-aligned grids with \(N_c\geq6\), an
occupancy-aware kernel instead combines the scan and output stages: output
blocks compute the chunk-local contractions while scan blocks generate
boundary states, with synchronization occurring only immediately before the
boundary-state readout so that the two stages can overlap. All other
configurations retain a separate scan kernel.

Beyond kernel scheduling, the deployment interface supports persistent
caller-managed workspaces, preallocated output buffers, and CUDA Graph
capture. In the end-to-end GTR implementation, the output epilogue further
fuses RMSNorm and the SiLU-activated output gate---distinct from the
subsequent S-SwiGLU channel mixer---directly into the GLA kernel, removing
one kernel launch and its intermediate global-memory round trip. This
epilogue fusion is excluded from the isolated operator measurements in
Figure~\ref{fig:chunk_gla_latency}.

\section{Qualitative Results}
\label{sec:qualitative_results}

We present qualitative results for object detection, instance segmentation,
human pose estimation, semantic segmentation, zero-shot monocular depth
estimation, and oriented object detection.
The COCO examples include occlusion, low illumination, motion blur, defocus,
object rotation, and crowded scenes. The additional transfer examples use
Cityscapes~\citep{cordts2016cityscapes} for semantic segmentation, the DTU
Robot Image Data Set~\citep{jensen2014large} for indoor depth,
nuScenes~\citep{caesar2020nuscenes} for outdoor depth, and the
DOTA-v1.0~\citep{xia2018dota} test set for oriented object detection.

\subsection{Object Detection Results}
\label{sec:detection_visualization}

\paragraph{Detection behavior.}
Figure~\ref{fig:detection_visualization} illustrates selected GTR-X
predictions under crowding and occlusion, motion blur, object rotation, and
low illumination. Across these examples, the model localizes small and
overlapping instances despite substantial appearance degradation.

\input{figs_tex/results_detection}

\subsection{Instance Segmentation Results}
\label{sec:segmentation_visualization}

\paragraph{Mask quality.}
Figure~\ref{fig:segmentation_visualization} visualizes GTR-X masks on the same
challenging images as Figure~\ref{fig:detection_visualization}. The selected
predictions follow visible object contours and separate adjacent instances;
together with Table~\ref{tab:coco_instance_segmentation}, they illustrate the
transfer of the output-aligned representation to instance segmentation.

\input{figs_tex/results_segmentation}

\subsection{Human Pose Estimation Results}
\label{sec:pose_visualization}

\paragraph{Pose behavior.}
The selected cases in Figure~\ref{fig:pose_visualization} show GTR-X predicting
articulated poses for multiple overlapping people under crowding, low
illumination, and motion blur.

\input{figs_tex/results_pose}

\subsection{Semantic Segmentation Results}
\label{sec:semantic_segmentation_visualization}

\paragraph{Dense semantic predictions.}
Figure~\ref{fig:semantic_segmentation_visualization_1} compares GTR-X predictions
with Cityscapes ground truth.

\input{figs_tex/results_semantic_segmentation}

\FloatBarrier

\subsection{Zero-Shot Monocular Depth Estimation Results}
\label{sec:depth_visualization}

We apply GTR-L without dataset-specific fine-tuning to indoor DTU and outdoor
nuScenes imagery.

\subsubsection{Indoor DTU Results}
\label{sec:depth_dtu_visualization}

\paragraph{Projected reference depth.}
For these examples, DTU provides calibrated point-cloud geometry rather than
dense per-view depth maps. We therefore construct reference depth by
projecting the point cloud through each camera using the provided intrinsics
and calibrated pose. The resulting reference maps cover 67.9\% of pixels on
average; missing support is concentrated in the background fabric, which was
not captured by structured light, and on specular surfaces.
Figure~\ref{fig:depth_dtu_predictions} compares the RGB inputs, GTR-L predictions,
and projected references.

\paragraph{Depth-based reconstruction.}
Figure~\ref{fig:depth_dtu_reconstruction} visualizes reconstructions obtained from all 49 calibrated views of each DTU object. Since a single image does not geometrically determine absolute depth scale, depth maps predicted independently across views may exhibit view-dependent scale drift. Prior to fusion, we therefore align the scale of each predicted depth map using a single scalar, computed as the median ratio between the projected structured-light reference depth and the prediction over valid pixels. This per-view rescaling preserves the relative depth structure within each view while correcting only its global scale.

We then back-project the rescaled depth maps using the calibrated camera intrinsics and extrinsics, fuse all 49 views into a TSDF~\citep{curless1996volumetric} volume with 2-mm voxels, and render the resulting surface as colored points. Because the structured-light reference supplies one global scale factor per view, these reconstructions illustrate the geometry after reference-assisted scale correction.

\input{figs_tex/results_depth_dtu}
\FloatBarrier

\subsubsection{Outdoor nuScenes Results}
\label{sec:depth_nuscenes_visualization}

\paragraph{Surround-view depth.}
Figures~\ref{fig:depth_nuscenes_1}--\ref{fig:depth_nuscenes_3} show GTR-L
predictions for the six synchronized surround-view cameras in three nuScenes
scenes. Each figure preserves the camera ordering between the RGB grid and
the corresponding depth grid, covering daytime, nighttime, and roadside
construction imagery without nuScenes-specific fine-tuning.

\input{figs_tex/results_depth_nuscenes}

\paragraph{Temporally fused local maps.}

To visualize multi-frame geometric consistency, for each reference time \(t\), we aggregate all available frames within \([t-4\,\mathrm{s},\,t+4\,\mathrm{s}]\). For each frame, we scale-align the GTR-L depth prediction using a single scalar estimated from the corresponding LiDAR measurements, back-project the corrected depth with the camera intrinsics, and transform the resulting points into a common coordinate frame using the calibrated pose. We then fuse all observations with TSDF~\citep{curless1996volumetric} to obtain a local map. Figures~\ref{fig:depth_nuscenes_bev_1} and~\ref{fig:depth_nuscenes_bev_2} show two examples from perspective and bird's-eye views. Since LiDAR is used only for per-frame scale alignment, these results qualitatively illustrate geometric consistency after scale correction.

\input{figs_tex/results_depth_nuscenes_mapping}

\subsection{Oriented Object Detection Results}
\label{sec:obb_visualization}

\paragraph{Oriented localization.}
Figure~\ref{fig:obb_visualization} shows the GTR-X predictions in the DOTA-v1.0 test
images.
The oriented boxes track ships and vehicles at various angles and remain
separated in densely packed harbor and road scenes, complementing the
quantitative results in Table~\ref{tab:dota_obb_full}.

\makeatletter
\setlength{\@fptop}{0pt}
\setlength{\@dblfptop}{0pt}
\makeatother
\input{figs_tex/results_obb}
\FloatBarrier

%% file: sec/2_related.tex
\section{Related Work}
\label{sec:related_work}

\paragraph{Object detection with Transformers.}
DETR~\citep{carion2020end} formulates object detection as set prediction
with Transformers, eliminating the need for separate non-maximum suppression (NMS). To address the slow convergence and high computational cost of the
original DETR, a series of follow-up works has been
proposed~\citep{zhu2021deformable,li2022dn, wang2022anchor,
meng2021conditional, chen2023group}. RT-DETR~\citep{lv2024rt} introduces a
hybrid encoder for real-time object detection. Building upon this framework,
D-FINE~\citep{peng2025dfine} further enhances detection performance through
fine-grained distribution refinement and self-distillation. Beyond
architectural improvements, several studies have explored incorporating Vision
Transformers (ViTs)~\citep{dosovitskiy2020image} into the DETR framework.
LW-DETR~\citep{chen2024lwdetr} adopts a self-supervised pre-trained ViT as the
backbone, while RF-DETR~\citep{robinson2025rf} combines neural architecture
search with the visual foundation model
DINOv2~\citep{oquab2023dinov2} to strengthen feature representation. More
recently, EdgeCrafter~\citep{liu2026edgecrafter} improves compact ViT backbones
via task-specific distillation from DINOv3~\citep{simeoni2025dinov3}.

\paragraph{Linear-complexity attention.}
Linear attention factorizes kernelized attention and exploits associativity to
avoid materializing the dense token-to-token matrix, reducing complexity from
$O(L^2)$ to $O(L)$ and admitting recurrent
evaluation~\citep{katharopoulos2020transformers}. Modern linear-time recurrent
families, including RetNet~\citep{sun2023retnet}, RWKV~\citep{peng2023rwkv}, Mamba~\citep{gu2024mamba}, DeltaNet~\citep{yang2024deltanet}, and GLA~\citep{yang2024gla}, enrich the recurrent
state with retention, selectivity, delta-rule updates, or data-dependent gates
while retaining parallel or chunkwise training
forms.
ViG~\citep{liao2025vig} is the closest architectural precedent: it adapts GLA to vision with within-block bidirectional modeling, direction-wise gating, and local two-dimensional interaction, and provides non-hierarchical and hierarchical variants. Its fused bidirectional kernel also targets practical efficiency. GTR uses one scan per block, alternates directions across depth, and introduces local mixing in the SwiGLU channel mixer. We evaluate these choices together with a detection-specialized distillation recipe and task-specific deployment.

Hybrid approaches such as MambaVision~\citep{hatamizadeh2025mambavision} and SoLA-Vision~\citep{li2026solavision} combine recurrent or linear mixers with softmax attention. Our hybrid ablation tests three softmax-layer placements within GTR-L. Kernel-based vision alternatives include LaplacianFormer's low-rank Laplacian attention~\citep{zhe2026rethinking}. GTR builds on the established GLA recurrence and chunkwise formulation~\citep{yang2024gla} and contributes a fused, hardware-aware execution schedule for low-latency inference.

\paragraph{Heterogeneous representation distillation.}

Cross-architecture transfer often uses several stages or complementary
targets to bridge different token mixers. ViT-Linearizer combines
intermediate token-affinity matching at multiple depths with Smooth-$\ell_1$
prediction of final teacher features at masked
positions~\citep{wei2025vitlinearizer}.
ViT-AdaLA~\citep{li2026vitalada} first matches
the output of every softmax- and linear-attention module with MSE, then aligns
the final representations of the complete models with MSE, and finally
performs supervised downstream fine-tuning. Attention to
Mamba~\citep{moudgil2026attention} first learns a Hedgehog feature map through
per-layer cosine output matching, uses it to initialize an adapted Mamba
model, and then fine-tunes the model with next-token cross-entropy. Related
Transformer-to-SSM studies likewise use staged conversion or additional
architecture-specific objectives~\citep{bick2024transformers,bick2025llamba,
wang2024mamba}. In the detection-specialized DINOv3-to-GLA setting studied here, we use
a single squared $\ell_2$ loss on the final output representations, without
intermediate or stage-specific distillation targets and without input masking.

%% file: tables/coco_main_table_layout.tex
\begin{table}[H]
  \centering
  \caption{\textbf{Comparison with real-time object detectors on COCO~\citep{lin2014microsoft} \texttt{val2017}, grouped by model scale.}
  Detailed latency settings are provided in Appendix~\ref{sec:latency_protocol}.
  For comparison methods, all non-latency entries are taken from EdgeCrafter~\citep{liu2026edgecrafter}, while latency is measured under our unified protocol.
  The epoch column indicates the COCO training schedule, where $^{*}$ denotes additional epochs without strong augmentation.
  $\ddagger$ marks models without Objects365~\citep{shao2019objects365} detector pre-training.
  Within each comparison group, the top three distinct results for median
  latency and every reported accuracy metric are highlighted in dark, medium,
  and light blue, respectively; rankings use unrounded values.}
  \label{tab:coco_main_full}

  \tiny
  \setlength{\tabcolsep}{2.15pt}
  \renewcommand{\arraystretch}{1.00}

  \begin{adjustbox}{width=\textwidth}
  \begin{tabular}{@{}lcccccccccccc@{}}
    \toprule
    \multirow{2}{*}{Model}
      & \multirow{2}{*}{\#Epochs}
      & \multirow{2}{*}{\makecell{\#Params\\(M)}}
      & \multirow{2}{*}{GFLOPs}
      & \multicolumn{3}{c}{Latency (ms) $\downarrow$}
      & \multicolumn{6}{c}{COCO AP $\uparrow$} \\    \cmidrule(lr){5-7}
    \cmidrule(lr){8-13}
      & & & & Min. & Mean & Median
      & AP$^{val}$ & AP$^{val}_{50}$ & AP$^{val}_{75}$
      & AP$^{val}_{S}$ & AP$^{val}_{M}$ & AP$^{val}_{L}$ \\    \midrule

    YOLOv9-S$^{\ddagger}$~\citep{wang2024yolov9} & 500 & 7 & 26
      & 1.651 & 2.326 & 2.294
      & 46.8 & 61.8 & 48.6 & 25.7 & 49.9 & 61.0 \\

    YOLOv10-S$^{\ddagger}$~\citep{wang2024yolov10} & 500 & 7 & 22
      & 0.870 & 0.934 & \rankfirst{0.933}
      & 46.3 & 63.0 & 50.4 & 26.8 & 51.0 & 63.8 \\

    YOLO11-S$^{\ddagger}$~\citep{yolo11} & 500 & 9 & 22
      & 1.078 & 1.523 & 1.490
      & 46.6 & 63.4 & 50.3 & 28.7 & 51.3 & 64.1 \\

    YOLOv12-S-Turbo$^{\ddagger}$~\citep{tian2025yolov12} & 600 & 9 & 19
      & 1.339 & 1.802 & 1.772
      & 47.6 & 64.5 & 51.5 & 28.3 & 52.7 & 65.9 \\

    RT-DETRv2-S$^{\ddagger}$~\citep{lv2024rt} & 120 & 20 & 60
      & 1.184 & 1.279 & 1.291
      & 48.1 & 65.1 & 52.1 & 30.2 & 51.5 & 63.9 \\

    DEIM-S$^{\ddagger}$~\citep{huang2025deim} & $132^{*}$ & 10 & 25
      & 1.090 & 1.171 & 1.180
      & 49.0 & 65.9 & 53.1 & 30.4 & 52.6 & 65.7 \\

    DEIMv2-S$^{\ddagger}$~\citep{huang2025deimv2} & $132^{*}$ & 10 & 26
      & 1.550 & 1.676 & 1.687
      & 50.9 & 68.4 & 55.1 & 31.3 & 55.3 & \rankthird{70.2} \\

    RT-DETRv4-S$^{\ddagger}$~\citep{liao2025rtdetrv4} & $132^{*}$ & 10 & 25
      & 1.079 & 1.166 & 1.174
      & 49.7 & 66.8 & 54.1 & 30.2 & 53.6 & 66.9 \\

    LW-DETR-S~\citep{chen2024lwdetr} & 60 & 15 & 17
      & 1.035 & 1.115 & \rankthird{1.123}
      & 48.0 & 66.9 & 51.7 & 26.8 & 52.5 & 65.5 \\

    D-FINE-S~\citep{peng2025dfine} & -- & 10 & 25
      & 1.103 & 1.198 & 1.194
      & 50.7 & 67.6 & 55.1 & \ranksecond{32.7} & 54.6 & 66.5 \\

    RF-DETR-S~\citep{robinson2025rf} & -- & 32 & 60
      & 1.207 & 1.285 & 1.299
      & \ranksecond{52.9} & \rankfirst{71.9} & \ranksecond{57.0}
      & 32.0 & \rankfirst{58.3} & \rankfirst{73.0} \\

    YOLO26-S~\citep{yolo26} & 70 & 10 & 21
      & 0.922 & 0.977 & \ranksecond{0.962}
      & 47.8 & 64.6 & 52.1 & 29.1 & 52.5 & 64.3 \\

    ECDet-S$^{\ddagger}$~\citep{liu2026edgecrafter} & 74 & 10 & 26
      & 1.511 & 1.641 & 1.652
      & \rankthird{51.7} & \rankthird{69.4} & \rankthird{55.8}
      & \rankthird{32.3} & \ranksecond{56.4} & \ranksecond{70.5} \\

    \textbf{GTR-S (ours)}$^{\ddagger}$ & 30 & 12.1 & 33.8
      & 1.215 & 1.225 & 1.225
      & 50.7 & 68.5 & 54.7 & 31.1 & \rankthird{55.5} & \rankthird{70.2} \\

    \textbf{GTR-S (ours)} & 30 & 12.1 & 33.8
      & 1.215 & 1.225 & 1.225
      & \rankfirst{53.6} & \ranksecond{71.1} & \rankfirst{58.3}
      & \rankfirst{36.4} & \rankfirst{58.3} & \rankthird{70.2} \\

    \midrule

    YOLOv9-M$^{\ddagger}$~\citep{wang2024yolov9} & 500 & 20 & 76
      & 2.023 & 2.713 & 2.675
      & 51.4 & 67.2 & 54.6 & 32.0 & 55.7 & 66.4 \\

    YOLOv10-M$^{\ddagger}$~\citep{wang2024yolov10} & 500 & 15 & 59
      & 1.415 & 1.437 & \rankfirst{1.426}
      & 51.1 & 68.1 & 55.8 & 33.8 & 56.5 & 67.0 \\

    YOLO11-M$^{\ddagger}$~\citep{yolo11} & 500 & 20 & 68
      & 1.643 & 2.086 & 2.061
      & 51.2 & 67.9 & 55.3 & 33.0 & 56.7 & 67.5 \\

    YOLOv12-M-Turbo$^{\ddagger}$~\citep{tian2025yolov12} & 600 & 20 & 60
      & 1.917 & 2.368 & 2.342
      & 52.5 & 69.9 & 57.1 & 35.2 & 57.8 & 69.7 \\

    RT-DETRv2-M$^{\ddagger}$~\citep{lv2024rt} & 120 & 31 & 92
      & 1.593 & 1.727 & 1.741
      & 49.9 & 67.5 & 54.1 & 32.0 & 53.2 & 66.5 \\

    DEIM-M$^{\ddagger}$~\citep{huang2025deim} & $102^{*}$ & 19 & 57
      & 1.612 & 1.732 & 1.744
      & 52.7 & 70.0 & 57.3 & 35.3 & 56.7 & 69.5 \\

    DEIMv2-M$^{\ddagger}$~\citep{huang2025deimv2} & $102^{*}$ & 18 & 52
      & 2.193 & 2.390 & 2.398
      & 53.0 & 70.2 & 57.6 & 34.2 & 57.4 & 71.5 \\

    RT-DETRv4-M$^{\ddagger}$~\citep{liao2025rtdetrv4} & $102^{*}$ & 19 & 57
      & 1.598 & 1.731 & 1.742
      & 53.5 & 71.1 & 58.1 & 34.9 & 57.7 & 72.1 \\

    LW-DETR-M~\citep{chen2024lwdetr} & 60 & 28 & 43
      & 1.342 & 1.433 & \rankthird{1.449}
      & 52.6 & 69.9 & 56.7 & 32.6 & 57.7 & 70.7 \\

    D-FINE-M~\citep{peng2025dfine} & -- & 19 & 57
      & 1.603 & 1.732 & 1.740
      & \ranksecond{55.1} & \rankthird{72.6} & \ranksecond{59.7}
      & \ranksecond{37.9} & \rankthird{59.4} & 71.7 \\

    RF-DETR-M~\citep{robinson2025rf} & -- & 34 & 79
      & 1.326 & 1.428 & \ranksecond{1.431}
      & \rankthird{54.7} & \ranksecond{73.5} & \rankthird{59.2}
      & 36.1 & \ranksecond{59.7} & \ranksecond{73.8} \\

    YOLO26-M~\citep{yolo26} & 80 & 20 & 68
      & 1.459 & 1.481 & 1.473
      & 52.5 & 69.8 & 57.2 & \rankthird{36.2} & 56.9 & 68.5 \\

    ECDet-M$^{\ddagger}$~\citep{liu2026edgecrafter} & 62 & 18 & 53
      & 1.901 & 2.080 & 2.095
      & 54.3 & 72.2 & 58.7 & 35.9 & 59.1 & 72.7 \\

    \textbf{GTR-M (ours)}$^{\ddagger}$ & 30 & 22.7 & 62.6
      & 1.450 & 1.462 & 1.462
      & 54.0 & 72.2 & 58.5 & 35.3 & 59.1 & \rankthird{73.1} \\

    \textbf{GTR-M (ours)} & 30 & 22.7 & 62.6
      & 1.450 & 1.462 & 1.462
      & \rankfirst{57.3} & \rankfirst{75.0} & \rankfirst{62.4}
      & \rankfirst{41.1} & \rankfirst{62.0} & \rankfirst{74.1} \\

    \midrule

    YOLOv9-C$^{\ddagger}$~\citep{wang2024yolov9} & 500 & 25 & 102
      & 2.049 & 2.716 & 2.680
      & 53.0 & 70.2 & 57.8 & 36.2 & 58.5 & 69.3 \\

    YOLOv10-L$^{\ddagger}$~\citep{wang2024yolov10} & 500 & 24 & 120
      & 2.042 & 2.059 & 2.056
      & 53.2 & 70.1 & 58.1 & 35.8 & 58.5 & 69.4 \\

    YOLO11-L$^{\ddagger}$~\citep{yolo11} & 500 & 25 & 87
      & 2.195 & 2.641 & 2.617
      & 53.4 & 70.1 & 58.2 & 35.6 & 59.1 & 69.2 \\

    YOLOv12-L-Turbo$^{\ddagger}$~\citep{tian2025yolov12} & 600 & 27 & 82
      & 2.879 & 3.328 & 3.305
      & 53.8 & 71.0 & 58.6 & 36.9 & 59.4 & 71.0 \\

    RT-DETRv2-L$^{\ddagger}$~\citep{lv2024rt} & 72 & 42 & 136
      & 2.127 & 2.317 & 2.332
      & 53.4 & 71.6 & 57.4 & 36.1 & 57.9 & 70.8 \\

    DEIM-L$^{\ddagger}$~\citep{huang2025deim} & $58^{*}$ & 31 & 91
      & 2.251 & 2.446 & 2.460
      & 54.7 & 72.4 & 59.4 & 36.9 & 59.6 & 71.8 \\

    DEIMv2-L$^{\ddagger}$~\citep{huang2025deimv2} & $68^{*}$ & 32 & 97
      & 2.560 & 2.759 & 2.776
      & 56.0 & 73.5 & 61.1 & 37.6 & 60.9 & \rankthird{74.9} \\

    RT-DETRv4-L$^{\ddagger}$~\citep{liao2025rtdetrv4} & $58^{*}$ & 31 & 91
      & 2.264 & 2.450 & 2.464
      & 55.4 & 73.0 & 60.3 & 37.1 & 60.1 & 72.9 \\

    LW-DETR-L~\citep{chen2024lwdetr} & 60 & 47 & 72
      & 2.010 & 2.152 & 2.174
      & 56.1 & 74.6 & 60.9 & 37.2 & 60.4 & 73.0 \\

    D-FINE-L~\citep{peng2025dfine} & -- & 31 & 91
      & 2.253 & 2.452 & 2.462
      & \ranksecond{57.1} & \rankthird{74.7} & \ranksecond{62.0}
      & \ranksecond{40.0} & \rankthird{61.5} & 74.2 \\

    RF-DETR-L~\citep{robinson2025rf} & -- & 34 & 126
      & 1.805 & 1.952 & \rankthird{1.968}
      & 56.5 & \ranksecond{75.1} & 61.3
      & \rankthird{39.0} & 61.0 & 73.9 \\

    YOLO26-L~\citep{yolo26} & 60 & 25 & 86
      & 1.953 & 1.969 & \ranksecond{1.967}
      & 54.3 & 71.5 & 59.4 & 37.8 & 58.6 & 70.3 \\

    ECDet-L$^{\ddagger}$~\citep{liu2026edgecrafter} & 50 & 31 & 101
      & 2.493 & 2.713 & 2.730
      & \rankthird{57.0} & \ranksecond{75.1} & \rankthird{61.7}
      & 38.7 & \ranksecond{62.5} & \ranksecond{75.0} \\

    \textbf{GTR-L (ours)}$^{\ddagger}$ & 30 & 37.2 & 106
      & 1.902 & 1.908 & \rankfirst{1.908}
      & 55.5 & 73.7 & 60.3 & 36.8 & 61.1 & 74.8 \\

    \textbf{GTR-L (ours)} & 30 & 37.2 & 106
      & 1.902 & 1.908 & \rankfirst{1.908}
      & \rankfirst{58.9} & \rankfirst{76.8} & \rankfirst{64.3}
      & \rankfirst{42.4} & \rankfirst{64.0} & \rankfirst{76.0} \\

    \midrule

    YOLOv9-E$^{\ddagger}$~\citep{wang2024yolov9} & 500 & 57 & 189
      & 3.960 & 4.633 & 4.598
      & 55.6 & 72.8 & 60.6 & 40.2 & 61.0 & 71.4 \\

    YOLOv10-X$^{\ddagger}$~\citep{wang2024yolov10} & 500 & 30 & 160
      & 2.462 & 2.484 & \ranksecond{2.482}
      & 54.4 & 71.3 & 59.3 & 37.0 & 59.8 & 70.9 \\

    YOLO11-X$^{\ddagger}$~\citep{yolo11} & 500 & 57 & 195
      & 3.331 & 3.771 & 3.747
      & 54.7 & 71.6 & 59.5 & 37.7 & 59.7 & 70.2 \\

    YOLOv12-X-Turbo$^{\ddagger}$~\citep{tian2025yolov12} & 600 & 59 & 185
      & 4.350 & 4.791 & 4.769
      & 55.4 & 72.5 & 60.3 & 38.9 & 60.8 & 70.9 \\

    RT-DETRv2-X$^{\ddagger}$~\citep{lv2024rt} & 72 & 76 & 259
      & 3.209 & 3.260 & 3.259
      & 54.3 & 72.8 & 58.8 & 35.8 & 58.8 & 72.1 \\

    DEIM-X$^{\ddagger}$~\citep{huang2025deim} & $58^{*}$ & 62 & 202
      & 3.336 & 3.388 & 3.384
      & 56.5 & 74.0 & 61.5 & 38.8 & 61.4 & 74.2 \\

    DEIMv2-X$^{\ddagger}$~\citep{huang2025deimv2} & $58^{*}$ & 50 & 152
      & 3.218 & 3.288 & 3.276
      & 57.8 & 75.3 & 63.2 & 39.1 & 62.9 & 75.9 \\

    RT-DETRv4-X$^{\ddagger}$~\citep{liao2025rtdetrv4} & $58^{*}$ & 62 & 202
      & 3.332 & 3.367 & 3.363
      & 57.0 & 74.6 & 62.1 & 39.5 & 61.9 & 74.8 \\

    LW-DETR-X~\citep{chen2024lwdetr} & 60 & 118 & 174
      & 3.643 & 3.689 & 3.686
      & 58.3 & \rankthird{76.9} & 63.3
      & 40.9 & 63.3 & 74.8 \\

    D-FINE-X~\citep{peng2025dfine} & -- & 62 & 202
      & 3.331 & 3.397 & 3.385
      & \ranksecond{59.3} & 76.8 & \ranksecond{64.6}
      & \rankfirst{42.3} & \ranksecond{64.2} & \rankfirst{76.4} \\

    RF-DETR-X~\citep{robinson2025rf} & -- & 126 & 300
      & 2.971 & 3.227 & 3.243
      & \rankthird{58.6} & \rankfirst{77.4} & \rankthird{63.8}
      & 40.3 & \rankthird{63.9} & \ranksecond{76.2} \\

    YOLO26-X~\citep{yolo26} & 40 & 55 & 194
      & 3.069 & 3.088 & 3.087
      & 56.9 & 74.1 & 62.1 & \rankthird{41.3} & 61.2 & 72.7 \\

    ECDet-X$^{\ddagger}$~\citep{liu2026edgecrafter} & 50 & 49 & 151
      & 2.950 & 3.025 & \rankthird{3.021}
      & 57.9 & 76.0 & 62.9 & 38.7
      & 63.4 & \rankthird{76.1} \\

    \textbf{GTR-X (ours)}$^{\ddagger}$ & 30 & 46.5 & 130.2
      & 2.104 & 2.117 & \rankfirst{2.115}
      & 56.2 & 74.5 & 61.1 & 37.3 & 61.8 & 74.9 \\

    \textbf{GTR-X (ours)} & 30 & 46.5 & 130.2
      & 2.104 & 2.117 & \rankfirst{2.115}
      & \rankfirst{59.4} & \ranksecond{77.3} & \rankfirst{64.7}
      & \ranksecond{42.1} & \rankfirst{64.6} & \rankfirst{76.4} \\

    \bottomrule
  \end{tabular}
  \end{adjustbox}
\end{table}

%% file: tables/coco_instance_segmentation_layout.tex
\begin{table}[!htbp]
  \centering
  \captionsetup{font=footnotesize}
  \caption{\textbf{Comparison with real-time instance segmentation models on COCO \texttt{val2017}~\citep{lin2014microsoft}.}
    Baseline results follow EdgeCrafter~\citep{liu2026edgecrafter}, with latency re-measured under our unified protocol (Appendix~\ref{sec:latency_protocol}).
    \(\dagger\) denotes Objects365 pre-training~\citep{shao2019objects365} with SAM2 pseudo masks~\citep{ravi2025sam} (\(35\times\) COCO boxes).
    Dark-to-light blue highlights the top three results within each model-size group.}

  \captionsetup{font=tiny}
  \label{tab:coco_instance_segmentation}
  \tiny
  \setlength{\tabcolsep}{1.75pt}
  \renewcommand{\arraystretch}{1.03}
  \begin{adjustbox}{width=\textwidth}
  \begin{tabular}{@{}lccccccccccc@{}}
    \toprule
    \multirow{2}{*}{Model}
      & \multirow{2}{*}{\makecell{\#Params\\(M)}}
      & \multirow{2}{*}{GFLOPs}
      & \multicolumn{3}{c}{Latency (ms) $\downarrow$}
      & \multicolumn{6}{c}{Mask AP $\uparrow$} \\    \cmidrule(lr){4-6}\cmidrule(lr){7-12}
      & & & Min. & Mean & Median
      & AP$^{val}$ & AP$^{val}_{50}$ & AP$^{val}_{75}$
      & AP$^{val}_{S}$ & AP$^{val}_{M}$ & AP$^{val}_{L}$ \\    \midrule

    YOLO26-Seg-S$^{\dagger}$~\citep{yolo26}
      & 10.4 & 34.2
      & 1.151 & 1.194 & \rankfirst{1.163}
      & 40.0 & 61.5 & 43.0 & \rankthird{21.0} & 44.5 & \rankthird{57.3} \\

    RF-DETR-Seg-S$^{\dagger}$~\citep{robinson2025rf}
      & 33.7 & 70.6
      & 1.339 & 1.427 & \ranksecond{1.445}
      & \ranksecond{43.1} & \ranksecond{66.2} & \rankthird{45.9}
      & \ranksecond{21.9} & \rankfirst{48.5} & \ranksecond{64.1} \\

    ECInsSeg-S~\citep{liu2026edgecrafter}
      & 10.3 & 33.1
      & 1.626 & 1.739 & 1.755
      & \rankthird{43.0} & \rankthird{65.7} & \ranksecond{46.0}
      & 20.8 & \rankthird{46.3} & \rankfirst{65.9} \\

    \textbf{GTR-S (ours)}
      & 12.6 & 46.8
      & 1.455 & 1.465 & \rankthird{1.465}
      & \rankfirst{45.0} & \rankfirst{67.9} & \rankfirst{48.3}
      & \rankfirst{23.8} & \ranksecond{48.4} & \rankfirst{65.9} \\

    \midrule

    YOLO26-Seg-M$^{\dagger}$~\citep{yolo26}
      & 23.6 & 121.5
      & 1.992 & 2.006 & \rankthird{2.004}
      & 44.1 & 66.8 & 47.7 & \ranksecond{25.6} & 48.9 & 60.2 \\

    RF-DETR-Seg-M$^{\dagger}$~\citep{robinson2025rf}
      & 35.7 & 102.0
      & 1.548 & 1.647 & \rankfirst{1.672}
      & \ranksecond{45.3} & \ranksecond{68.4} & \ranksecond{48.8}
      & \rankthird{25.5} & \ranksecond{50.4} & \rankthird{65.3} \\

    ECInsSeg-M~\citep{liu2026edgecrafter}
      & 20.1 & 64.2
      & 2.077 & 2.252 & 2.252
      & \rankthird{45.2} & \rankthird{68.2} & \rankthird{48.3}
      & 22.9 & \rankthird{49.0} & \ranksecond{68.1} \\

    \textbf{GTR-M (ours)}
      & 23.6 & 84.2
      & 1.797 & 1.807 & \ranksecond{1.807}
      & \rankfirst{47.7} & \rankfirst{71.3} & \rankfirst{51.6}
      & \rankfirst{27.7} & \rankfirst{51.3} & \rankfirst{69.3} \\

    \midrule

    YOLO26-Seg-L$^{\dagger}$~\citep{yolo26}
      & 28.0 & 139.8
      & 2.574 & 2.598 & \rankthird{2.597}
      & \rankthird{45.5} & 68.7 & 49.2 & \rankthird{27.1} & 50.4 & 62.8 \\

    RF-DETR-Seg-L$^{\dagger}$~\citep{robinson2025rf}
      & 36.2 & 151.1
      & 1.907 & 2.033 & \rankfirst{2.055}
      & \ranksecond{47.1} & \rankthird{70.5} & \ranksecond{50.9}
      & \ranksecond{28.4} & \ranksecond{52.1} & \rankthird{65.6} \\

    ECInsSeg-L~\citep{liu2026edgecrafter}
      & 33.6 & 110.8
      & 2.662 & 2.862 & 2.869
      & \ranksecond{47.1} & \ranksecond{70.9} & \rankthird{50.5}
      & 24.8 & \rankthird{51.1} & \ranksecond{69.6} \\

    \textbf{GTR-L (ours)}
      & 38.1 & 127.6
      & 2.241 & 2.252 & \ranksecond{2.250}
      & \rankfirst{49.5} & \rankfirst{73.5} & \rankfirst{53.6}
      & \rankfirst{28.5} & \rankfirst{53.5} & \rankfirst{71.3} \\

    \midrule

    YOLO26-Seg-X$^{\dagger}$~\citep{yolo26}
      & 62.8 & 313.5
      & 4.096 & 4.143 & 4.139
      & 47.0 & \rankthird{70.8} & 51.1 & \ranksecond{29.7} & 51.8 & 63.1 \\

    RF-DETR-Seg-X$^{\dagger}$~\citep{robinson2025rf}
      & 38.1 & 260.0
      & 3.064 & 3.299 & \rankthird{3.311}
      & \ranksecond{48.8} & \ranksecond{72.2} & \ranksecond{53.1}
      & \rankfirst{30.6} & \ranksecond{53.3} & \rankthird{65.9} \\

    ECInsSeg-X~\citep{liu2026edgecrafter}
      & 49.9 & 168.1
      & 3.125 & 3.147 & \ranksecond{3.140}
      & \rankthird{48.4} & \ranksecond{72.2} & \rankthird{52.0}
      & 26.3 & \rankthird{52.7} & \ranksecond{71.1} \\

    \textbf{GTR-X (ours)}
      & 47.4 & 151.6
      & 2.458 & 2.473 & \rankfirst{2.472}
      & \rankfirst{49.8} & \rankfirst{74.2} & \rankfirst{53.8}
      & \rankthird{28.5} & \rankfirst{53.7} & \rankfirst{71.4} \\

    \bottomrule
  \end{tabular}
  \end{adjustbox}
\end{table}

%% file: tables/coco_pose.tex
\begin{table}[H]
  \centering
  \caption{\textbf{Complete comparison with human pose estimators on COCO~\citep{lin2014microsoft}
  \texttt{val2017}.}
  Non-latency baseline values follow EdgeCrafter~\citep{liu2026edgecrafter};
  latencies use our unified protocol.
  $\dagger$ denotes additional Objects365 supervision~\citep{shao2019objects365}.
  Dark-to-light blue marks the first three ranks for median latency and
  accuracy; ties share a rank, and rankings use unrounded values.}
  \label{tab:coco_pose_full}

  \tiny
  \setlength{\tabcolsep}{2.15pt}
  \renewcommand{\arraystretch}{1.03}

  \begin{adjustbox}{width=\textwidth}
  \begin{tabular}{@{}lccccccccccc@{}}
    \toprule
    \multirow{2}{*}{Model}
      & \multirow{2}{*}{\makecell{\#Params\\(M)}}
      & \multirow{2}{*}{GFLOPs}
      & \multicolumn{3}{c}{Latency (ms) $\downarrow$}
      & \multicolumn{6}{c}{Keypoint metrics $\uparrow$} \\
    \cmidrule(lr){4-6}\cmidrule(lr){7-12}
      & & & Min. & Mean & Median
      & AP$^{val}$ & AP$^{val}_{50}$ & AP$^{val}_{75}$
      & AP$^{val}_{M}$ & AP$^{val}_{L}$ & AR$^{val}$ \\
    \midrule

    RTMO-S~\citep{rtmo} & 9.9 & 30.7
      & 0.645 & 0.648 & \rankfirst{0.646}
      & \rankthird{67.7} & \rankthird{87.8} & \rankthird{73.7}
      & -- & -- & 71.5 \\
    YOLO11-Pose-S~\citep{yolo11} & 9.9 & 23.2
      & 0.701 & 0.704 & \ranksecond{0.703}
      & 58.9 & 86.3 & 64.8 & 54.0 & 68.0 & 66.1 \\
    YOLO26-Pose-S~\citep{yolo26} & 10.4 & 23.9
      & 0.811 & 0.814 & \rankthird{0.813}
      & 63.1 & 86.6 & 68.8 & 56.5 & \rankthird{73.7} & 69.0 \\
    DETRPose-S$^{\dagger}$~\citep{janampa2025detrpose} & 11.5 & 33.1
      & 1.215 & 1.219 & 1.218
      & 67.0 & 87.6 & 72.8
      & \rankthird{60.2} & \ranksecond{77.4} & \rankthird{73.5} \\
    ECPose-S~\citep{liu2026edgecrafter} & 9.9 & 30.4
      & 1.516 & 1.522 & 1.519
      & \ranksecond{68.9} & \ranksecond{89.1} & \ranksecond{75.2}
      & \ranksecond{60.7} & \rankfirst{81.1} & \ranksecond{74.6} \\
    \textbf{GTR-S (ours)} & 11.9 & 37.0
      & 1.447 & 1.455 & 1.455
      & \rankfirst{70.1} & \rankfirst{89.7} & \rankfirst{76.8}
      & \rankfirst{62.6} & \rankfirst{81.1} & \rankfirst{76.1} \\
    \midrule

    RTMO-M~\citep{rtmo} & 22.6 & 69
      & 1.147 & 1.153 & \rankfirst{1.153}
      & \rankthird{70.9} & 89.0 & \rankthird{77.8}
      & -- & -- & 74.7 \\
    YOLO11-Pose-M~\citep{yolo11} & 20.9 & 71.7
      & 1.220 & 1.227 & \ranksecond{1.227}
      & 64.9 & 89.4 & 72.4 & 62.2 & 71.6 & 72.2 \\
    YOLO26-Pose-M~\citep{yolo26} & 21.5 & 73.1
      & 1.350 & 1.355 & \rankthird{1.353}
      & 68.8 & \rankthird{89.6} & 75.5
      & \rankthird{64.0} & 77.2 & 74.6 \\
    DETRPose-M$^{\dagger}$~\citep{janampa2025detrpose} & 20.8 & 67.3
      & 1.846 & 1.853 & 1.851
      & 69.4 & 89.2 & 75.4 & 63.2
      & \rankthird{79.0} & \rankthird{75.5} \\
    ECPose-M~\citep{liu2026edgecrafter} & 19.8 & 62.8
      & 2.099 & 2.110 & 2.109
      & \ranksecond{72.4} & \ranksecond{90.9} & \ranksecond{78.6}
      & \ranksecond{65.2} & \ranksecond{83.6} & \ranksecond{78.2} \\
    \textbf{GTR-M (ours)} & 22.8 & 69.8
      & 1.849 & 1.860 & 1.860
      & \rankfirst{74.1} & \rankfirst{91.5} & \rankfirst{80.7}
      & \rankfirst{67.5} & \rankfirst{84.1} & \rankfirst{79.6} \\
    \midrule

    RTMO-L~\citep{rtmo} & 44.8 & 136.7
      & 1.769 & 1.779 & \ranksecond{1.778}
      & 72.4 & 89.9 & 78.8 & -- & -- & 76.8 \\
    YOLO11-Pose-L~\citep{yolo11} & 26.2 & 90.7
      & 1.701 & 1.708 & \rankfirst{1.705}
      & 66.1 & 89.9 & 73.6 & 63.2 & 73.1 & 73.3 \\
    YOLO26-Pose-L~\citep{yolo26} & 25.9 & 91.3
      & 1.780 & 1.788 & \rankthird{1.788}
      & 70.4 & 90.5 & 77.4 & 65.7 & 78.4 & 75.9 \\
    DETRPose-L$^{\dagger}$~\citep{janampa2025detrpose} & 32.8 & 107.1
      & 2.639 & 2.653 & 2.650
      & \rankthird{72.5} & \rankthird{90.6} & \rankthird{79.0}
      & \rankthird{66.3} & \rankthird{82.2} & \rankthird{78.7} \\
    ECPose-L~\citep{liu2026edgecrafter} & 34.3 & 111.7
      & 2.768 & 2.783 & 2.783
      & \ranksecond{73.5} & \ranksecond{91.7} & \ranksecond{79.9}
      & \ranksecond{66.4} & \ranksecond{84.4} & \ranksecond{78.8} \\
    \textbf{GTR-L (ours)} & 38.4 & 115.5
      & 2.320 & 2.331 & 2.328
      & \rankfirst{74.7} & \rankfirst{91.9} & \rankfirst{81.6}
      & \rankfirst{68.3} & \rankfirst{84.5} & \rankfirst{79.9} \\
    \midrule

    ED-Pose~\citep{edpose} & 218 & 422.6
      & 10.266 & 10.293 & 10.292
      & \rankthird{74.3} & 91.5 & \ranksecond{81.7}
      & \ranksecond{68.5} & \rankthird{82.7} & -- \\
    YOLO11-Pose-X~\citep{yolo11} & 58.8 & 203.3
      & 2.765 & 2.777 & \ranksecond{2.777}
      & 69.5 & 91.1 & 77.4 & 66.6 & 76.0 & 76.3 \\
    YOLO26-Pose-X~\citep{yolo26} & 57.6 & 201.7
      & 2.774 & 2.790 & \rankthird{2.788}
      & 71.6 & \rankthird{91.6} & 78.9 & 67.4 & 79.5 & 77.2 \\
    DETRPose-X$^{\dagger}$~\citep{janampa2025detrpose} & 73.3 & 239.5
      & 4.129 & 4.147 & 4.145
      & 73.3 & 90.5 & 79.4 & 67.5
      & \rankthird{82.7} & \rankthird{79.4} \\
    ECPose-X~\citep{liu2026edgecrafter} & 50.6 & 172.2
      & 3.275 & 3.286 & 3.281
      & \ranksecond{74.8} & \ranksecond{92.2} & \rankthird{81.5}
      & \rankthird{68.0} & \rankfirst{85.4} & \ranksecond{80.1} \\
    \textbf{GTR-X (ours)} & 47.6 & 142.9
      & 2.573 & 2.588 & \rankfirst{2.590}
      & \rankfirst{75.5} & \rankfirst{92.4} & \rankfirst{81.8}
      & \rankfirst{69.1} & \ranksecond{85.3} & \rankfirst{80.7} \\
    \bottomrule
  \end{tabular}
  \end{adjustbox}
\end{table}

%% file: tables/dota_obb_full.tex
\begin{table}[H]
  \centering
  \captionsetup{font=footnotesize}
  \caption{\textbf{Complete oriented object detection comparison on the
  DOTA-v1.0~\citep{xia2018dota} test set.}
  We report overall and available per-class AP at an IoU threshold of $0.50$.
  Methods use multi-scale training and testing unless marked $^{\dagger}$,
  which denotes single-scale training and testing. YOLO26 accuracy and
  complexity follow its official end-to-end results; RiO-DETR values are
  taken from~\citet{hu2026rio}. Latencies use our RTX~4090 protocol;
  dashes denote unreported entries. Within each detector
  family, dark and light blue mark the first and second ranks, respectively,
  for median latency and accuracy; rankings are determined at the reported
  precision, and ties share a rank.}
  \label{tab:dota_obb_full}

  \tiny
  \setlength{\tabcolsep}{1.15pt}
  \renewcommand{\arraystretch}{1.03}

  \begin{adjustbox}{width=\textwidth}
  \begin{tabular}{@{}llccccccccccccccccccccc@{}}
    \toprule
    \multirow{2}{*}{Model}
      & \multirow{2}{*}{Backbone}
      & \multirow{2}{*}{\makecell{\#Params\\(M)}}
      & \multirow{2}{*}{GFLOPs}
      & \multicolumn{3}{c}{Latency (ms) $\downarrow$}
      & \multirow{2}{*}{AP$_{50}$ $\uparrow$}
      & \multicolumn{15}{c}{Per-class AP$_{50}$ $\uparrow$} \\
    \cmidrule(lr){5-7}\cmidrule(lr){9-23}
      & & & & Min. & Mean & Median & & PL & BD & BR & GTF & SV & LV
      & SH & TC & BC & ST & SBF & RA & HA & SP & HC \\
    \midrule

    \multicolumn{23}{l}{\textit{CNN-based oriented object detectors}} \\
    RTMDet-R-m~\citep{lyu2022rtmdet} & CSPNeXt-m & 24.7 & 100
      & 2.356 & 2.364 & 2.363 & 80.3
      & \ranksecond{87.1} & \rankfirst{85.8} & \ranksecond{56.3}
      & \ranksecond{80.3} & \ranksecond{80.0} & \ranksecond{84.7}
      & \rankfirst{88.2} & \rankfirst{90.9} & \rankfirst{88.5}
      & \rankfirst{87.6} & \rankfirst{70.7} & \ranksecond{70.0}
      & \ranksecond{78.4} & \ranksecond{80.9} & \ranksecond{74.5} \\
    RTMDet-R-l~\citep{lyu2022rtmdet} & CSPNeXt-l & 52.3 & 205
      & 3.663 & 3.673 & 3.671 & 80.5
      & \rankfirst{88.4} & \ranksecond{85.0} & \rankfirst{57.3}
      & \rankfirst{80.5} & \rankfirst{80.6} & \rankfirst{84.9}
      & \ranksecond{88.1} & \rankfirst{90.9} & \ranksecond{86.3}
      & \rankfirst{87.6} & \ranksecond{69.3} & \rankfirst{70.6}
      & \rankfirst{78.6} & \rankfirst{81.0} & \rankfirst{79.2} \\
    YOLO26n-obb~\citep{yolo26} & YOLO26n & 2.4 & 14.8
      & 0.691 & 0.694 & \rankfirst{0.693} & 78.9
      & - & - & - & - & - & -
      & - & - & - & - & - & -
      & - & - & - \\
    YOLO26s-obb~\citep{yolo26} & YOLO26s & 9.8 & 56.7
      & 1.092 & 1.099 & \ranksecond{1.099} & 80.9
      & - & - & - & - & - & -
      & - & - & - & - & - & -
      & - & - & - \\
    YOLO26m-obb~\citep{yolo26} & YOLO26m & 21.2 & 184.9
      & 2.082 & 2.093 & 2.093 & 81.0
      & - & - & - & - & - & -
      & - & - & - & - & - & -
      & - & - & - \\
    YOLO26l-obb~\citep{yolo26} & YOLO26l & 25.6 & 232.4
      & 2.700 & 2.709 & 2.707 & \ranksecond{81.6}
      & - & - & - & - & - & -
      & - & - & - & - & - & -
      & - & - & - \\
    YOLO26x-obb~\citep{yolo26} & YOLO26x & 57.6 & 520.1
      & 5.073 & 5.087 & 5.085 & \rankfirst{81.7}
      & - & - & - & - & - & -
      & - & - & - & - & - & -
      & - & - & - \\
    \midrule

    \multicolumn{23}{l}{\textit{DETR-based oriented object detectors}} \\
    RHINO-DETR$^{\dagger}$~\citep{lee2025rhino} & R-50 & 47.6 & 566
      & 9.287 & 9.319 & 9.314 & 78.7
      & 88.2 & 85.1 & 55.8 & 72.7 & 80.2 & 83.1
      & 89.0 & \ranksecond{90.8} & 87.1 & 86.8 & 65.3 & 71.6
      & 77.7 & 81.2 & 64.7 \\
    RHINO-DETR$^{\dagger}$~\citep{lee2025rhino} & Swin-T & 50.8 & 609
      & 10.345 & 10.368 & 10.365 & 79.4
      & 88.2 & 84.8 & 58.5 & 77.7 & 81.1 & 85.6
      & \rankfirst{89.2} & \rankfirst{90.9} & 87.0 & 86.4 & 65.5 & 71.3
      & 78.2 & \ranksecond{82.8} & 64.3 \\
    Oriented-DETR$^{\dagger}$~\citep{zhao2024oriented} & R-50 & 57.2 & 302
      & 14.904 & 14.929 & 14.927 & 79.1
      & 89.2 & 86.4 & 57.7 & 75.3 & 81.1 & 84.7
      & \ranksecond{89.1} & \rankfirst{90.9} & 86.1 & 87.0 & 59.5 & 70.3
      & 79.3 & 81.5 & 68.8 \\
    Oriented-DETR$^{\dagger}$~\citep{zhao2024oriented} & Swin-T & 57.7 & 309
      & 15.462 & 15.492 & 15.492 & 79.8
      & \ranksecond{89.4} & 85.1 & 57.8 & 75.0 & 81.2 & 86.1
      & \ranksecond{89.1} & \rankfirst{90.9} & \ranksecond{88.7} & 87.0 & 62.9 & 69.1
      & \ranksecond{80.7} & \ranksecond{82.8} & 71.0 \\
    RiO-DETR-n$^{\dagger}$~\citep{hu2026rio} & HGNetv2-B0 & 4.0 & 17
      & - & - & - & 78.4
      & 87.3 & 86.1 & 54.9 & 73.6 & 80.5 & 85.1
      & 88.0 & \ranksecond{90.8} & 87.4 & 86.9 & 60.9 & 72.0
      & 76.8 & 73.4 & 71.8 \\
    RiO-DETR-s$^{\dagger}$~\citep{hu2026rio} & HGNetv2-B0 & 8.2 & 53
      & - & - & - & 80.3
      & 85.6 & 84.8 & 57.5 & 75.5 & 80.7 & 86.2
      & 89.0 & \ranksecond{90.8} & 88.4 & 88.2 & 65.2 & 74.2
      & 78.3 & 80.1 & 79.6 \\
    RiO-DETR-m$^{\dagger}$~\citep{hu2026rio} & HGNetv2-B2 & 18.6 & 158
      & - & - & - & 80.9
      & 85.9 & 86.4 & 61.7 & 78.1 & 82.0 & \rankfirst{86.8}
      & \rankfirst{89.2} & \ranksecond{90.8} & \rankfirst{88.9} & 87.7 & 67.4 & 73.3
      & 78.8 & 77.4 & 78.8 \\
    RiO-DETR-l$^{\dagger}$~\citep{hu2026rio} & HGNetv2-B4 & 27.5 & 230
      & - & - & - & \ranksecond{81.7}
      & 86.7 & 86.8 & 60.8 & 79.4 & 82.0 & 86.4
      & 89.0 & \ranksecond{90.8} & \ranksecond{88.7} & \ranksecond{88.5} & 70.0 & \rankfirst{76.0}
      & 79.2 & \rankfirst{83.4} & 76.9 \\
    RiO-DETR-x$^{\dagger}$~\citep{hu2026rio} & HGNetv2-B5 & 62.5 & 527
      & - & - & - & \rankfirst{81.8}
      & 88.3 & \rankfirst{87.5} & \ranksecond{61.9} & 79.2 & \rankfirst{83.2} & \ranksecond{86.5}
      & \rankfirst{89.2} & \ranksecond{90.8} & 88.4 & 87.8 & 71.2 & 75.1
      & 78.5 & 82.2 & 77.1 \\
    RiO-DETR-m~\citep{hu2026rio} & HGNetv2-B2 & 18.6 & 158
      & - & - & - & 81.5
      & 87.3 & 87.2 & 61.7 & \rankfirst{81.8} & 82.5 & 86.2
      & \rankfirst{89.2} & 90.6 & \ranksecond{88.7} & \rankfirst{88.7} & \rankfirst{74.5} & 71.4
      & 78.8 & 74.4 & \ranksecond{79.7} \\
    RiO-DETR-x~\citep{hu2026rio} & HGNetv2-B5 & 62.5 & 527
      & - & - & - & \rankfirst{81.8}
      & 83.8 & \ranksecond{87.3} & \rankfirst{62.6} & \ranksecond{79.9} & \ranksecond{82.7} & 86.3
      & \ranksecond{89.1} & 90.6 & 88.5 & 88.0 & \ranksecond{72.5} & \ranksecond{75.2}
      & 78.9 & 81.8 & 79.2 \\
    \textbf{GTR-S (ours)} & GLA & 12.1 & 82.8
      & 1.920 & 1.943 & \rankfirst{1.946} & 80.1
      & 89.2 & 83.9 & 58.7 & 77.5 & 82.6 & 86.1
      & 88.5 & \ranksecond{90.8} & 87.8 & 87.7 & 61.4 & 68.6
      & 78.5 & 79.8 & \rankfirst{81.7} \\

    \textbf{GTR-X (ours)} & GLA & 46.3 & 324
      & 3.927 & 3.961 & \ranksecond{3.960} & 81.3
      & \rankfirst{89.5} & 85.1 & 59.6 & 78.0 & 82.6 & 86.4
      & 88.6 & \rankfirst{90.9} & 87.9 & 87.6 & 67.2 & 73.3
      & \rankfirst{84.5} & 80.1 & 77.9 \\
    \bottomrule
  \end{tabular}
  \end{adjustbox}
\end{table}

%% file: tables/cityscapes_semantic.tex
\begin{table}[H]
  \centering
  \caption{\textbf{Complete semantic segmentation comparison on the Cityscapes~\citep{cordts2016cityscapes} validation set.}
Latency is measured in FP16 with batch size 1. Dark blue indicates the better value within each matched model-size pair; ties share the same rank.}
  \label{tab:cityscapes_semantic_full}
  \small
  \setlength{\tabcolsep}{5.0pt}
  \renewcommand{\arraystretch}{1.04}
  \begin{tabular}{@{}lcccccc@{}}
    \toprule
    \multirow{2}{*}{Model}
      & \multirow{2}{*}{\makecell{\#Params\\(M)}}
      & \multirow{2}{*}{GFLOPs}
      & \multicolumn{3}{c}{Latency (ms) $\downarrow$}
      & \multirow{2}{*}{mIoU $\uparrow$} \\
    \cmidrule(lr){4-6}
      & & & Min. & Mean & Median & \\
    \midrule
    YOLO26s-sem~\citep{yolo26} & 6.5 & 88.8
      & 0.752 & 0.758 & \rankfirst{0.758} & 80.8 \\
    \textbf{GTR-S (ours)} & 6.9 & 82.8
      & 1.489 & 1.495 & 1.495 & \rankfirst{81.5} \\
    \midrule
    YOLO26m-sem~\citep{yolo26} & 14.3 & 304.5
      & 1.601 & 1.617 & \rankfirst{1.617} & 82.0 \\
    \textbf{GTR-M (ours)} & 12.9 & 153.0
      & 1.961 & 1.971 & 1.970 & \rankfirst{83.0} \\
    \midrule
    YOLO26l-sem~\citep{yolo26} & 17.9 & 384.7
      & 2.155 & 2.167 & \rankfirst{2.167} & 82.9 \\
    \textbf{GTR-L (ours)} & 25.0 & 258.7
      & 2.922 & 2.958 & 2.958 & \rankfirst{83.2} \\
    \midrule
    YOLO26x-sem~\citep{yolo26} & 40.2 & 861.7
      & 4.291 & 4.326 & 4.325 & \rankfirst{83.6} \\
    \textbf{GTR-X (ours)} & 32.2 & 317.1
      & 3.442 & 3.483 & \rankfirst{3.482} & \rankfirst{83.6} \\
    \bottomrule
  \end{tabular}
\end{table}

%% file: tables/nyu_depth_aligned.tex
\begin{table}[H]
  \centering
  \caption{\textbf{Complete ground-truth-aligned monocular depth comparison on the NYU Depth
  V2~\citep{silberman2012indoor} Eigen test split~\citep{eigen2014depth}.}
  ZipDepth is re-evaluated using the same filled ground-truth data as YOLO26 and GTR.
  YOLO26 and GTR scores use six test-time views and a per-image log-affine fit to ground truth,
  without NYU-specific fine-tuning (Appendix~\ref{sec:metric_depth_details}).
  Latency measures a single forward pass. Dark blue
  marks the better value within each matched YOLO26/GTR pair for median latency
  and each accuracy metric.}
  \label{tab:nyu_depth_full}
  \small
  \setlength{\tabcolsep}{4.2pt}
  \renewcommand{\arraystretch}{1.04}
  \begin{adjustbox}{width=\textwidth}
  \begin{tabular}{@{}lcccccccc@{}}
    \toprule
    \multirow{2}{*}{Model}
      & \multirow{2}{*}{\makecell{\#Params\\(M)}}
      & \multirow{2}{*}{GFLOPs}
      & \multicolumn{3}{c}{Latency (ms) $\downarrow$}
      & \multirow{2}{*}{$\delta_1$ $\uparrow$}
      & \multirow{2}{*}{AbsRel $\downarrow$}
      & \multirow{2}{*}{RMSE $\downarrow$} \\
    \cmidrule(lr){4-6}
      & & & Min. & Mean & Median & & & \\
    \midrule
    ZipDepth~\citep{tosi2026zipdepth} & 6.1 & 7.9
      & 0.332 & 0.333 & 0.334
      & 0.919 & 0.091 & 0.385 \\
    \midrule
    YOLO26s-depth~\citep{yolo26} & 13.2 & 67.9
      & 0.847 & 0.850 & \rankfirst{0.850}
      & 0.896 & 0.104 & 0.399 \\
    \textbf{GTR-S (ours)} & 11.5 & 65.0
      & 1.290 & 1.297 & 1.296
      & \rankfirst{0.946} & \rankfirst{0.074} & \rankfirst{0.336} \\
    \midrule
    YOLO26m-depth~\citep{yolo26} & 23.3 & 130.7
      & 1.304 & 1.308 & \rankfirst{1.307}
      & 0.921 & 0.089 & 0.364 \\
    \textbf{GTR-M (ours)} & 19.4 & 93.2
      & 1.501 & 1.509 & 1.508
      & \rankfirst{0.952} & \rankfirst{0.069} & \rankfirst{0.319} \\
    \midrule
    YOLO26l-depth~\citep{yolo26} & 27.7 & 157.2
      & 1.744 & 1.747 & \rankfirst{1.747}
      & 0.930 & 0.083 & 0.351 \\
    \textbf{GTR-L (ours)} & 33.9 & 136.6
      & 1.936 & 1.948 & 1.948
      & \rankfirst{0.951} & \rankfirst{0.069} & \rankfirst{0.328} \\
    \midrule
    YOLO26x-depth~\citep{yolo26} & 57.0 & 302.0
      & 2.721 & 2.730 & 2.730
      & 0.933 & 0.080 & 0.344 \\
    \textbf{GTR-X (ours)} & 41.1 & 159.4
      & 2.144 & 2.159 & \rankfirst{2.158}
      & \rankfirst{0.954} & \rankfirst{0.067} & \rankfirst{0.317} \\
    \bottomrule
  \end{tabular}
  \end{adjustbox}
\end{table}

%% file: tables/ablation_positional_encoding.tex
\begin{table}[H]
  \centering
  \setlength{\intextsep}{4pt}
  \captionsetup{
    font=footnotesize,
    skip=2pt,
    belowskip=0pt
  }
  \caption{\textbf{S-SwiGLU vs. learned positional embeddings on GTR-S.} S-SwiGLU uses no explicit positional embedding.}

  \label{tab:ablation_positional_encoding}
  \tiny
  \setlength{\tabcolsep}{2.6pt}
  \renewcommand{\arraystretch}{1.15}
  \begin{tabular}{@{}clcccccccc@{}}
    \toprule
    \multirow{2}{*}{\makecell{Input\\resolution}}
      & \multirow{2}{*}{Spatial mechanism}
      & \multirow{2}{*}{\makecell{\#Params\\(M)}}
      & \multirow{2}{*}{GFLOPs}
      & \multicolumn{6}{c}{COCO AP $\uparrow$} \\
    \cmidrule(lr){5-10}
      & & &
      & AP$^{val}$ & AP$^{val}_{50}$ & AP$^{val}_{75}$
      & AP$^{val}_{S}$ & AP$^{val}_{M}$ & AP$^{val}_{L}$ \\
    \midrule
      & Learned positional embedding
      & 12.1 & 47.6 & 53.9 & 71.6 & 58.6 & 38.0 & 57.8 & 70.2 \\
    \rowcolor{gtrblue!10}
    \multirow{-2}{*}{$768^2$}
      & \textbf{S-SwiGLU (ours)}
      & 12.1 & 47.8
      & \textbf{55.1} & \textbf{72.5} & \textbf{60.2}
      & \textbf{38.7} & \textbf{59.6} & \textbf{70.7} \\
    \midrule
      & Learned positional embedding
      & 12.1 & 83.2 & 55.3 & 72.7 & 60.3 & 40.6 & 58.7 & \textbf{70.7} \\
    \rowcolor{gtrblue!10}
    \multirow{-2}{*}{$1024^2$}
      & \textbf{S-SwiGLU (ours)}
      & 12.1 & 83.7
      & \textbf{56.6} & \textbf{73.9} & \textbf{61.9}
      & \textbf{42.1} & \textbf{60.8} & 70.4 \\
    \bottomrule
  \end{tabular}
\end{table}

%% file: tables/ablation_scan_direction_full.tex
\begin{table}[H]
 \centering

 \setlength{\intextsep}{4pt}

 \captionsetup{
   font=footnotesize,
   skip=2pt,              
   belowskip=0pt
 }

 \caption{\textbf{Effect of scan direction and spatial mechanism on GTR-L.}
 LR/RL/TB/BT denote left-to-right, right-to-left, top-to-bottom, and bottom-to-top scans.}
 \label{tab:ablation_scan_direction}

 \tiny


 \begin{adjustbox}{width=\textwidth}
 \begin{tabular}{@{}lcccccccc@{}}
   \toprule
   \multirow{2}{*}{Variant}
     & \multirow{2}{*}{\makecell{\#Params\\(M)}}
     & \multirow{2}{*}{GFLOPs}
     & \multicolumn{6}{c}{COCO AP $\uparrow$} \\
   \cmidrule(lr){4-9}
     & &
     & AP$^{val}$ & AP$^{val}_{50}$ & AP$^{val}_{75}$
     & AP$^{val}_{S}$ & AP$^{val}_{M}$ & AP$^{val}_{L}$ \\
   \midrule
   \rowcolor{gtrblue!10}
   \textbf{Four directions (LR/RL/TB/BT) + S-SwiGLU}
     & 37.2 & 106
     & \textbf{55.5} & \textbf{73.7} & \textbf{60.3}
     & \textbf{36.8} & \textbf{61.1} & 74.8 \\
   Horizontal bidirectional (LR/RL) + S-SwiGLU
     & 37.2 & 106 & 55.1 & 73.4 & 59.8 & 36.1 & 60.5 & \textbf{75.2} \\
   Vertical bidirectional (TB/BT) + S-SwiGLU
     & 37.2 & 106 & 55.0 & 73.3 & 59.8 & 35.8 & 60.7 & 74.6 \\
   Single direction + S-SwiGLU
     & 37.2 & 106 & 54.4 & 72.6 & 59.0 & 35.1 & 59.7 & 73.9 \\
   Single direction + learned positional embedding
     & 37.2 & 106 & 45.6 & 62.3 & 49.1 & 27.5 & 49.1 & 63.5 \\
   \bottomrule
 \end{tabular}
 \end{adjustbox}
\end{table}

%% file: tables/object365_pretraining_ablation.tex
\begin{table}[H]
  \centering
  \caption{\textbf{Effect of Objects365~\citep{shao2019objects365} detector pre-training.}
  A filled circle indicates that detector pre-training on Objects365 is enabled.}
  \label{tab:objects365_ablation}
  \small
  \setlength{\tabcolsep}{5.0pt}
  \renewcommand{\arraystretch}{1.03}
  \begin{tabular}{@{}ccccccc@{}}
    \toprule
    Objects365 pre-training
      & AP$^{val}$ & AP$^{val}_{50}$ & AP$^{val}_{75}$
      & AP$^{val}_{S}$ & AP$^{val}_{M}$ & AP$^{val}_{L}$ \\
    \midrule
    --
      & 50.7 & 68.5 & 54.7 & 31.1 & 55.5 & 70.2 \\
    \rowcolor{gtrblue!10}
    $\bullet$
      & \textbf{53.6} & \textbf{71.1} & \textbf{58.3}
      & \textbf{36.4} & \textbf{58.3} & \textbf{70.2} \\
    \bottomrule
  \end{tabular}
\end{table}

%% file: tables/gtr_architecture.tex
\begin{table}[!ht]
  \centering
  \caption{\textbf{Architecture configurations of the four GTR variants.} }
  \label{tab:gtr_architecture}
  \setlength{\tabcolsep}{8pt}
  \renewcommand{\arraystretch}{1.08}
  \small
  \begin{tabular}{lcccc}
    \toprule
    Setting & GTR-S & GTR-M & GTR-L & GTR-X \\
    \midrule
    Embedding dimension & 192 & 256 & 384 & 384 \\
    GLA heads & 3 & 4 & 6 & 6 \\
    S-SwiGLU expansion ratio & 4 & 4 & 4 & 6 \\
    Encoder width & 192 & 256 & 256 & 256 \\
    Decoder FFN dimension & 512 & 1024 & 1024 & 2048 \\
    Decoder layers & 4 & 4 & 4 & 4 \\
    Object queries & 300 & 300 & 300 & 300 \\
    Training query groups & 3 & 3 & 3 & 3 \\
    \bottomrule
  \end{tabular}
\end{table}

%% file: tables/distillation.tex
\begin{table}[!ht]
    \centering
    \caption{\textbf{Training configuration for final-output representation alignment.}}
    \label{tab:distillation_recipe}
    \small
    \setlength{\tabcolsep}{5pt}
    \renewcommand{\arraystretch}{1.05}
    \begin{tabular}{lcc}
        \toprule
        Configuration & GTR-S/M & GTR-L/X \\
        \midrule
        Optimizer                         & \multicolumn{2}{c}{AdamW} \\
        Peak learning rate                & \multicolumn{2}{c}{$1.5\times10^{-3}$} \\
        Minimum learning rate             & \multicolumn{2}{c}{$1.0\times10^{-5}$} \\
        Weight decay                      & \multicolumn{2}{c}{$0.05$} \\
        Training epochs                   & \multicolumn{2}{c}{$300$} \\
        Optimizer betas                   & \multicolumn{2}{c}{$(0.9,\,0.999)$} \\
        Global batch size                 & \multicolumn{2}{c}{$2048$} \\
        Warm-up epochs                    & $10$ & $20$ \\
        Drop-path rate                    & $0.1$ & $0.2$ \\
        Layer-wise learning-rate decay    & \multicolumn{2}{c}{\xmark} \\
        Label smoothing                   & \multicolumn{2}{c}{\xmark} \\
        Random erasing                    & \multicolumn{2}{c}{\xmark} \\
        RandAugment                       & \multicolumn{2}{c}{\xmark} \\
        Repeated augmentation             & \multicolumn{2}{c}{\cmark} \\
        ThreeAugmentation                 & \multicolumn{2}{c}{\cmark} \\
        \bottomrule
    \end{tabular}
\end{table}

%% file: figs_tex/compare_distillation.tex
\begin{figure}[t]
    \centering
    \includegraphics[width=0.9\textwidth]{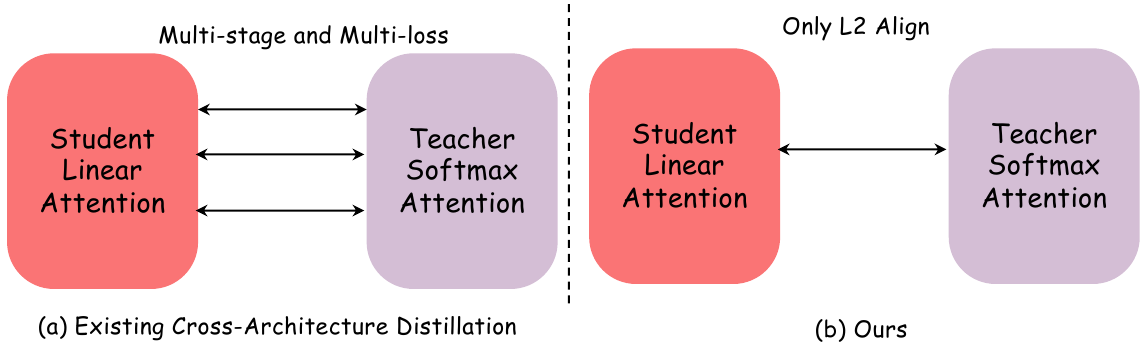}
    \caption{
    \textbf{Comparison of cross-architecture transfer designs.}
    Representative prior approaches use multiple alignment stages or losses.
    GTR instead aligns only the final representations of a frozen softmax
    teacher and a linear-attention student with a squared $\ell_2$ objective.
    }
    \label{fig:distillation_comparison}
\end{figure}

%% file: tables/obj365_pre_training.tex
\begin{table}[!ht]
  \centering
  \caption{\textbf{Objects365~\citep{shao2019objects365} detector pre-training hyperparameters.} }
  \label{tab:objects365_hparams}
  \setlength{\tabcolsep}{8pt}
  \renewcommand{\arraystretch}{1.08}
  \small
  \begin{tabular}{lcccc}
    \toprule
    Setting & GTR-S & GTR-M & GTR-L & GTR-X \\
    \midrule
    Epochs & 36 & 36 & 36 & 36 \\
    Global batch size & 128 & 128 & 128 & 128 \\
    Base LR & $2\times10^{-3}$ & $2\times10^{-3}$ & $2\times10^{-3}$ & $2\times10^{-3}$ \\
    Backbone LR & $1\times10^{-4}$ & $1\times10^{-4}$ & $2\times10^{-5}$ & $1\times10^{-5}$ \\
    Weight decay & $1\times10^{-4}$ & $1\times10^{-4}$ & $1.25\times10^{-4}$ & $1.25\times10^{-4}$ \\
    AdamW $\beta$ & 0.9/0.999 & 0.9/0.999 & 0.9/0.999 & 0.9/0.999 \\
    LR schedule & flat-cosine & flat-cosine & flat-cosine & flat-cosine \\
    Flat phase ends at epoch & 15 & 15 & 15 & 15 \\
    Minimum/base LR ratio & 0.1 & 0.1 & 0.1 & 0.1 \\
    Warm-up iterations & 2000 & 2000 & 2000 & 2000 \\
    \bottomrule
  \end{tabular}
\end{table}

%% file: tables/coco_finetune.tex
\begin{table}[!ht]
  \centering
  \caption{\textbf{COCO~\citep{lin2014microsoft} fine-tuning hyperparameters.} Bias and normalization parameters use zero weight decay in both training stages.}
  \label{tab:coco_hparams}
  \setlength{\tabcolsep}{8pt}
  \renewcommand{\arraystretch}{1.08}
  \small
  \begin{tabular}{lcccc}
    \toprule
    Setting & GTR-S & GTR-M & GTR-L & GTR-X \\
    \midrule
    Epochs & 30 & 30 & 30 & 30 \\
    Global batch size & 32 & 32 & 32 & 32 \\
    Base LR & $5\times10^{-4}$ & $5\times10^{-4}$ & $5\times10^{-4}$ & $5\times10^{-4}$ \\
    Backbone LR & $1.5\times10^{-5}$ & $1.5\times10^{-5}$ & $2.5\times10^{-6}$ & $8\times10^{-6}$ \\
    Weight decay & $1\times10^{-4}$ & $1\times10^{-4}$ & $1.25\times10^{-4}$ & $1.25\times10^{-4}$ \\
    LR schedule & flat-cosine & flat-cosine & flat-cosine & flat-cosine \\
    Flat phase ends at epoch & 6 & 6 & 6 & 6 \\
    Minimum/base LR ratio & 0.5 & 0.5 & 0.5 & 0.1 \\
    Warm-up iterations & 2000 & 2000 & 2000 & 2000 \\
    \bottomrule
  \end{tabular}
\end{table}

%% file: tables/coco_pose_training.tex
\begin{table}[!ht]
  \centering
  \caption{\textbf{Human pose estimation hyperparameters.}}
  \label{tab:pose_hparams}
  \setlength{\tabcolsep}{8pt}
  \renewcommand{\arraystretch}{1.08}
  \small
  \begin{tabular}{lcccc}
    \toprule
    Setting & GTR-S & GTR-M & GTR-L & GTR-X \\
    \midrule
    Optimizer & AdamW & AdamW & AdamW & AdamW \\
    Backbone LR & $2.5\times10^{-5}$ & $2.5\times10^{-5}$ & $2.5\times10^{-6}$ & $2.5\times10^{-6}$ \\
    Base LR & $5\times10^{-4}$ & $5\times10^{-4}$ & $5\times10^{-4}$ & $5\times10^{-4}$ \\
    Weight decay & $1\times10^{-4}$ & $1\times10^{-4}$ & $1.25\times10^{-4}$ & $1.25\times10^{-4}$ \\
    \midrule
    Epochs (with/without aug.) & 90 + 2 & 90 + 2 & 72 + 2 & 72 + 2 \\
    Prob$_\text{mosaic}$ & 0.5 & 0.5 & 0.5 & 0.5 \\
    Epochs$_\text{mosaic}$ & 45 & 45 & 48 & 48 \\
    Prob$_\text{mixup}$ & 0.25 & 0.25 & 0.25 & 0.25 \\
    Epochs$_\text{mixup}$ & 45 & 45 & 48 & 48 \\
    Prob$_\text{copypaste}$ & 0.75 & 0.75 & 0.75 & 0.75 \\
    Epochs$_\text{copypaste}$ & 45 & 45 & 48 & 48 \\
    Global batch size & 32 & 32 & 32 & 32 \\
    \midrule
    $\lambda_{\mathrm{cls}}$ & 2 & 2 & 2 & 2 \\
    $\lambda_{\mathrm{kpt}}$ & 10 & 10 & 10 & 10 \\
    $\lambda_{\mathrm{oks}}$ & 4 & 4 & 4 & 4 \\
    \bottomrule
  \end{tabular}
\end{table}

%% file: tables/cityscapes_training.tex
\begin{table}[!ht]
  \centering
  \caption{\textbf{Cityscapes~\citep{cordts2016cityscapes} semantic-segmentation fine-tuning hyperparameters.}}
  \label{tab:cityscapes_hparams}
  \setlength{\tabcolsep}{8pt}
  \renewcommand{\arraystretch}{1.08}
  \small
  \begin{tabular}{lcccc}
    \toprule
    Setting & GTR-S & GTR-M & GTR-L & GTR-X \\
    \midrule
    Epochs & 30 & 30 & 30 & 30 \\
    Global batch size & 8 & 8 & 8 & 8 \\
    Base LR & $5\times10^{-4}$ & $5\times10^{-4}$ & $5\times10^{-4}$ & $5\times10^{-4}$ \\
    Backbone LR & $1.5\times10^{-4}$ & $1.8\times10^{-4}$ & $1.2\times10^{-4}$ & $1.2\times10^{-4}$ \\
    Weight decay & $1\times10^{-4}$ & $1\times10^{-4}$ & $1.25\times10^{-4}$ & $1.25\times10^{-4}$ \\
    LR schedule & flat-cosine & flat-cosine & flat-cosine & flat-cosine \\
    Flat phase ends at epoch & 6 & 6 & 6 & 6 \\
    Minimum/base LR ratio & 0.5 & 0.5 & 0.5 & 0.5 \\
    Warm-up iterations & 2,000 & 2,000 & 2,000 & 2,000 \\
    \midrule
    Training crop size & $1024^2$ & $1024^2$ & $1024^2$ & $1024^2$ \\
    Sliding-window stride & 768 & 768 & 768 & 768 \\
    \bottomrule
  \end{tabular}
\end{table}

%% file: tables/depth_pretraining.tex
\begin{table}[!ht]
  \centering
  \caption{\textbf{Mixed-data depth pre-training hyperparameters.}}
  \label{tab:depth_pretrain_hparams}
  \setlength{\tabcolsep}{5pt}
  \renewcommand{\arraystretch}{1.08}
  \small
  \begin{tabular}{lcccc}
    \toprule
    Setting & GTR-S & GTR-M & GTR-L & GTR-X \\
    \midrule
    Epochs & 30 & 30 & 30 & 30 \\
    Global batch size & 256 & 256 & 256 & 256 \\
    Base LR & $2\times10^{-4}$ & $2\times10^{-4}$ & $2\times10^{-4}$ & $2\times10^{-4}$ \\
    Backbone LR & $1\times10^{-4}$ & $1\times10^{-4}$ & $2\times10^{-5}$ & $1\times10^{-5}$ \\
    Weight decay & $1\times10^{-4}$ & $1\times10^{-4}$ & $1.25\times10^{-4}$ & $1.25\times10^{-4}$ \\
    LR schedule & flat-cosine & flat-cosine & flat-cosine & flat-cosine \\
    Flat phase ends at epoch & 12 & 12 & 12 & 12 \\
    Minimum/base LR ratio & 0.1 & 0.1 & 0.1 & 0.1 \\
    Warm-up iterations & 2,000 & 2,000 & 2,000 & 2,000 \\
    No-augmentation phase starts at epoch & 27 & 27 & 27 & 27 \\
    \midrule
    Valid depth range (m) & $[10^{-3},100]$ & $[10^{-3},100]$ & $[10^{-3},100]$ & $[10^{-3},100]$ \\
    Output parameterization & log-depth & log-depth & log-depth & log-depth \\
    \bottomrule
  \end{tabular}
\end{table}

%% file: tables/dota_obb_training.tex
\begin{table}[!ht]
  \centering
  \caption{\textbf{DOTA-v1.0~\citep{xia2018dota} oriented-object-detection fine-tuning hyperparameters.}}
  \label{tab:dota_obb_hparams}
  \setlength{\tabcolsep}{8pt}
  \renewcommand{\arraystretch}{1.08}
  \small
  \begin{tabular}{lcc}
    \toprule
    Setting & GTR-S & GTR-X \\
    \midrule
    Epochs & 12 & 14 \\
    Global batch size & 16 & 16 \\
    Base LR & $5\times10^{-4}$ & $5\times10^{-4}$ \\
    Backbone LR & $1.8\times10^{-4}$ & $8\times10^{-6}$ \\
    Weight decay & $1\times10^{-4}$ & $1.25\times10^{-4}$ \\
    LR schedule & constant after warm-up & constant after warm-up \\
    Warm-up iterations & 2,000 & 2,000 \\
    Drop-path rate & 0.0 & 0.1 \\
    \midrule
    Training patch size & $1024^2$ & $1024^2$ \\
    Multi-scale resize ratios & $\{0.5,1.0,1.5\}$ & $\{0.5,1.0,1.5\}$ \\
    \bottomrule
  \end{tabular}
\end{table}

%% file: figs_tex/roofline_model.tex
\begin{figure*}[t]
  \centering
  \includegraphics[
    width=0.96\textwidth
  ]{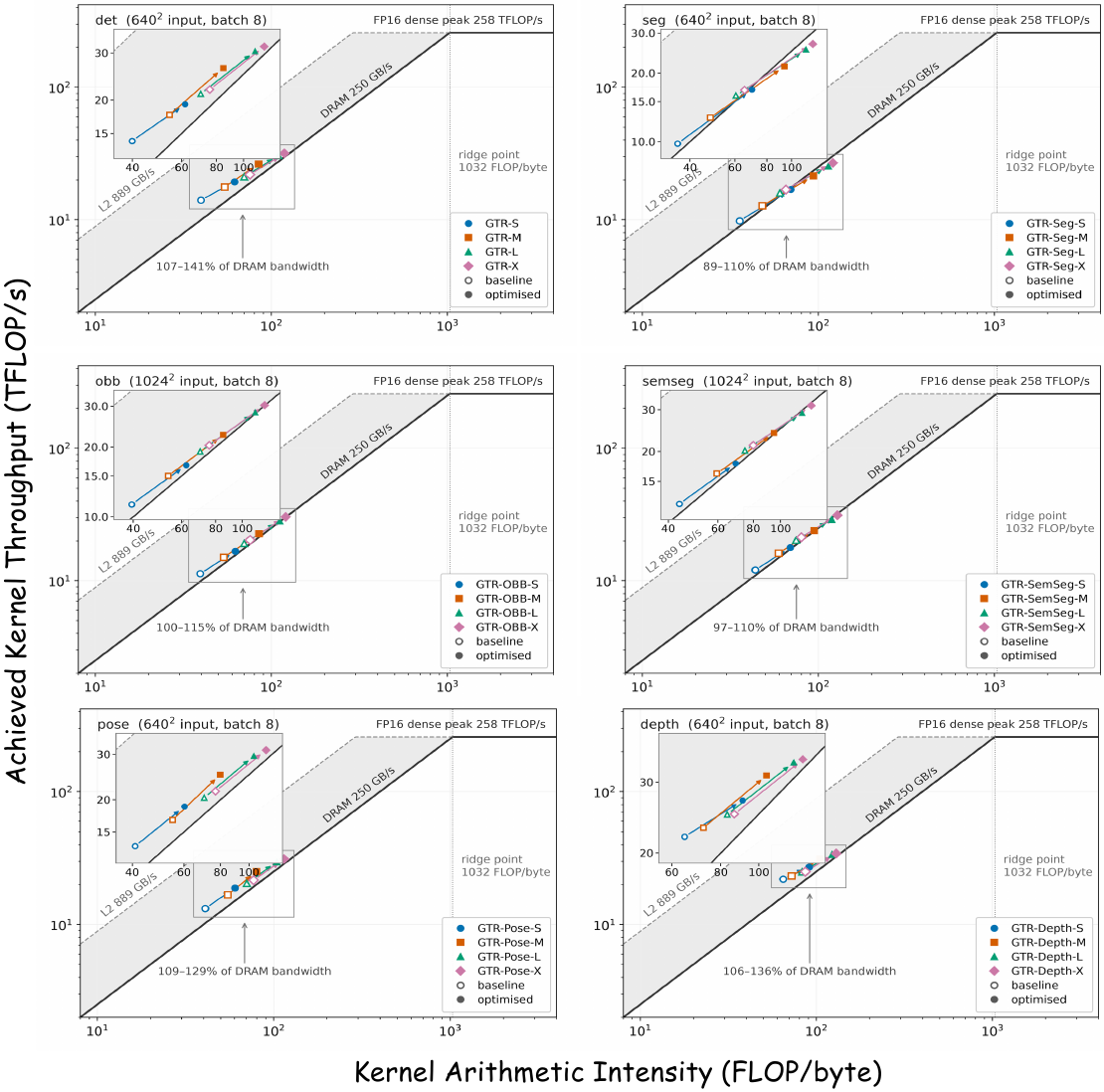}
  \caption{\textbf{Roofline analysis of the TensorRT deployment on DRIVE AGX
  Thor.} Open markers denote the original TensorRT baseline,
  which uses only TensorRT's built-in optimizations; filled markers denote our
  graph--kernel-optimized implementation.}
  \label{fig:thor_roofline}
\end{figure*}

%% file: tables/thor_deployment_bs1.tex
\begin{table*}[t]
  \centering
  \caption{\textbf{FP16 TensorRT deployment on DRIVE AGX Thor at batch size
  one.} Cosine similarity compares FP16 TensorRT outputs with FP32
  PyTorch reference outputs.}
  \label{tab:thor_bs1}
  \scriptsize
  \setlength{\tabcolsep}{7pt}
  \renewcommand{\arraystretch}{0.93}
  \begin{tabular}{@{}llcrrr@{}}
    \toprule
    Task & Model & Input & Median (ms) $\downarrow$ & Images/s $\uparrow$ & Cosine sim. $\uparrow$ \\
    \midrule
    \multirow{4}{*}{Detection}
      & GTR-S & $640^2$ & 2.282 & 438 & 0.9997 \\
      & GTR-M & $640^2$ & 2.721 & 368 & 0.9999 \\
      & GTR-L & $640^2$ & 3.527 & 284 & 0.9998 \\
      & GTR-X & $640^2$ & 4.080 & 245 & 0.9999 \\
    \addlinespace[2pt]
    \multirow{4}{*}{Instance seg.}
      & GTR-Seg-S & $640^2$ & 3.474 & 288 & 0.9992 \\
      & GTR-Seg-M & $640^2$ & 4.363 & 229 & 0.9997 \\
      & GTR-Seg-L & $640^2$ & 5.130 & 195 & 0.9993 \\
      & GTR-Seg-X & $640^2$ & 5.703 & 175 & 0.9993 \\
    \addlinespace[2pt]
    \multirow{4}{*}{Oriented det.}
      & GTR-OBB-S & $1024^2$ & 4.310 & 232 & 0.9998 \\
      & GTR-OBB-M & $1024^2$ & 5.491 & 182 & 0.9998 \\
      & GTR-OBB-L & $1024^2$ & 7.602 & 132 & 0.9999 \\
      & GTR-OBB-X & $1024^2$ & 8.769 & 114 & 0.9998 \\
    \addlinespace[2pt]
    \multirow{4}{*}{Pose}
      & GTR-Pose-S & $640^2$ & 2.459 & 407 & 0.9994 \\
      & GTR-Pose-M & $640^2$ & 3.082 & 324 & 0.9996 \\
      & GTR-Pose-L & $640^2$ & 3.909 & 256 & 0.9989 \\
      & GTR-Pose-X & $640^2$ & 4.455 & 224 & 0.9989 \\
    \addlinespace[2pt]
    \multirow{4}{*}{Semantic seg.}
      & GTR-SemSeg-S & $1024^2$ & 3.579 & 279 & 0.9993 \\
      & GTR-SemSeg-M & $1024^2$ & 4.655 & 215 & 0.9998 \\
      & GTR-SemSeg-L & $1024^2$ & 6.792 & 147 & 0.9996 \\
      & GTR-SemSeg-X & $1024^2$ & 7.916 & 126 & 0.9996 \\
    \addlinespace[2pt]
    \multirow{4}{*}{Depth}
      & GTR-Depth-S & $640^2$ & 2.675 & 374 & 1.0000 \\
      & GTR-Depth-M & $640^2$ & 3.071 & 326 & 1.0000 \\
      & GTR-Depth-L & $640^2$ & 3.872 & 258 & 1.0000 \\
      & GTR-Depth-X & $640^2$ & 4.389 & 228 & 1.0000 \\
    \bottomrule
  \end{tabular}
\end{table*}

%% file: tables/thor_deployment_bs8.tex
\begin{table*}[t]
  \centering
  \caption{\textbf{FP16 TensorRT deployment on DRIVE AGX Thor at batch size
  eight.} Cosine similarity compares FP16 TensorRT outputs with FP32
  PyTorch reference outputs.}
  \label{tab:thor_bs8}
  \scriptsize
  \setlength{\tabcolsep}{7pt}
  \renewcommand{\arraystretch}{0.93}
  \begin{tabular}{@{}llcrrr@{}}
    \toprule
    Task & Model & Input & Median (ms) $\downarrow$ & Images/s $\uparrow$ & Cosine sim. $\uparrow$ \\
    \midrule
    \multirow{4}{*}{Detection}
      & GTR-S & $640^2$ & 14.017 & 571 & 0.9998 \\
      & GTR-M & $640^2$ & 19.126 & 418 & 0.9998 \\
      & GTR-L & $640^2$ & 27.817 & 288 & 0.9999 \\
      & GTR-X & $640^2$ & 32.701 & 245 & 0.9998 \\
    \addlinespace[2pt]
    \multirow{4}{*}{Instance seg.}
      & GTR-Seg-S & $640^2$ & 22.085 & 362 & 0.9993 \\
      & GTR-Seg-M & $640^2$ & 31.529 & 254 & 0.9996 \\
      & GTR-Seg-L & $640^2$ & 40.203 & 199 & 0.9990 \\
      & GTR-Seg-X & $640^2$ & 45.142 & 177 & 0.9990 \\
    \addlinespace[2pt]
    \multirow{4}{*}{Oriented det.}
      & GTR-OBB-S & $1024^2$ & 39.619 & 202 & 0.9998 \\
      & GTR-OBB-M & $1024^2$ & 54.312 & 147 & 0.9998 \\
      & GTR-OBB-L & $1024^2$ & 74.545 & 107 & 0.9998 \\
      & GTR-OBB-X & $1024^2$ & 85.553 & 94 & 0.9998 \\
    \addlinespace[2pt]
    \multirow{4}{*}{Pose}
      & GTR-Pose-S & $640^2$ & 16.227 & 493 & 0.9993 \\
      & GTR-Pose-M & $640^2$ & 23.108 & 346 & 0.9995 \\
      & GTR-Pose-L & $640^2$ & 31.877 & 251 & 0.9993 \\
      & GTR-Pose-X & $640^2$ & 37.228 & 215 & 0.9994 \\
    \addlinespace[2pt]
    \multirow{4}{*}{Semantic seg.}
      & GTR-SemSeg-S & $1024^2$ & 36.953 & 216 & 0.9991 \\
      & GTR-SemSeg-M & $1024^2$ & 50.860 & 157 & 0.9998 \\
      & GTR-SemSeg-L & $1024^2$ & 70.900 & 113 & 0.9997 \\
      & GTR-SemSeg-X & $1024^2$ & 81.234 & 98 & 0.9994 \\
    \addlinespace[2pt]
    \multirow{4}{*}{Depth}
      & GTR-Depth-S & $640^2$ & 19.212 & 416 & 1.0000 \\
      & GTR-Depth-M & $640^2$ & 23.953 & 334 & 1.0000 \\
      & GTR-Depth-L & $640^2$ & 32.588 & 245 & 1.0000 \\
      & GTR-Depth-X & $640^2$ & 37.328 & 214 & 1.0000 \\
    \bottomrule
  \end{tabular}
\end{table*}

%% file: figs_tex/results_detection.tex
\begin{figure*}[!t]
    \centering
    \includegraphics[width=0.90\textwidth]
    {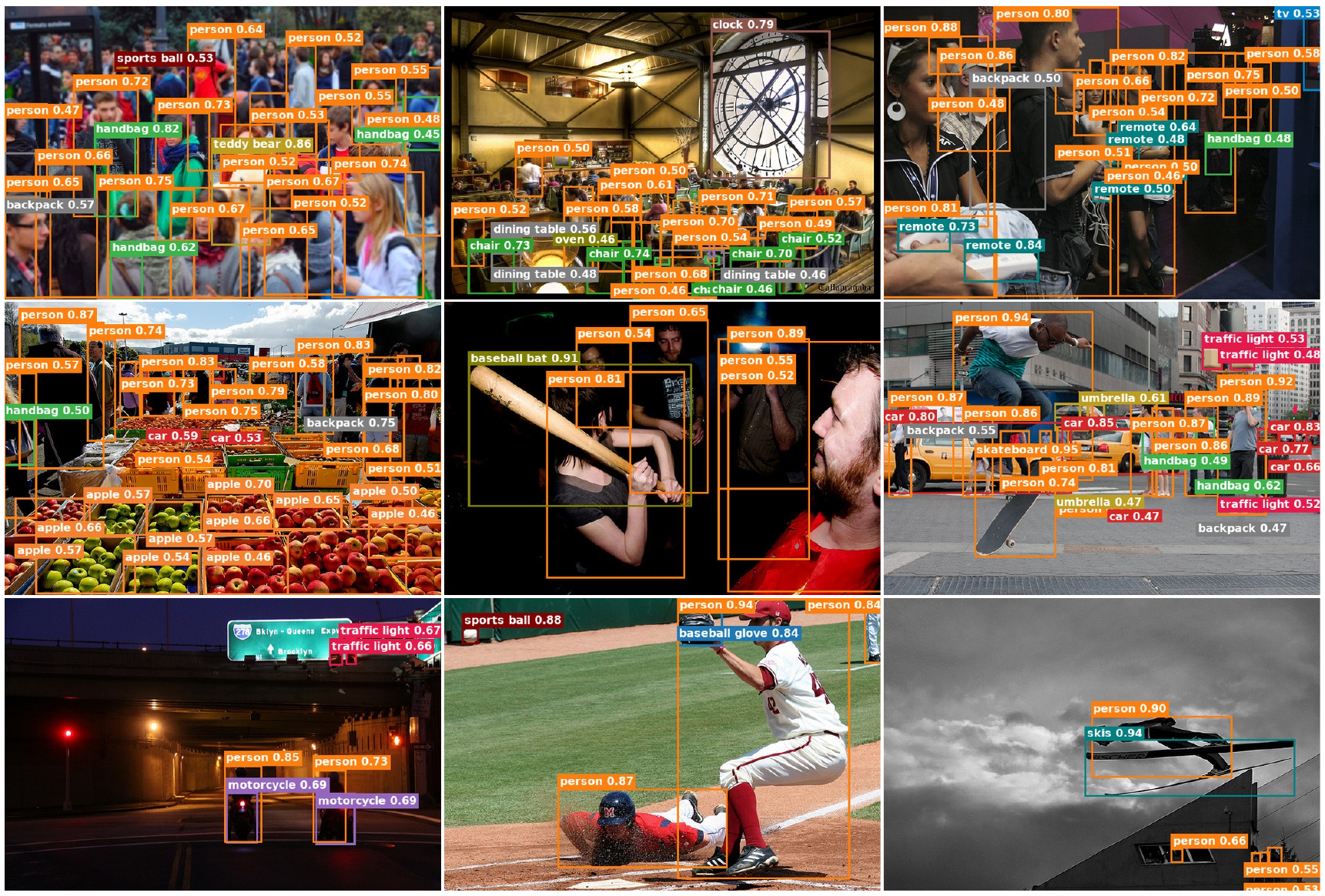}
    \caption{\textbf{Qualitative object detection results of GTR-X.}}
    \label{fig:detection_visualization}
\end{figure*}

%% file: figs_tex/results_segmentation.tex
\begin{figure*}[!t]
    \centering
    \includegraphics[width=0.90\textwidth]
    {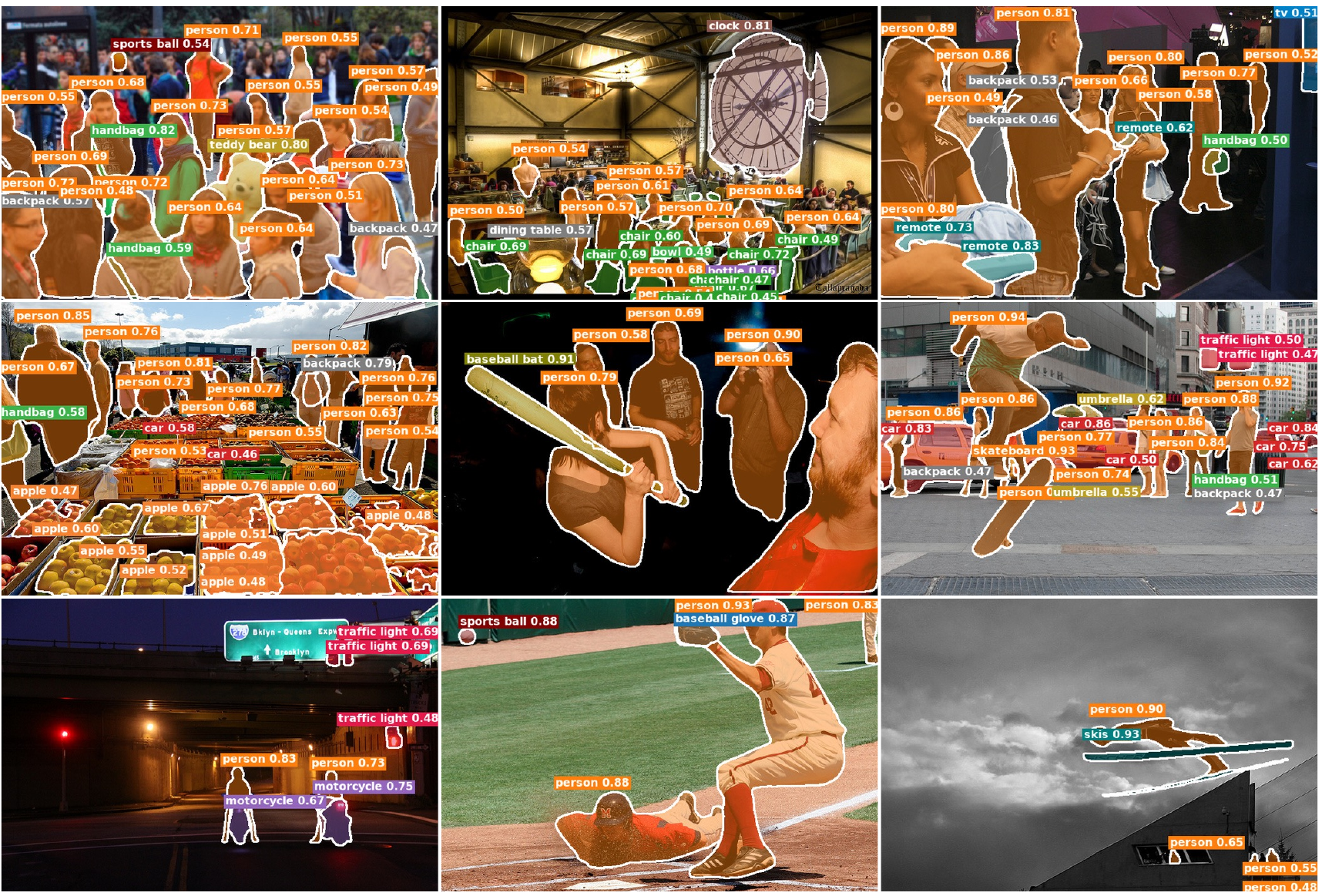}
    \caption{\textbf{Qualitative instance segmentation results of GTR-X.}
    Predictions are visualized on the same image set used in
    Figure~\ref{fig:detection_visualization}.
    }
    \label{fig:segmentation_visualization}
\end{figure*}

%% file: figs_tex/results_pose.tex
\begin{figure*}[!t]
    \centering
    \includegraphics[
        width=0.90\textwidth,
        pagebox=cropbox
    ]{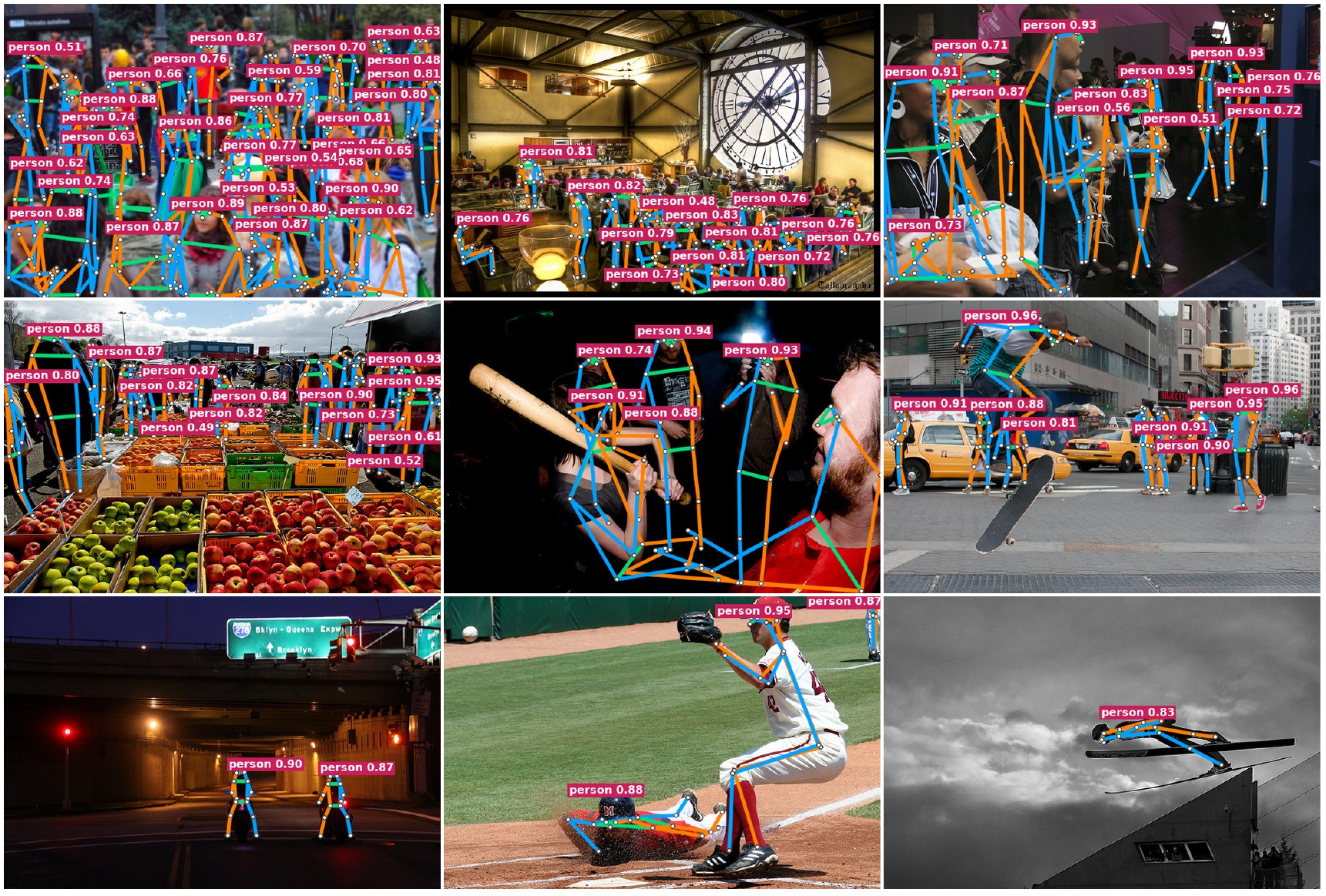}
    \caption{\textbf{Qualitative human pose estimation results of GTR-X.}
    The examples emphasize crowded scenes, low illumination, and motion
    blur.}
    \label{fig:pose_visualization}
\end{figure*}

%% file: figs_tex/results_semantic_segmentation.tex
\begin{figure*}[!t]
    \centering
    \includegraphics[
        width=0.90\textwidth,
        pagebox=cropbox
    ]{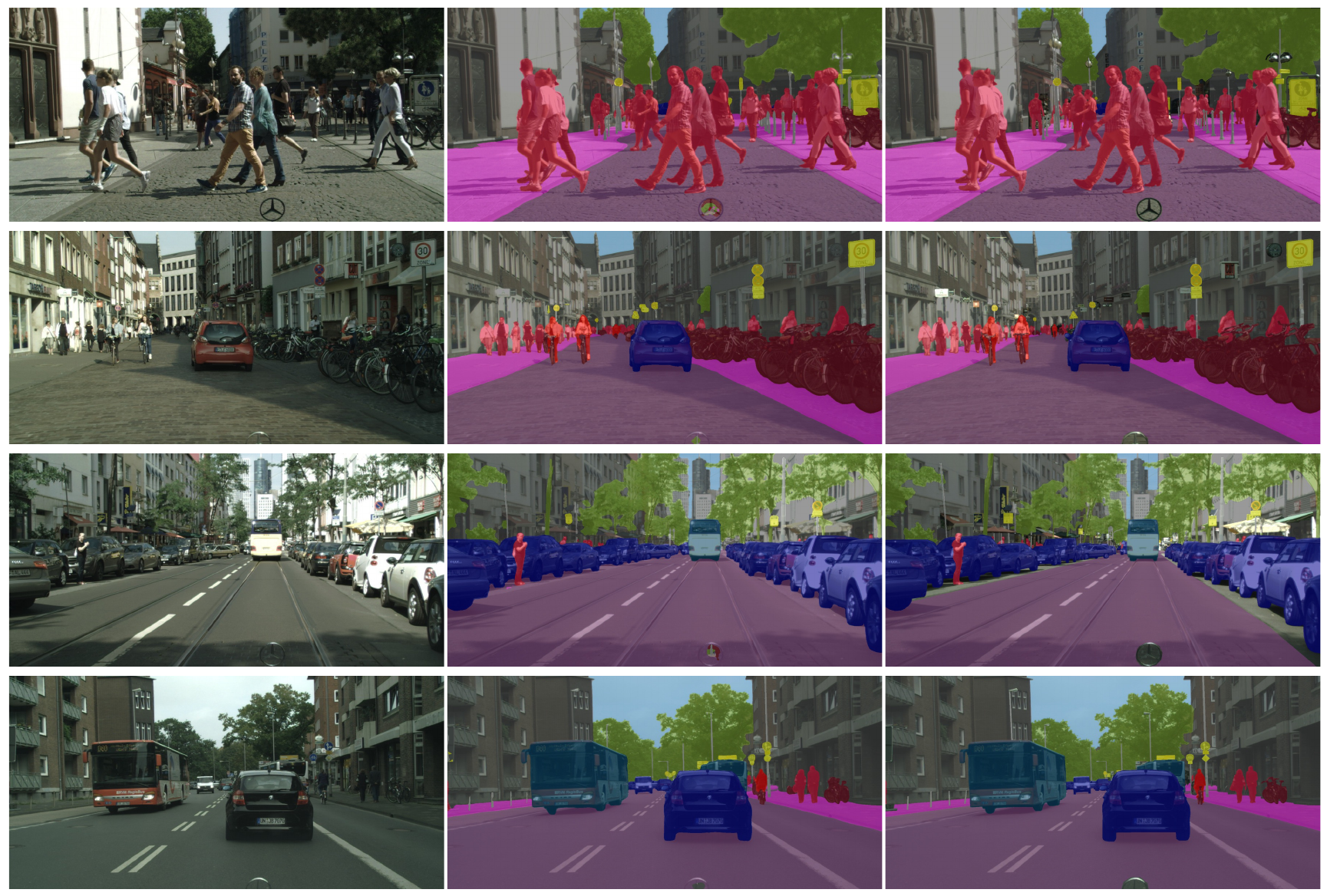}
    \caption{\textbf{Qualitative semantic segmentation results of GTR-X.}
    Each row shows, from left to right, the input RGB image, GTR-X prediction,
    and ground-truth annotation.}
    \label{fig:semantic_segmentation_visualization_1}
\end{figure*}

%% file: figs_tex/results_depth_dtu.tex
\begin{figure*}[!t]
    \centering
    \includegraphics[
        width=\textwidth,
        pagebox=cropbox
    ]{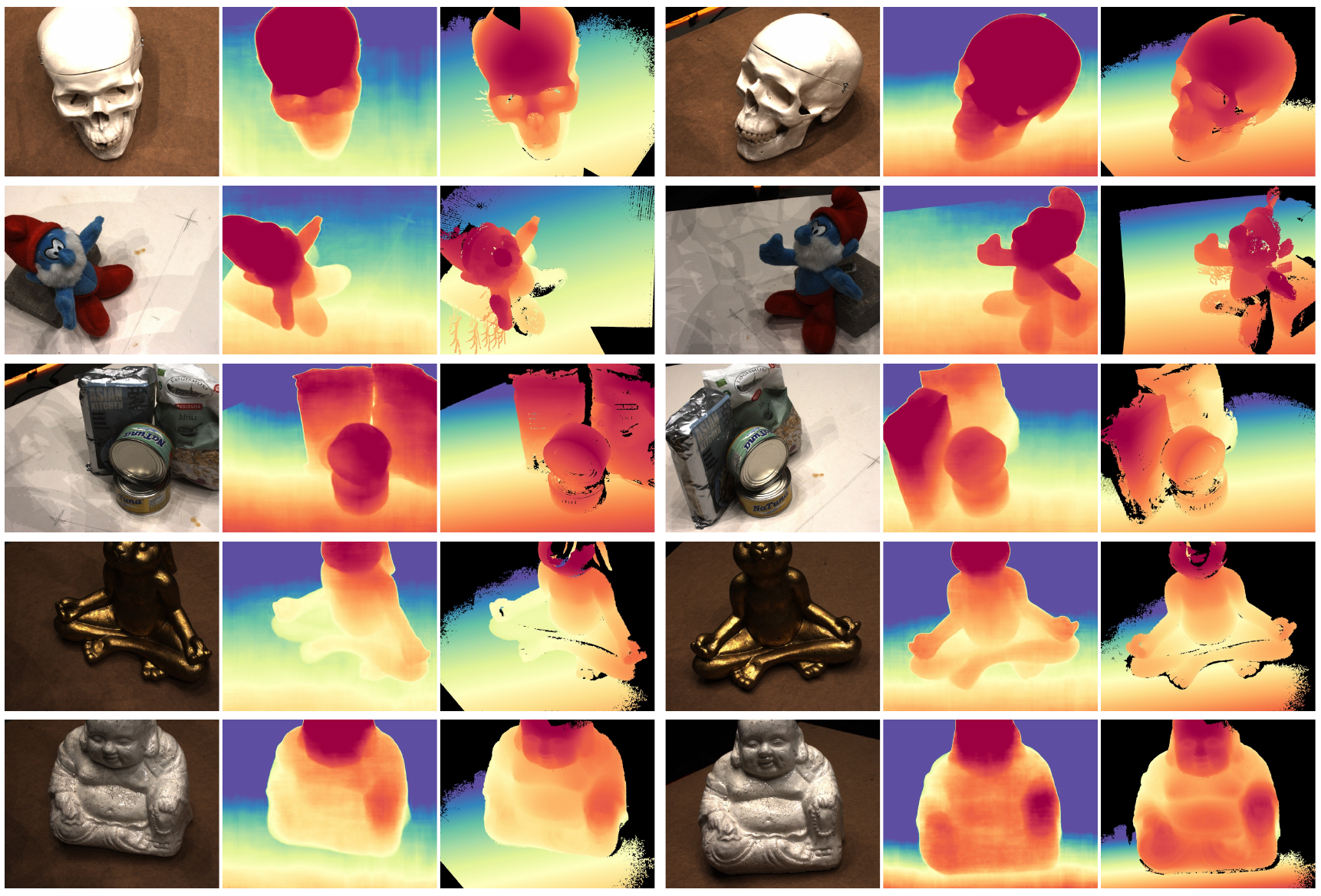}
    \caption{\textbf{Indoor monocular depth predictions of GTR-L on
    the DTU Robot Image Data Set~\citep{jensen2014large}.}
    Each three-column group shows the input RGB image, GTR-L prediction, and
    reference depth obtained by projecting the calibrated DTU point cloud.
    Black regions in the reference maps contain no valid projected depth.}
    \label{fig:depth_dtu_predictions}
\end{figure*}

\begin{figure*}[!t]
    \centering
    \includegraphics[
        width=\textwidth,
        pagebox=cropbox
    ]{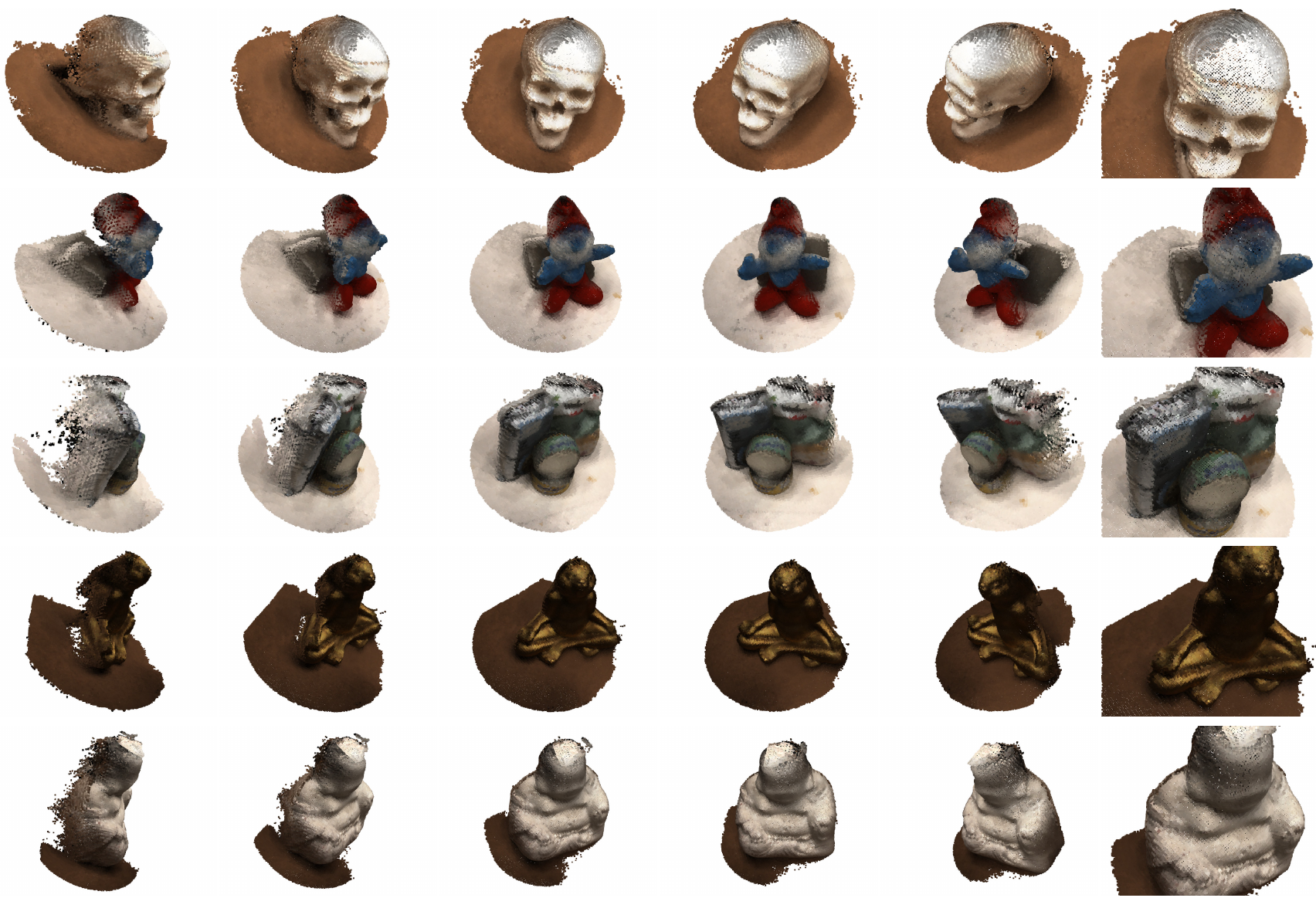}
    \caption{\textbf{Colored point clouds reconstructed from zero-shot GTR-L
    depth predictions on DTU~\citep{jensen2014large}.} Before fusion, each view is scale-aligned using the projected structured-light reference.}
    \label{fig:depth_dtu_reconstruction}
\end{figure*}

%% file: figs_tex/results_depth_nuscenes.tex
\begin{figure*}[!t]
    \centering
    \includegraphics[
        width=0.86\textwidth,
        pagebox=cropbox
    ]{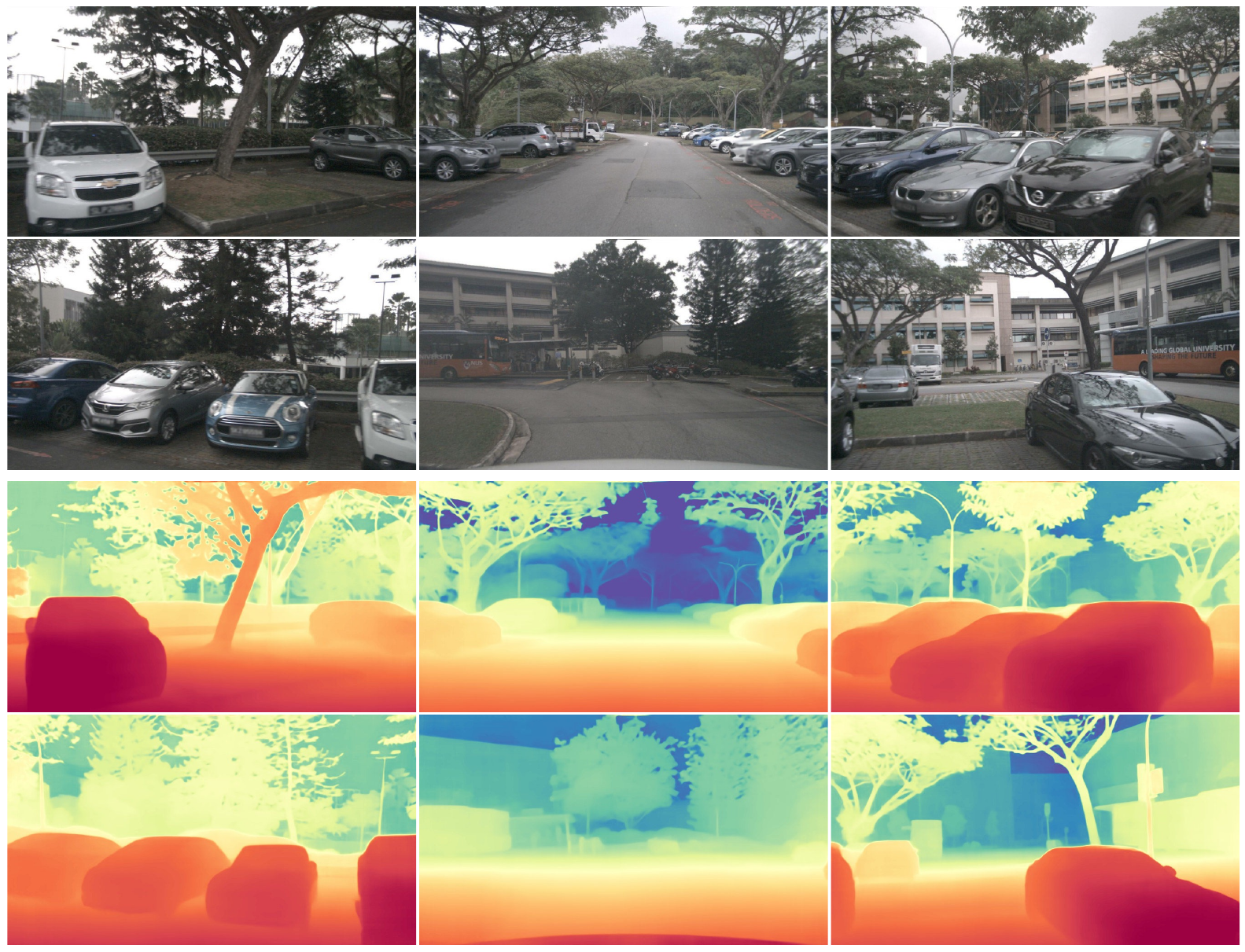}
    \caption{\textbf{Outdoor monocular depth predictions of GTR-L on
    nuScenes~\citep{caesar2020nuscenes} (set 1).}
    The first two rows show the six surround-view RGB images, and the last
    two rows show the corresponding predictions in the same row-major order.}
    \label{fig:depth_nuscenes_1}
\end{figure*}

\begin{figure*}[!t]
    \centering
    \includegraphics[
        width=0.86\textwidth,
        pagebox=cropbox
    ]{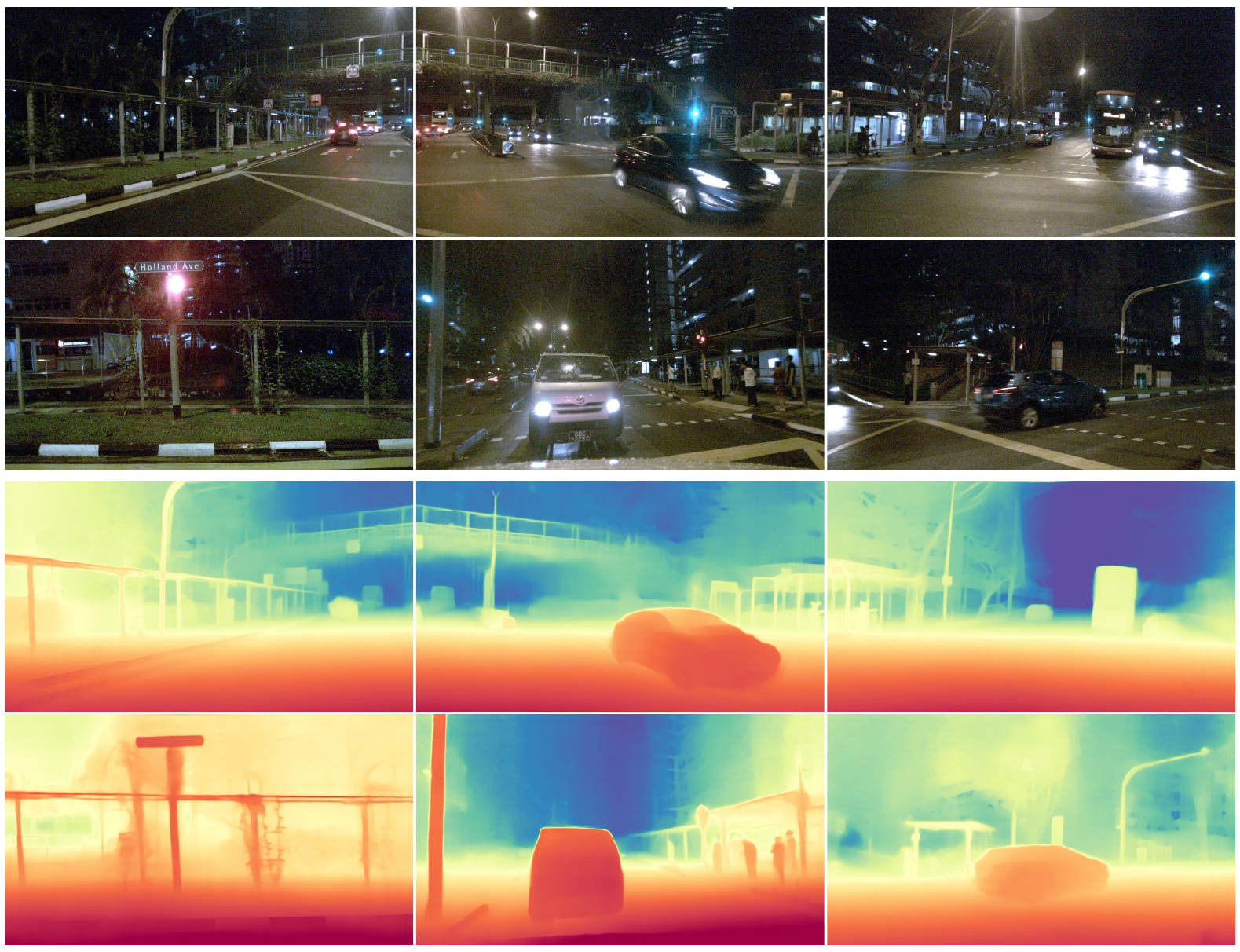}
    \caption{\textbf{Outdoor monocular depth predictions of GTR-L on
    nuScenes~\citep{caesar2020nuscenes} (set 2).}
    RGB inputs occupy the first two rows and their corresponding predictions
    occupy the last two rows in the same row-major order.}
    \label{fig:depth_nuscenes_2}
\end{figure*}

\begin{figure*}[!t]
    \centering
    \includegraphics[
        width=0.86\textwidth,
        pagebox=cropbox
    ]{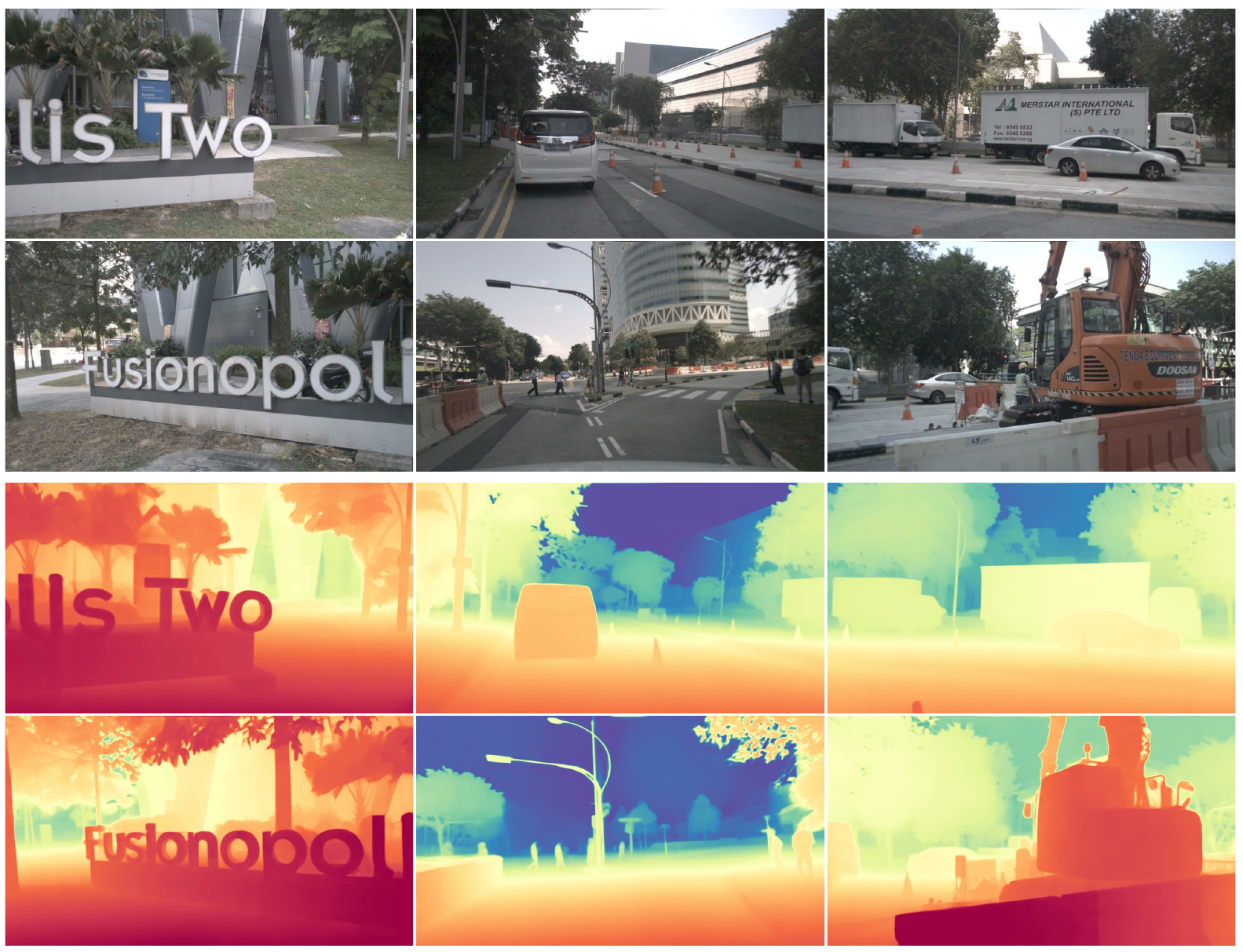}
    \caption{\textbf{Outdoor monocular depth predictions of GTR-L on
    nuScenes~\citep{caesar2020nuscenes} (set 3).}
    The layout follows Figure~\ref{fig:depth_nuscenes_1}: six RGB inputs are
    followed by their six corresponding depth predictions.}
    \label{fig:depth_nuscenes_3}
\end{figure*}

%% file: figs_tex/results_depth_nuscenes_mapping.tex
\begin{figure*}[!t]
    \centering
    \includegraphics[
        width=0.86\textwidth,
        pagebox=cropbox
    ]{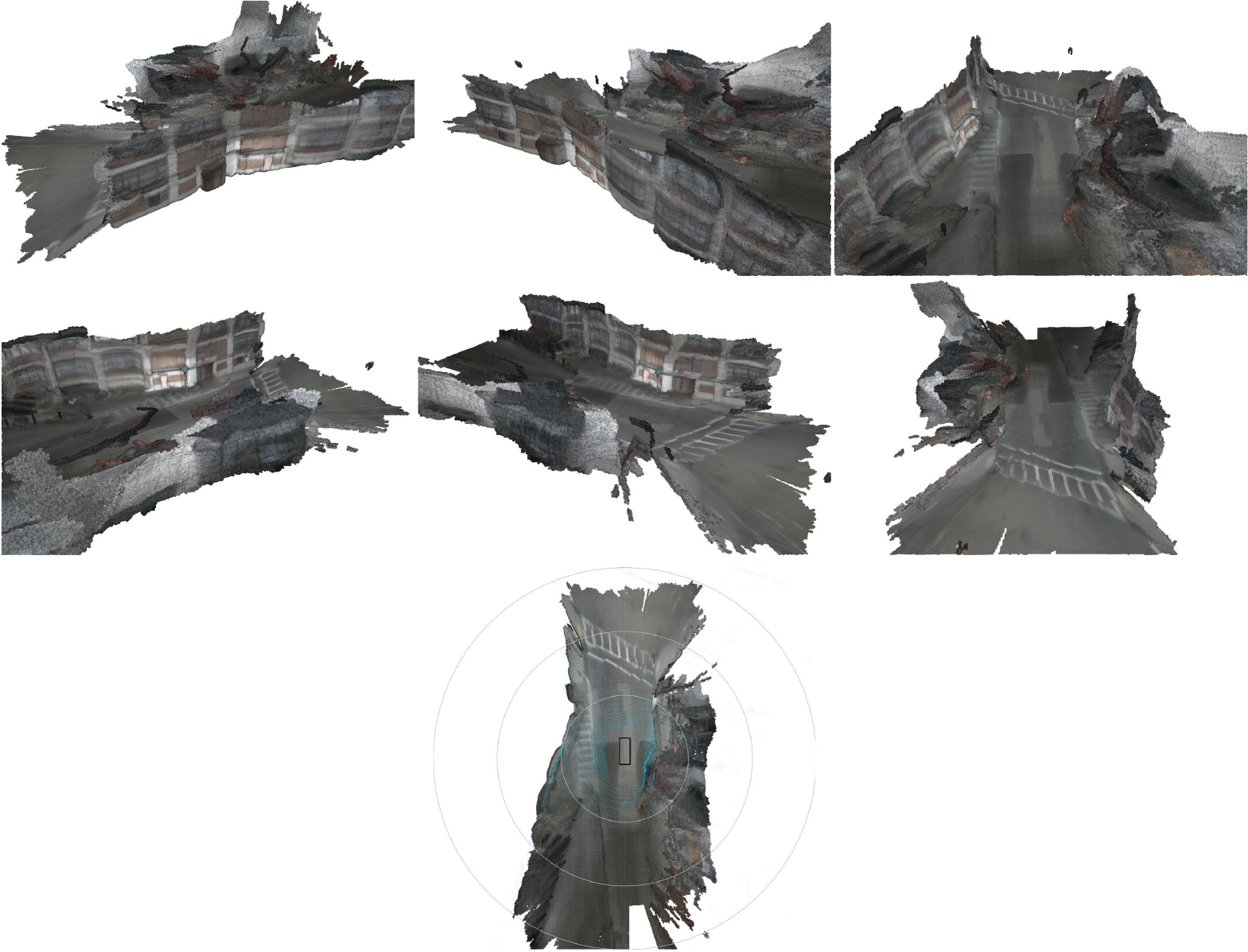}
    \caption{\textbf{Temporally fused local reconstruction on
    nuScenes~\citep{caesar2020nuscenes} (scene 1).} Each frame uses LiDAR-assisted scale alignment before TSDF fusion.}
    \label{fig:depth_nuscenes_bev_1}
\end{figure*}

\begin{figure*}[!t]
    \centering
    \includegraphics[
        width=0.86\textwidth,
        pagebox=cropbox
    ]{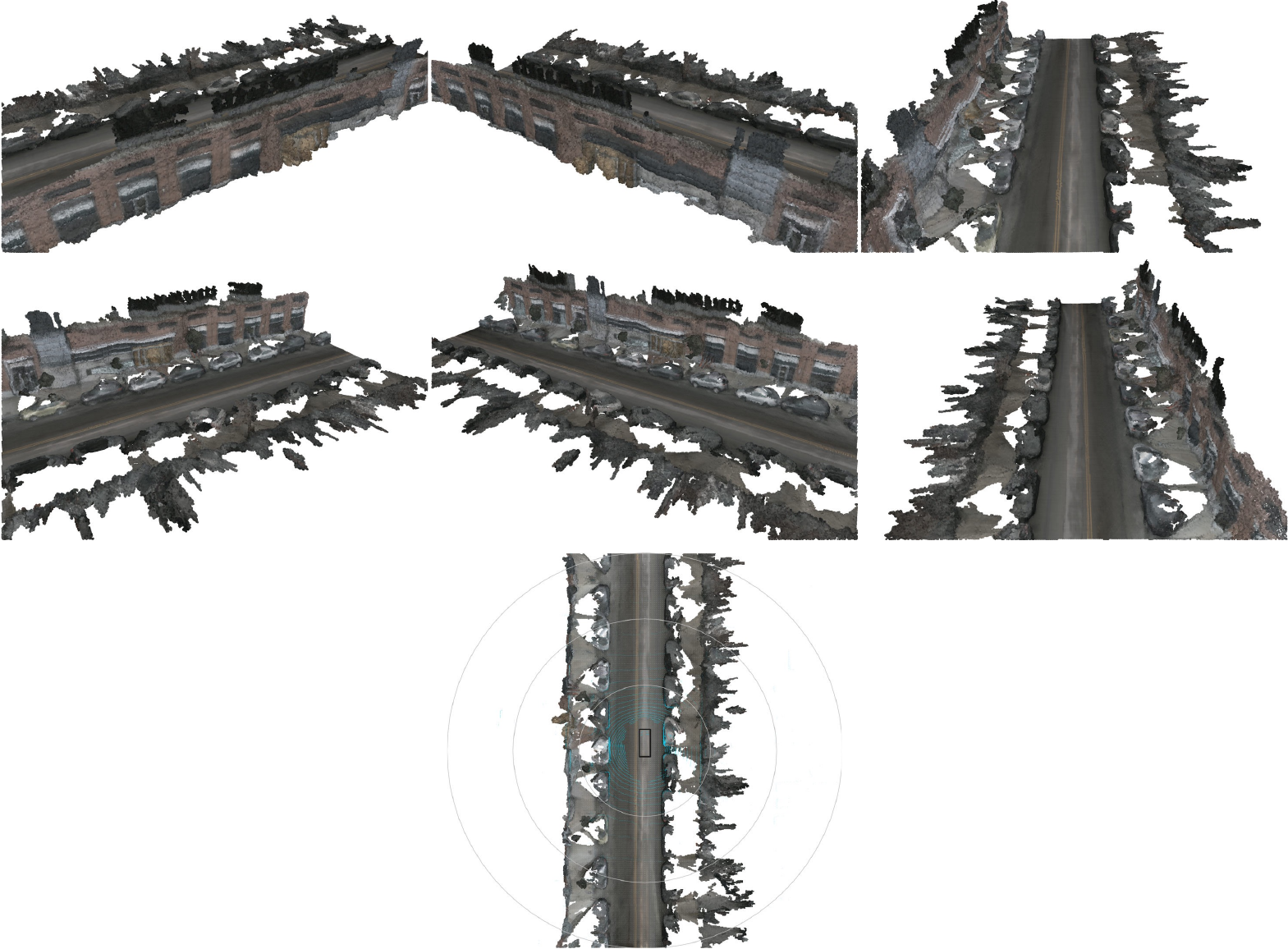}
    \caption{\textbf{Temporally fused local reconstruction on
    nuScenes~\citep{caesar2020nuscenes} (scene 2).} Each frame uses LiDAR-assisted scale alignment before TSDF fusion.}
    \label{fig:depth_nuscenes_bev_2}
\end{figure*}

%% file: figs_tex/results_obb.tex
\begin{figure*}[!t]
    \centering
    \includegraphics[
        width=\textwidth,
        pagebox=cropbox
    ]{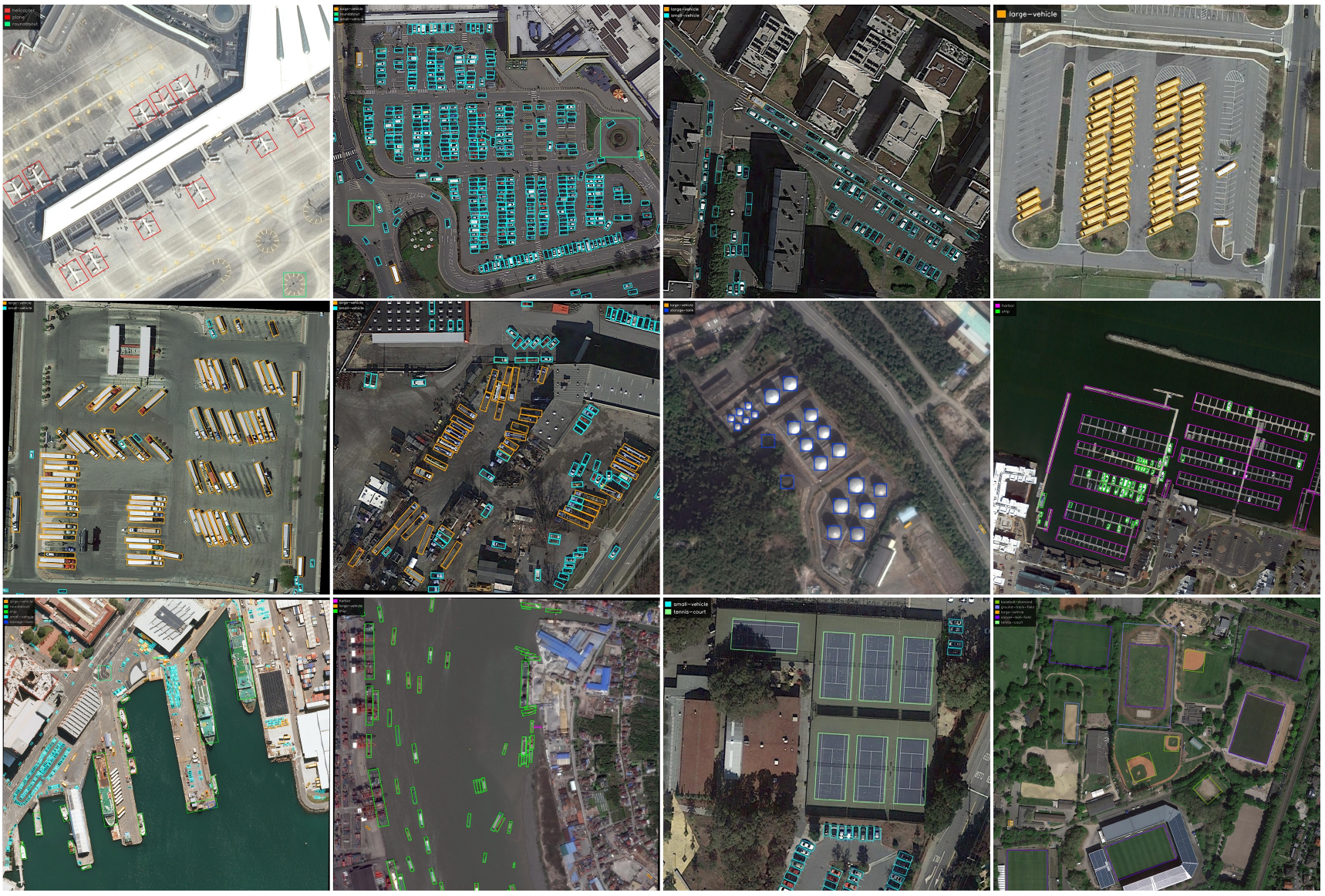}
    \caption{\textbf{Qualitative oriented object detection results on the
    DOTA-v1.0~\citep{xia2018dota} test set using GTR-X.}}
    \label{fig:obb_visualization}
\end{figure*}